\documentclass[11pt]{article}

\usepackage{amsmath, amssymb, amsthm}
\usepackage{graphicx}
\usepackage{fullpage}
\usepackage{hyperref}

\usepackage{epsf}
\usepackage{fancyheadings}
\usepackage{graphics}
\usepackage{amsfonts}
\usepackage{amsmath}
\usepackage{psfrag}

\def\real{\mathbb{R}}
\def\R{\mathbb{R}}
\def\pdim{d}
\newcommand{\usedim}{\ensuremath{\pdim}}
\def\betaopt{\theta^*}
\def\betahat{\widehat{\theta}}
\def\parset{\Omega}
\def\err{\Delta}

\def\norm#1{\|#1\|}
\def\loss{\mathcal{L}}
\def\reg{\mathcal{R}}

\def\half{\frac{1}{2}}
\def\numobs{n}
\def\spindex{s}

\theoremstyle{plain}

\newtheorem{theo}{Theorem}[section]

\newtheorem{lem}{Lemma}[section]
\newtheorem{prop}{Proposition}[section]
\newtheorem{cor}{Corollary}[section]

\theoremstyle{definition} 

\newtheorem{nota}{Notation}[section]
\newtheorem{de}{Definition}[section]
\newtheorem{exa}{Example}[section]
\newtheorem{as}{Assumption}[section]
\newtheorem{alg}{Algorithm}[section]

\newcommand{\btheo}{\begin{theo}}
\newcommand{\bde}{\begin{de}}
\newcommand{\ble}{\begin{lem}}
\newcommand{\bpr}{\begin{prop}}
\newcommand{\bno}{\begin{nota}}
\newcommand{\bex}{\begin{exa}}
\newcommand{\bcor}{\begin{cor}}
\newcommand{\spro}{\begin{proof}}
\newcommand{\bas}{\begin{as}}
\newcommand{\balg}{\begin{alg}}

\newcommand{\etheo}{\end{theo}}
\newcommand{\ede}{\end{de}}
\newcommand{\ele}{\end{lem}}
\newcommand{\epr}{\end{prop}}
\newcommand{\eno}{\end{nota}}
\newcommand{\eex}{\end{exa}}
\newcommand{\ecor}{\end{cor}}
\newcommand{\fpro}{\end{proof}}
\newcommand{\eas}{\end{as}}
\newcommand{\ealg}{\end{alg}}

\theoremstyle{plain}
\newtheorem{theos}{Theorem}
\newtheorem{props}{Proposition}
\newtheorem{lems}{Lemma}
\newtheorem{cors}{Corollary}
\newtheorem{defns}{Definition}

\theoremstyle{definition}
\newtheorem{exas}{Example}
\newtheorem{algs}{Algorithm}
\newtheorem{asss}{Assumption}

\newcommand{\btheos}{\begin{theos}}
\newcommand{\etheos}{\end{theos}}
\newcommand{\bprops}{\begin{props}}
\newcommand{\eprops}{\end{props}}
\newcommand{\bdes}{\begin{defns}}
\newcommand{\edes}{\end{defns}}
\newcommand{\blems}{\begin{lems}}
\newcommand{\elems}{\end{lems}}
\newcommand{\bcors}{\begin{cors}}
\newcommand{\ecors}{\end{cors}}
\newcommand{\bexs}{\begin{exas}}
\newcommand{\eexs}{\end{exas}}
\newcommand{\balgs}{\begin{algs}}
\newcommand{\ealgs}{\end{algs}}
\newcommand{\bass}{\begin{asss}}
\newcommand{\eass}{\end{asss}}

\renewcommand{\P}{\ensuremath{\mathbb{P}}}
\newcommand{\ip}[2]{\left\langle #1, #2\right\rangle}
\def\grad{\nabla}

\newcommand{\matsnorm}[2]{|\!|\!| #1 | \! | \!|_{{#2}}}
\newcommand{\myop}{\ensuremath{\operatorname{op}}}
\newcommand{\opnorm}[1]{\ensuremath{\matsnorm{#1}{\myop}}}
\newcommand{\frobnorm}[1]{\ensuremath{\matsnorm{#1}{F}}}
\newcommand{\order}{\mathcal{O}}
\newcommand{\ALLSLACK}{v}

\newcommand{\staterr}{\bar{\epsilon}_{\text{stat}}}
\newcommand{\coneslack}{\varepsilon}

\newcommand{\defn}{\ensuremath{:  =}}

\newcommand{\Xop}{\ensuremath{\mathfrak{X}}}

\newcommand{\inprod}[2]{\ensuremath{\langle #1 , \, #2 \rangle}}

\newcommand{\ThetaStar}{\ensuremath{\Theta^*}}
\newcommand{\ThetaHat}{\ensuremath{\widehat{\Theta}}}

\newcommand{\rdim}{\ensuremath{r}}

\newcommand{\errstat}{\epsilon_{\mbox{\tiny{stat}}}}
\newcommand{\sample}{\ensuremath{Z}}
\newcommand{\sampleset}{\ensuremath{\mathcal{Z}}}

\DeclareMathOperator{\trace}{trace}

\newcommand{\ind}{\ensuremath{\mathbb{I}}}

\makeatletter
\long\def\@makecaption#1#2{
        \vskip 0.8ex
        \setbox\@tempboxa\hbox{\small {\bf #1:} #2}
        \parindent 1.5em  
        \dimen0=\hsize
        \advance\dimen0 by -3em
        \ifdim \wd\@tempboxa >\dimen0
                \hbox to \hsize{
                        \parindent 0em
                        \hfil 
                        \parbox{\dimen0}{\def\baselinestretch{0.96}\small
                                {\bf #1.} #2
                                } 
                        \hfil}
        \else \hbox to \hsize{\hfil \box\@tempboxa \hfil}
        \fi
        }
\makeatother

\newenvironment{carlist}
 {\begin{list}{$\bullet$}
 {\setlength{\topsep}{0in} \setlength{\partopsep}{0in}
  \setlength{\parsep}{0in} \setlength{\itemsep}{\parskip}
  \setlength{\leftmargin}{0.07in} \setlength{\rightmargin}{0.08in}
  \setlength{\listparindent}{0in} \setlength{\labelwidth}{0.08in}
  \setlength{\labelsep}{0.1in} \setlength{\itemindent}{0.05in}}}
 {\end{list}}

\newcommand{\bcar}{\begin{carlist}}
\newcommand{\ecar}{\end{carlist}}

\newenvironment{proof-of-theorem}[1][{}]{\noindent{\bf Proof of
    Theorem~{#1}} \hspace*{1em}}{\qed\smallskip\\}

\newlength{\widebarargwidth}
\newlength{\widebarargheight}
\newlength{\widebarargdepth}
\DeclareRobustCommand{\widebar}[1]{%
  \settowidth{\widebarargwidth}{\ensuremath{#1}}%
  \settoheight{\widebarargheight}{\ensuremath{#1}}%
  \settodepth{\widebarargdepth}{\ensuremath{#1}}%
  \addtolength{\widebarargwidth}{-0.3\widebarargheight}%
  \addtolength{\widebarargwidth}{-0.3\widebarargdepth}%
  \makebox[0pt][l]{\hspace{0.3\widebarargheight}%
    \hspace{0.3\widebarargdepth}%
    \addtolength{\widebarargheight}{0.3ex}%
    \rule[\widebarargheight]{0.95\widebarargwidth}{0.1ex}}%
  {#1}}

\DeclareMathOperator{\col}{col}
\DeclareMathOperator{\row}{row}
\newcommand{\smallgap}{\delta}

\newcommand{\ball}[2]{\mathbb{B}_{#1}(#2)}

\newcommand{\mjwcomment}[1]{{\bf{\emph{MJW comment -- #1}}}}

\newcommand{\tracer}[2]{\ensuremath{\langle \!\langle {#1}, \; {#2}
\rangle \!\rangle}}

\long\def\comment#1{}

\newcommand{\binprod}[2]{\ensuremath{\big\langle #1 , \, #2 \big\rangle}}

\newcommand{\qpar}{\ensuremath{q}}
\newcommand{\radq}{\ensuremath{R_\qpar}}

\newcommand{\Parset}{\ensuremath{\parset}}
\newcommand{\mprob}{\ensuremath{\P}}

\newcommand{\iter}[1]{\ensuremath{\thetapar^{#1}}}

\newcommand{\radlag}{\ensuremath{\bar{\rho}}}

\newcommand{\Data}{\ensuremath{\sample_1^\numobs}}

\newcommand{\Reg}[1]{\ensuremath{\mathcal{R}(#1)}}

\newcommand{\ccon}{\ensuremath{\kappa}}
\newcommand{\cconu}{\ensuremath{\ccon_u}}
\newcommand{\cconl}{\ensuremath{\ccon_\ell}}

\newcommand{\Exs}{\ensuremath{\mathbb{E}}}

\newcommand{\betastar}{\ensuremath{\theta^*}}

\newcommand{\Sset}{\ensuremath{S}}

\newcommand{\CovMat}{\ensuremath{\Sigma}}

\newcommand{\kdim}{\ensuremath{s}}

\newcommand{\Ball}{\ensuremath{\mathbb{B}}}

\newcommand{\mdim}{\ensuremath{d}}

\newcommand{\plainnorm}[1]{\ensuremath{\|#1\|}}

\newcommand{\radbound}{\ensuremath{\radlag}}

\newcommand{\thetapar}{\ensuremath{\theta}}

\newcommand{\betahatlag}{\ensuremath{\widehat{\thetapar}_{\scriptsize{\regpar}}}}
\newcommand{\thetahatlag}{\ensuremath{\widehat{\thetapar}_{\scriptsize{\regpar}}}}
\newcommand{\betahatcon}{\ensuremath{\widehat{\thetapar}_{\scriptsize{\radcon}}}}

\newcommand{\composite}{\ensuremath{\phi}}

\newcommand{\matX}{X}
\newcommand{\Xmat}{\matX}

\renewcommand{\thetapar}{\ensuremath{\theta}}

\newcommand{\regpar}{\ensuremath{\lambda_\numobs}}

\newcommand{\thetastar}{\ensuremath{\theta^*}}

\newcommand{\BallReg}{\ensuremath{\mathbb{B}_{\tiny{\Regplain}}}}

\newcommand{\lossrsc}{\ensuremath{\gamma_\ell}}
\newcommand{\uplossrsc}{\ensuremath{\widebar{\lossrsc}}}

\newcommand{\Tay}{\ensuremath{\mathcal{T}_\loss}}

\newcommand{\Lsmooth}{\ensuremath{\gamma_u}}

\newcommand{\thetaparodd}{\ensuremath{\thetapar'}}

\newcommand{\thetahat}{\ensuremath{\widehat{\theta}}}

\newcommand{\PLAINCON}{\ensuremath{c}}

\newcommand{\widgraph}[2]{\includegraphics[keepaspectratio,width=#1]{#2}}

\newcommand{\toep}{\ensuremath{\omega}}

\newcommand{\Loss}{\ensuremath{\loss}}

\newcommand{\contrac}{\ensuremath{\kappa}}

\newcommand{\plaincon}{\ensuremath{c}}

\newcommand{\CovMatSqrt}{\ensuremath{\CovMat^{1/2}}}

\newcommand{\apar}{\ensuremath{v}}

\newcommand{\Iter}[1]{\ensuremath{\Theta^{#1}}}
\newcommand{\DelIt}[1]{\ensuremath{\Delta^{#1}}}
\newcommand{\DelHatIt}[1]{\ensuremath{\widehat{\Delta}^{#1}}}
\newcommand{\obj}{\phi} 

\newcommand{\objerror}{\widebar{\eta}}
\newcommand{\mjwerror}{\widebar{\eta}}

\newcommand{\FunHatIt}[1]{\FunPlain{#1}_\obj}
\newcommand{\FunPlain}[1]{\eta^{#1}}

\newcommand{\zhat}{\ensuremath{\widehat{z}}}
\newcommand{\Term}{\ensuremath{T}}

\newcommand{\LSPEC}{\ensuremath{L}}

\newcommand{\DeltaStar}{\ensuremath{\Delta^*}}

\newcommand{\BigProjModel}[1]{\Pi_{\ModelSet}(#1)}
\newcommand{\BigProjModelPerp}[1]{\Pi_{\ModelSet^\perp}(#1)}
\newcommand{\BigProjBset}[1]{\Pi_{\BsetSmall}(#1)}
\newcommand{\BigProjBsetPerp}[1]{\Pi_{\BsetSmall^\perp}(#1)}

\newcommand{\ProjModel}[1]{#1_{\ModelSet}}
\newcommand{\ProjModelPerp}[1]{#1_{\ModelSet^\perp}}
\newcommand{\ProjBset}[1]{#1_{\BsetSmall}}
\newcommand{\ProjBsetPerp}[1]{#1_{\BsetSmall^\perp}}

\newcommand{\ModelSet}{\ensuremath{\mathcal{M}}}
\newcommand{\ModelSetPerp}{\ensuremath{\ModelSet^\perp}}
\newcommand{\ModelPerp}{\ModelSetPerp}

\newcommand{\Bset}{\ensuremath{\widebar{\mathcal{M}}}}
\newcommand{\BsetSmall}{\ensuremath{\bar{\mathcal{M}}}}
\newcommand{\BsetPerp}{\ensuremath{{\Bset}^\perp}}

\newcommand{\Compat}{\ensuremath{\Psi}}
\newcommand{\slopa}{\xi}
\newcommand{\slopb}{\beta}

\newcommand{\radcon}{\ensuremath{\rho}}

\newcommand{\Plainreg}{\ensuremath{\mathcal{R}}}
\newcommand{\Regplain}{\Plainreg}

\newcommand{\NEWPHI}{\ensuremath{\varphi}}

\newcommand{\RegSq}[1]{\ensuremath{\mathcal{R}^2(#1)}}
\newcommand{\FUN}{\ensuremath{\tau}}
\newcommand{\FUNUP}{\ensuremath{\FUN_u}}
\newcommand{\FUNLOW}{\ensuremath{\FUN_\ell}}

\newcommand{\FUNLOWALL}{\FUNLOW(\LossN)}

\newcommand{\KeySet}{\ensuremath{\mathbb{S}}}

\newcommand{\HACKERRSQ}{\ensuremath{\nu^2(\DeltaStar; \ModelSet,
\Bset)}}

\newcommand{\HACKSTATERR}{\ensuremath{\epsilon^2(\DeltaStar;
\ModelSet, \Bset)}}

\newcommand{\LossN}{\ensuremath{\Loss_\numobs}}

\newcommand{\vsmall}{\vspace*{.05in}}

\newcommand{\lammin}{\ensuremath{\singval_{\operatorname{min}}}}
\newcommand{\lammax}{\ensuremath{\singval_{\operatorname{max}}}}

\newcommand{\phin}{\ensuremath{\chi_\numobs}}

\newcommand{\mutrun}{\ensuremath{\mu}}
\newcommand{\Stau}{\ensuremath{\Sset_\mutrun}}

\newcommand{\XopN}{\ensuremath{\Xop_\numobs}} 

\newcommand{\nuclear}[1]{\ensuremath{\matsnorm{#1}{1}}}
\newcommand{\frob}[1]{\ensuremath{\matsnorm{#1}{F}}}

\newcommand{\SHORT}{\Gamma^t} 

\newcommand{\DEL}{\ensuremath{\delta}}

\newcommand{\ANNOY}{\ensuremath{\alpha}}

\newcommand{\rhovec}{\ensuremath{\zeta}}
\newcommand{\rhomat}{\ensuremath{\zeta_{\tiny{\mbox{mat}}}}}

\newcommand{\Regdual}[1]{\Regplain^*(#1)}
\newcommand{\Plainregdual}{\Regplain^*}

\newcommand{\var}{\ensuremath{\operatorname{var}}}

\newcommand{\vect}{\ensuremath{\operatorname{vec}}}

\newcommand{\vtiny}{\ensuremath{\vspace*{.05in}}}

\newcommand{\Zdata}{\ensuremath{\sample_1^\numobs}}

\newcommand{\TmpVarA}{\ensuremath{\alpha}}
\newcommand{\TmpVarB}{\ensuremath{\beta}}
\newcommand{\Sbar}{\ensuremath{{\Sset^c}}}

\newcommand{\pdima}{\ensuremath{d_1}}
\newcommand{\pdimb}{\ensuremath{d_2}}

\newcommand{\singval}{\ensuremath{\sigma}}

\newcommand{\thetacon}{\ensuremath{\theta_t}}

\newcommand{\noisevar}{\ensuremath{\nu}}

\newcommand{\cov}{\ensuremath{\operatorname{cov}}}

\newcommand{\PREFACT}{\ensuremath{\alpha}}
\newcommand{\mynorm}[3]{\ensuremath{\|#1\|_{#2, #3}}}
\newcommand{\spike}{\ensuremath{\alpha}}
\newcommand{\GammaStar}{\ensuremath{\Gamma^*}}
\newcommand{\GammaHat}{\ensuremath{\widehat{\Gamma}}}
\newcommand{\mdima}{\ensuremath{{d_1}}}
\newcommand{\mdimb}{\ensuremath{{d_2}}}

\newcommand{\LinkFun}{\ensuremath{\Phi}}

\newcommand{\tol}{\ensuremath{\mjwerror}} 

\newcommand{\CRIT}{\ensuremath{\epsilon_{\tiny{\mbox{{tol}}}}}}

\newcommand{\PARNEW}{\ensuremath{\Parset'}}
\DeclareRobustCommand{\robrefthmcon}{\ref{ThmMainCon}}
\DeclareRobustCommand{\robrefthmlag}{\ref{ThmMainLag}}
\DeclareRobustCommand{\robreflemcone}{\ref{LemConeOpt}}
\DeclareRobustCommand{\robrefcorcomp}{\ref{CorMatComp}}
\newcommand{\barcontrac}{\ensuremath{\widebar{\contrac}}}

\begin{document}

\begin{center}
{\LARGE{{\bf{Fast global convergence of gradient methods for
        high-dimensional statistical recovery}}}}

\vspace*{.2in}

\begin{tabular}{ccc}
  Alekh Agarwal$^\dagger$
&
  Sahand N. Negahban$^\ddag$
&
  Martin J. Wainwright$^{\star, \dagger}$ \\
  \texttt{alekh@eecs.berkeley.edu} & 
  \texttt{sahandn@mit.edu} &
  \texttt{wainwrig@stat.berkeley.edu}
\end{tabular}

\vspace*{.2in}

\begin{tabular}{cc}
  Department of Statistics$^\star$, and & Department of EECS$^\ddag$, \\
  Department of EECS$^\dagger$, & Massachusetts Institute of Technology, \\
  University of California, Berkeley & 32 Vassar Street\\
  Berkeley, CA 94720-1776 & Cambirdge MA 02139
\end{tabular}

\vspace*{.2in}

\begin{abstract}
Many statistical $M$-estimators are based on convex optimization
problems formed by the combination of a data-dependent loss function
with a norm-based regularizer.  We analyze the convergence rates of
projected gradient and composite gradient methods for solving such
problems, working within a high-dimensional framework that allows the
data dimension $\pdim$ to grow with (and possibly exceed) the sample
size $\numobs$.  This high-dimensional structure precludes the usual
global assumptions---namely, strong convexity and smoothness
conditions---that underlie much of classical optimization analysis.
We define appropriately restricted versions of these conditions, and
show that they are satisfied with high probability for various
statistical models.  Under these conditions, our theory guarantees
that projected gradient descent has a globally geometric rate of
convergence up to the \emph{statistical precision} of the model,
meaning the typical distance between the true unknown parameter
$\theta^*$ and an optimal solution $\widehat{\theta}$.  This result is
substantially sharper than previous convergence results, which yielded
sublinear convergence, or linear convergence only up to the noise
level.  Our analysis applies to a wide range of $M$-estimators and
statistical models, including sparse linear regression using Lasso
($\ell_1$-regularized regression); group Lasso for block sparsity;
log-linear models with regularization; low-rank matrix recovery using
nuclear norm regularization; and matrix decomposition.  Overall, our
analysis reveals interesting connections between statistical precision
and computational efficiency in high-dimensional estimation.
\end{abstract}

\end{center}

\section{Introduction} 
\label{SecIntro}

High-dimensional data sets present challenges that are both
statistical and computational in nature.  On the statistical side,
recent years have witnessed a flurry of results on consistency and
rates for various estimators under non-asymptotic high-dimensional
scaling, meaning that error bounds are provided for general settings
of the sample size $\numobs$ and problem dimension $\pdim$, allowing
for the possibility that $\pdim \gg \numobs$.  These results typically
involve some assumption regarding the underlying structure of the
parameter space, such as sparse vectors, structured covariance
matrices, low-rank matrices, or structured regression functions, as
well as some regularity conditions on the data-generating process.  On
the computational side, many estimators for statistical recovery are
based on solving convex programs.  Examples of such $M$-estimators
include $\ell_1$-regularized quadratic programs (also known as the
Lasso) for sparse linear regression (e.g., see the
papers~\cite{Tibshirani96,Chen98,Huang06,Meinshausen06,BiRiTsy08,Bun07,GeerBuhl09}
and references therein), second-order cone programs (SOCP) for the
group Lasso (e.g.,~\cite{ZhaRocYu09,Lou09,HuaZha09} and references
therein), and semidefinite programming relaxations (SDP) for various
problems, including sparse PCA and low-rank matrix estimation
(e.g.,~\cite{CanRec08,Recht09,SreAloJaa05,
  AmiWai08,RohTsy10,NegWai09,RecFazPar10} and references therein).

Many of these programs are instances of convex conic programs, and so
can (in principle) be solved to $\epsilon$-accuracy in polynomial time
using interior point methods, and other standard methods from convex
programming (e.g., see the books~\cite{Bertsekas_nonlin,Boyd02}).
However, the complexity of such quasi-Newton methods can be
prohibitively expensive for the very large-scale problems that arise
from high-dimensional data sets. Accordingly, recent years have
witnessed a renewed interest in simpler first-order methods, among
them the methods of projected gradient descent and mirror descent.
Several authors (e.g.,~\cite{becker09nesta,Yi09,BeckTeb09}) have used
variants of Nesterov's accelerated gradient method~\cite{Nesterov07}
to obtain algorithms for high-dimensional statistical problems with a
sublinear rate of convergence.  Note that an optimization algorithm,
generating a sequence of iterates $\{\iter{t}\}_{t=0}^\infty$, is said
to exhibit \emph{sublinear convergence} to an optimum $\thetahat$ if
the optimization error $\plainnorm{\iter{t} - \thetahat}$ decays at
the rate $1/t^\kappa$, for some exponent $\kappa > 0$ and norm
$\plainnorm{\cdot}$.  Although this type of convergence is quite slow,
it is the best possible with gradient descent-type methods for convex
programs under only Lipschitz conditions~\cite{Nesterov04}.

It is known that much faster global rates---in particular, a linear or
geometric rate---can be achieved if global regularity conditions like
strong convexity and smoothness are imposed~\cite{Nesterov04}.  An
optimization algorithm is said to exhibit \emph{linear or geometric}
convergence if the optimization error $\plainnorm{\iter{t} -
  \thetahat}$ decays at a rate $\contrac^t$, for some contraction
coefficient $\contrac \in (0,1)$.  Note that such convergence is
exponentially faster than sub-linear convergence. For certain classes
of problems involving polyhedral constraints and global smoothness,
Tseng and Luo~\cite{LuoTse93} have established geometric convergence.
However, a challenging aspect of statistical estimation in high
dimensions is that the underlying optimization problems can never be
strongly convex in a global sense when $\pdim > \numobs$ (since the
$\pdim \times \pdim$ Hessian matrix is rank-deficient), and global
smoothness conditions cannot hold when $\pdim/\numobs \rightarrow
+\infty$.  Some more recent work has exploited structure specific to
the optimization problems that arise in statistical settings. For the
special case of sparse linear regression with random isotropic designs
(also referred to as compressed sensing), some authors have
established fast convergence rates in a local sense, meaning
guarantees that apply once the iterates are close enough to the
optimum~\cite{BreLor08,HalWotZha08}.  The intuition underlying these
results is that once an algorithm identifies the support set of the
optimal solution, the problem is then effectively reduced to a
lower-dimensional subspace, and thus fast convergence can be
guaranteed in a local sense.  Also in the setting of compressed
sensing, Tropp and Gilbert~\cite{TroppGil07} studied finite
convergence of greedy algorithms based on thresholding techniques, and
showed linear convergence up to a certain tolerance.  For the same
class of problems, Garg and Khandekar~\cite{GarKha09} showed that a
thresholded gradient algorithm converges rapidly up to some tolerance.
In both of these results, the convergence tolerance is of the order of
the noise variance, and hence substantially larger than the true
statistical precision of the problem.

\begin{figure}[h]
\begin{center}
\begin{tabular}{ccc}
\includegraphics[scale=0.4]{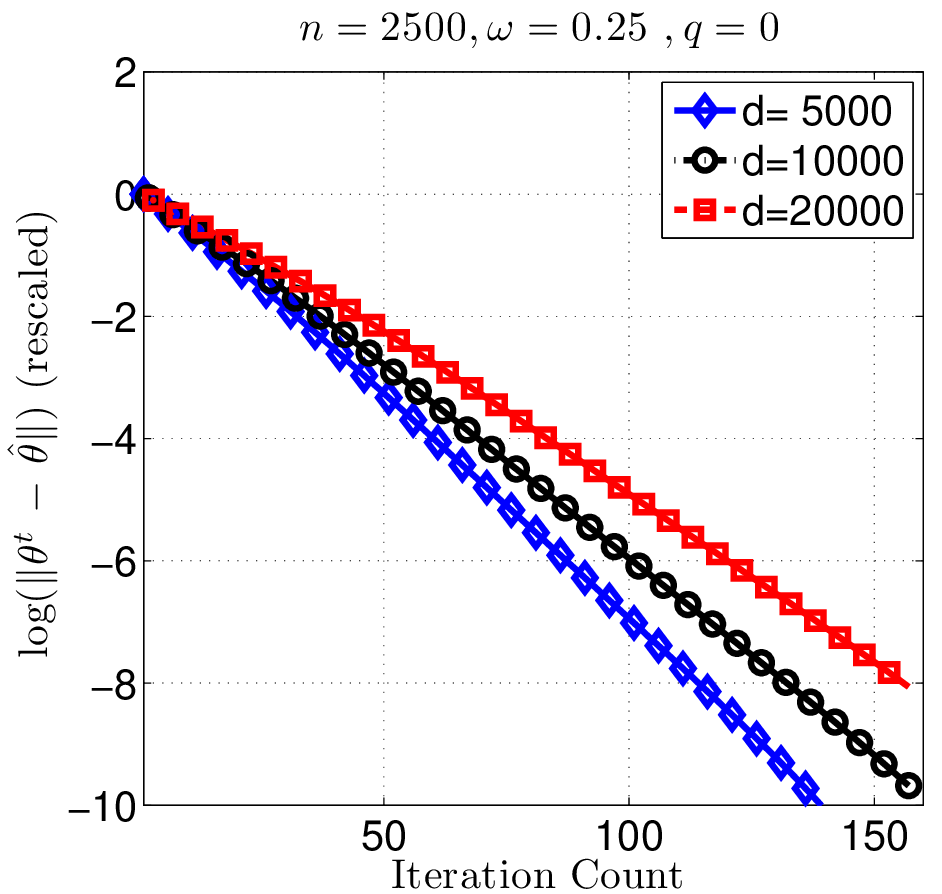} & \hspace*{.2in} & 
\includegraphics[scale=0.4]{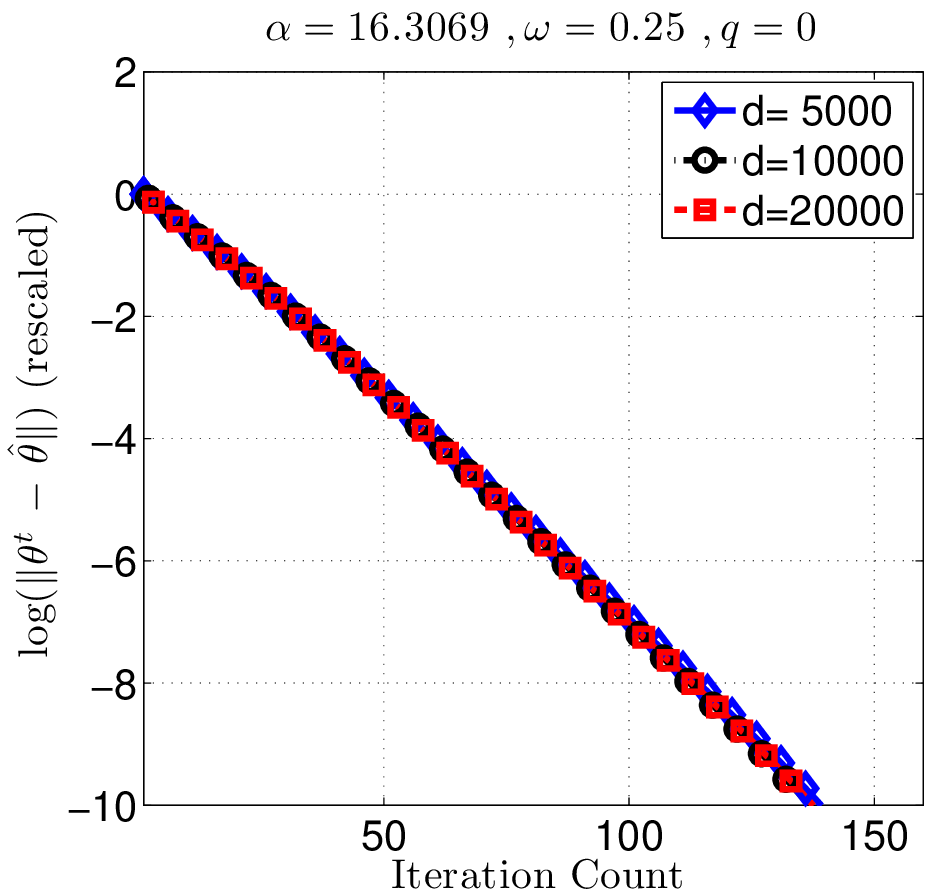} \\
(a) & & (b)
\end{tabular}
\caption{Convergence rates of projected gradient descent in
  application to Lasso programs ($\ell_1$-constrained least-squares).
  Each panel shows the log optimization error $\log
  \plainnorm{\iter{t} - \thetahat}$ versus the iteration number $t$.
  Panel (a) shows three curves, corresponding to dimensions $\pdim \in
  \{5000, 10000, 20000\}$, sparsity $\spindex = \lceil \sqrt{\pdim}
  \rceil$, and all with the same sample size $\numobs = 2500$.  All
  cases show geometric convergence, but the rate for larger problems
  becomes progressively slower.  (b) For an appropriately rescaled
  sample size ($\alpha = \frac{\numobs}{\spindex \log \pdim}$), all
  three convergence rates should be roughly the same, as predicted by
  the theory.}
\label{FigDim}
\end{center}
\end{figure}

The focus of this paper is the convergence rate of two simple
gradient-based algorithms for solving optimization problems that
underlie regularized $M$-estimators.  For a constrained problem with a
differentiable objective function, the projected gradient method
generates a sequence of iterates $\{\iter{t}\}_{t=0}^\infty$ by taking
a step in the negative gradient direction, and then projecting the
result onto the constraint set.  The composite gradient method of
Nesterov~\cite{Nesterov07} is well-suited to solving regularized
problems formed by the sum of a differentiable and (potentially)
non-differentiable component.  The main contribution of this paper is
to establish a form of global geometric convergence for these
algorithms that holds for a broad class of high-dimensional
statistical problems.  In order to provide intuition for this
guarantee, Figure~\ref{FigDim} shows the performance of projected
gradient descent for a Lasso problem ($\ell_1$-constrained
least-squares).  In panel (a), we have plotted the logarithm of the
optimization error, measured in terms of the Euclidean norm
$\plainnorm{\iter{t} - \thetahat}$ between the current iterate
$\iter{t}$ and an optimal solution $\thetahat$, versus the iteration
number $t$.  The plot includes three different curves, corresponding
to sparse regression problems in dimension $\pdim \in \{5000, 10000,
20000\}$, and a fixed sample size $\numobs = 2500$.  Note that all
curves are linear (on this logarithmic scale), revealing the geometric
convergence predicted by our theory.  Such convergence is not
predicted by classical optimization theory, since the objective
function cannot be strongly convex whenever $\numobs < \pdim$.
Moreover, the convergence is geometric even at early iterations, and
takes place to a precision far less than the noise level ($\noisevar^2
= 0.25$ in this example). We also note that the design matrix does not
satisfy the restricted isometry property, as assumed in some past
work.

The results in panel (a) exhibit an interesting property: the
convergence rate is \emph{dimension-dependent}, meaning that for a
fixed sample size, projected gradient descent converges more slowly
for a large problem than a smaller problem---compare the squares for
$\pdim = 20000$ to the diamonds for $\pdim = 5000$.  This phenomenon
reflects the natural intuition that larger problems are, in some
sense, ``harder'' than smaller problems.  A notable aspect of our
theory is that in addition to guaranteeing geometric convergence, it
makes a quantitative prediction regarding the extent to which a larger
problem is harder than a smaller one.  In particular, our convergence
rates suggest that if the sample size $\numobs$ is re-scaled in a
certain way according to the dimension $\pdim$ and also other model
parameters such as sparsity, then convergence rates should be roughly
similar.  Panel (b) provides a confirmation of this prediction: when
the sample size is rescaled according to our theory (in particular,
see Corollary~\ref{CorSparse} in Section~\ref{SecSparseVec}), then all
three curves lie essentially on top of another.

Although high-dimensional optimization problems are typically neither
strongly convex nor smooth, this paper shows that it is fruitful to
consider suitably restricted notions of strong convexity and
smoothness.  Our notion of restricted strong convexity (RSC) is
related to but slightly different than that introduced in a recent
paper by Negahban et al.~\cite{NegRavWaiYu09} for establishing
statistical consistency.  As we discuss in the sequel, bounding the
optimization error introduces new challenges not present when
analyzing the statistical error.  We also introduce a related notion
of restricted smoothness (RSM), not needed for proving
statistical rates but essential in the setting of optimization.  Our
analysis consists of two parts. We first show that for optimization
problems underlying many regularized $M$-estimators, appropriately
modified notions of restricted strong convexity (RSC) and smoothness
(RSM) are sufficient to guarantee global linear convergence of
projected gradient descent.  Our second contribution is to prove that
for the iterates generated by our first-order method, these RSC/RSM
assumptions do indeed hold with high probability for a broad class of
statistical models, among them sparse linear models, models with group
sparsity constraints, and various classes of matrix estimation
problems, including matrix completion and matrix decomposition.

An interesting aspect of our results is that the global geometric
convergence is not guaranteed to an arbitrary numerical precision, but
only to an accuracy related to \emph{statistical precision} of the
problem.  For a given error norm $\plainnorm{\cdot}$, given by the
Euclidean or Frobenius norm for most examples in this paper, the
statistical precision is given by the mean-squared error
$\Exs[\plainnorm{\betahat - \betaopt}^2]$ between the true parameter
$\betaopt$ and the estimate $\betahat$ obtained by solving the
optimization problem, where the expectation is taken over randomness
in the statistical model.  Note that this is very natural from the
statistical perspective, since it is the true parameter $\betaopt$
itself (as opposed to the solution $\betahat$ of the $M$-estimator)
that is of primary interest, and our analysis allows us to approach it
as close as is statistically possible.  Our analysis shows that we can
geometrically converge to a parameter $\theta$ such that
$\norm{\theta - \thetastar} = \norm{\thetahat - \thetastar} +
o(\norm{\thetahat - \thetastar})$, which is the best we can hope for
statistically, ignoring lower order terms. Overall, our results reveal
an interesting connection between the statistical and computational
properties of $M$-estimators---that is, the properties of the
underlying statistical model that make it favorable for estimation
also render it more amenable to optimization procedures.

The remainder of this paper is organized as follows.  We begin in
Section~\ref{SecBackground} with a precise formulation of the class of
convex programs analyzed in this paper, along with background on the
notions of a decomposable regularizer, and properties of the loss
function.  Section~\ref{SecMain} is devoted to the statement of our
main convergence result, as well as to the development and discussion
of its various corollaries for specific statistical models.  In
Section~\ref{SecSimulations}, we provide a number of empirical results
that confirm the sharpness of our theoretical predictions.  Finally,
Section~\ref{SecProofs} contains the proofs, with more technical
aspects of the arguments deferred to the Appendix.

\section{Background and problem formulation}
\label{SecBackground}

In this section, we begin by describing the class of regularized
$M$-estimators to which our analysis applies, as well as the
optimization algorithms that we analyze.  Finally, we introduce some
important notions that underlie our analysis, including the notions of
a decomposable regularization, and the properties of restricted strong
convexity and smoothness.

\subsection{Loss functions, regularization and gradient-based methods}

Given a random variable $\sample \sim \P$ taking values in some set
$\sampleset$, let \mbox{$\sample_1^\numobs =
  \{\sample_1,\dots,\sample_\numobs\}$} be a collection of $\numobs$
observations.  Here the integer $\numobs$ is the \emph{sample size} of
the problem.  Assuming that $\P$ lies within some indexed family
$\{\mprob_\thetapar, \thetapar \in \Parset\}$, the goal is to recover
an estimate of the unknown true parameter $\betaopt \in \Parset$
generating the data.  Here $\Parset$ is some subset of $\real^\pdim$,
and the integer $\pdim$ is known as the \emph{ambient dimension} of
the problem.  In order to measure the ``fit'' of any given parameter
$\thetapar \in \Parset$ to a given data set $\Zdata$, we introduce a
loss function \mbox{$\LossN: \parset \times \sampleset^\numobs
  \rightarrow \real_+$.}  By construction, for any given
$\numobs$-sample data set $\Zdata \in \sampleset^\numobs$, the loss
function assigns a cost $\LossN(\thetapar; \Zdata) \geq 0$ to the
parameter $\thetapar \in \Parset$.  In many (but not all)
applications, the loss function has a separable structure across the
data set, meaning that $\LossN(\thetapar; \Zdata) = \frac{1}{\numobs}
\sum_{i=1}^\numobs \ell(\thetapar; Z_i)$ where $\ell: \Parset \times
\sampleset: \rightarrow \real_+$ is the loss function associated with
a single data point. \\

Of primary interest in this paper are estimation problems that are
under-determined, meaning that the number of observations $\numobs$ is
smaller than the ambient dimension $\pdim$.  In such settings, without
further restrictions on the parameter space $\Parset$, there are
various impossibility theorems, asserting that consistent estimates of
the unknown parameter $\betaopt$ cannot be obtained.  For this reason,
it is necessary to assume that the unknown parameter $\betaopt$ either
lies within a smaller subset of $\Parset$, or is well-approximated by
some member of such a subset.  In order to incorporate these types of
structural constraints, we introduce a \emph{regularizer}
$\mathcal{R}: \parset \rightarrow \real_+$ over the parameter space.
With these ingredients, the analysis of this paper applies to the
\emph{constrained $M$-estimator}
\begin{align}
\label{EqnMestCon}
\betahatcon & \in \arg \min_{\Reg{\thetapar} \leq \radcon} \big \{
\LossN(\theta; \Zdata) \},
\end{align}
where $\radcon > 0$ is a user-defined radius, as well as to the
\emph{regularized $M$-estimator}
\begin{align}
\label{EqnMestReg}
\betahatlag & \in \arg \min_{\Reg{\thetapar} \leq \radlag} \bigl \{
\underbrace{\LossN(\thetapar; \sample_1^\numobs) + \regpar
  \Reg{\thetapar}}_{\composite_\numobs(\thetapar)} \bigr \}
\end{align}
where the regularization weight $\regpar > 0$ is user-defined. Note
that the radii $\radcon$ and $\radlag$ may be different in
general. Throughout this paper, we impose the following two
conditions:
\begin{enumerate}
\item[(a)] for any data set $\Zdata$, the function $\LossN(\cdot;
\Zdata)$ is convex and differentiable over $\Parset$, and
\item[(b)] the regularizer $\Regplain$ is a norm.
\end{enumerate}
These conditions ensure that the overall problem is convex, so that by
Lagrangian duality, the optimization problems~\eqref{EqnMestCon}
and~\eqref{EqnMestReg} are equivalent. However, as our analysis will
show, solving one or the other can be computationally more preferable
depending upon the assumptions made. Some remarks on notation: when the
radius $\radcon$ or the regularization parameter $\lambda_n$ is clear
from the context, we will drop the subscript on $\thetahat$ to ease
the notation.  Similarly, we frequently adopt the shorthand
$\LossN(\thetapar)$, with the dependence of the loss function on the
data being implicitly understood.  Procedures based on optimization
problems of either form are known as $M$-estimators in the statistics
literature.

The focus of this paper is on two simple algorithms for solving the
above optimization problems. The method of \emph{projected gradient
  descent} applies naturally to the constrained
problem~\eqref{EqnMestCon}, whereas the \emph{composite gradient
  descent} method due to Nesterov~\cite{Nesterov07} is suitable for
solving the regularized problem~\eqref{EqnMestReg}. Each routine
generates a sequence $\{\iter{t}\}_{t=0}^\infty$ of iterates by first
initializing to some parameter $\iter{0} \in \Parset$, and then
applying the recursive update
\begin{align}
\label{EqnAlgProj}
\iter{t+1} & = \arg \min_{\thetapar \in \BallReg(\radcon)} \big \{
\LossN(\iter{t}) + \inprod{\nabla \LossN(\iter{t})}{\theta - \iter{t}}
+ \frac{\Lsmooth}{2} \norm{\theta - \iter{t}}^2 \big \}, \qquad
\mbox{for $t = 0, 1, 2, \ldots$,}
\end{align}
in the case of projected gradient descent, or the update
\begin{align}
\label{EqnAlgReg}
\iter{t+1} & = \arg \min_{\thetapar \in \BallReg(\radlag)} \big \{
\LossN(\iter{t}) + \inprod{\nabla \LossN(\iter{t})}{\theta - \iter{t}}
+ \frac{\Lsmooth}{2} \norm{\theta - \iter{t}}^2 + \regpar
\Reg{\thetapar} \big \}, \qquad \mbox{for $t = 0, 1, 2, \ldots$,}
\end{align}
for the composite gradient method. Note that the only difference
between the two updates is the addition of the regularization term in
the objective.  These updates have a natural intuition: the next
iterate $\iter{t+1}$ is obtained by constrained minimization of a
first-order approximation to the loss function, combined with a
smoothing term that controls how far one moves from the current
iterate in terms of Euclidean norm.  Moreover, it is easily seen that
the update~\eqref{EqnAlgProj} is equivalent to
\begin{align}
  \label{EqnProjVersion}
  \iter{t+1} &= \Pi \bigg(\iter{t} - \frac{1}{\Lsmooth}\nabla
  \LossN(\iter{t})\bigg),
\end{align}
where $\Pi \equiv \Pi_{\BallReg(\radcon)}$ denotes Euclidean
projection onto the ball $\BallReg(\radcon) = \{ \thetapar \in \Parset
\, \mid \, \Reg{\thetapar} \leq \radcon \}$ of radius $\radcon$.  In
this formulation, we see that the algorithm takes a step in the
negative gradient direction, using the quantity $1/\Lsmooth$ as
stepsize parameter, and then projects the resulting vector onto the
constraint set. The update~\eqref{EqnAlgReg} takes an analogous form,
however, the projection will depend on both $\regpar$ and
$\Lsmooth$. As will be illustrated in the examples to follow, for many
problems, the updates~\eqref{EqnAlgProj} and~\eqref{EqnAlgReg}, or
equivalently~\eqref{EqnProjVersion}, have a very simple solution.  For
instance, in the case of $\ell_1$-regularization, it can be obtained
by an appropriate form of the soft-thresholding operator.

\subsection{Restricted strong convexity and smoothness}
\label{SecRSC}

In this section, we define the conditions on the loss function and
regularizer that underlie our analysis.  Global smoothness and strong
convexity assumptions play an important role in the classical analysis
of optimization
algorithms~\cite{Bertsekas_nonlin,Boyd02,Nesterov04}. In application
to a differentiable loss function $\LossN$, both of these properties
are defined in terms of a first-order Taylor series expansion around a
vector $\thetaparodd$ in the direction of $\thetapar$---namely, the
quantity
\begin{align}
\label{EqnTayError}
\Tay(\thetapar; \thetaparodd) & \defn \LossN(\thetapar) -
\LossN(\thetaparodd) - \inprod{\nabla \LossN(\thetaparodd)}{\thetapar -
  \thetaparodd}.
\end{align}
By the assumed convexity of $\LossN$, this error is always
non-negative, and global strong convexity is equivalent to imposing a
stronger condition, namely that for some parameter $\lossrsc > 0$, the
first-order Taylor error $\Tay(\thetapar; \thetaparodd)$ is lower
bounded by a quadratic term $\frac{\lossrsc}{2} \,
\plainnorm{\thetapar - \thetaparodd}^2$ for all $\thetapar,
\thetaparodd \in \Parset$.  Global smoothness is defined in a similar
way, by imposing a quadratic upper bound on the Taylor error.  It is
known that under global smoothness and strong convexity assumptions,
the method of projected gradient descent~\eqref{EqnAlgProj} enjoys a
\emph{globally geometric convergence rate}, meaning that there is some
$\contrac \in (0,1)$ such that\footnote{In this statement (and
  throughout the paper), we use $\lesssim$ to mean an inequality that
  holds with some universal constant $c$, independent of the problem
  parameters.}
\begin{align}
\norm{\iter{t} - \betahat}^2 & \lesssim \; \contrac^t \;
\norm{\iter{0} - \betahat}^2 \qquad \mbox{for all iterations $t = 0,
  1, 2, \ldots$}.
\end{align}
We refer the reader to Bertsekas~\cite[Prop. 1.2.3,
  p. 145]{Bertsekas_nonlin}, or Nesterov~\cite[Thm. 2.2.8,
  p. 88]{Nesterov04} for such results on projected gradient descent,
and to Nesterov~\cite{Nesterov07} for composite gradient descent.

Unfortunately, in the high-dimensional setting ($\pdim > \numobs$), it
is usually impossible to guarantee strong convexity of the
problem~\eqref{EqnMestCon} in a global sense.  For instance, when the
data is drawn i.i.d., the loss function consists of a sum of $\numobs$
terms. If the loss is twice differentiable, the resulting $\pdim
\times \pdim$ Hessian matrix $\nabla^2 \loss(\thetapar; \Data)$ is
often a sum of $\numobs$ matrices each with rank one, so that the
Hessian is rank-degenerate when $\numobs < \pdim$.  However, as we
show in this paper, in order to obtain fast convergence rates for the
optimization method~\eqref{EqnAlgProj}, it is sufficient that (a) the
objective is strongly convex and smooth in a restricted set of
directions, and (b) the algorithm approaches the optimum $\betahat$
only along these directions.  Let us now formalize these ideas.\\

\bdes[{\bf{Restricted strong convexity (RSC)}}] The loss function
$\LossN$ satisfies restricted strong convexity with respect to
$\Regplain$ and with parameters $(\lossrsc, \FUNLOW(\LossN))$ over the
set $\PARNEW$ if
\begin{align}
\label{EqnRSC}
\Tay(\thetapar; \thetaparodd) & \geq \frac{\lossrsc}{2} \,
\plainnorm{\thetapar - \thetaparodd}^2 - \FUNLOW(\LossN) \;
\RegSq{\thetapar - \thetaparodd} \qquad \mbox{for all $\thetapar,
  \thetaparodd \in \PARNEW$.}
\end{align}
\edes 
\noindent We refer to the quantity $\lossrsc$ as the \emph{(lower)
  curvature parameter}, and to the quantity $\FUNLOW$ as the
\emph{tolerance parameter}.  The set $\PARNEW$ corresponds to a
suitably chosen subset of the space $\Parset$ of all possible
parameters.

In order to gain intuition for this definition, first suppose that the
condition~\eqref{EqnRSC} holds with tolerance parameter $\FUNLOW = 0$.
In this case, the regularizer plays no role in the definition, and
condition~\eqref{EqnRSC} is equivalent to the usual definition of
strong convexity on the optimization set $\Parset$.  As discussed
previously, this type of global strong convexity typically
\emph{fails} to hold for high-dimensional inference problems.  In
contrast, when tolerance parameter $\FUNLOW$ is strictly positive, the
condition~\eqref{EqnRSC} is much milder, in that it only applies to a
\emph{limited set} of vectors.  For a given pair $\thetapar \neq
\thetaparodd$, consider the inequality
\begin{align}
\label{EqnFACup}
\frac{\RegSq{\thetapar - \thetaparodd}}{\plainnorm{\thetapar -
    \thetaparodd}^2} & < \frac{\lossrsc}{2 \, \FUNLOW(\LossN)}.
\end{align}
If this inequality is violated, then the right-hand side of the
bound~\eqref{EqnRSC} is non-positive, in which case the RSC
constraint~\eqref{EqnRSC} is vacuous.  Thus, restricted strong
convexity imposes a non-trivial constraint only on pairs $\thetapar
\neq \thetaparodd$ for which the inequality~\eqref{EqnRSC} holds, and
a central part of our analysis will be to prove that, for the sequence
of iterates generated by projected gradient descent, the optimization
error $\DelHatIt{t} \defn \iter{t} - \thetahat$ satisfies a constraint
of the form~\eqref{EqnFACup}.  We note that since the regularizer
$\Regplain$ is convex, strong convexity of the loss function $\LossN$
also implies the strong convexity of the regularized loss
$\composite_\numobs$ as well.

For the least-squares loss, the RSC definition depends purely on the
direction (and not the magnitude) of the difference vector $\thetapar
- \thetaparodd$.  For other types of loss functions---such as those
arising in generalized linear models---it is essential to localize the
RSC definition, requiring that it holds only for pairs for which the
norm $\|\thetapar - \thetaparodd\|_2$ is not too large.  We refer the
reader to Section~\ref{SecLogLinear} for further discussion of this
issue.

Finally, as pointed out by a reviewer, our restricted version of
strong convexity can be seen as an instance of the general theory of
paraconvexity (e.g.,~\cite{NgailPe2008}); however, we are not aware of
convergence rates for minimizing general paraconvex functions. \\

\noindent We also specify an analogous notion of restricted smoothness:

\bdes[{\bf{Restricted smoothness (RSM)}}] We say the loss function
$\LossN$ satisfies restricted smoothness with respect to $\Regplain$
and with parameters $(\Lsmooth, \FUNUP(\LossN))$ over the set
$\PARNEW$ if
\begin{align}
\label{EqnRSM}
\Tay(\thetapar; \thetaparodd) & \leq \frac{\Lsmooth}{2} \,
\plainnorm{\thetapar - \thetaparodd}^2 + \FUNUP(\LossN) \;
\RegSq{\thetapar - \thetaparodd} \qquad \mbox{for all $\thetapar,
\thetaparodd \in \PARNEW$.}
\end{align}
\edes ~\\

\noindent 
As with our definition of restricted strong convexity, the additional
tolerance $\FUNUP(\LossN)$ is not present in analogous smoothness
conditions in the optimization literature, but it is essential in our
set-up.

\subsection{Decomposable regularizers}
\label{SecDecomp}

In past work on the statistical properties of regularization, the
notion of a decomposable regularizer has been shown to be
useful~\cite{NegRavWaiYu09}.  Although the focus of this paper is a
rather different set of questions---namely, optimization as opposed to
statistics---decomposability also plays an important role here.  
Decomposability is defined with respect to a pair of subspaces defined
with respect to the parameter space $\Parset \subseteq \real^\pdim$.
The set $\ModelSet$ is known as the \emph{model subspace}, whereas the
set $\BsetPerp$, referred to as the \emph{perturbation subspace},
captures deviations away from the model subspace.
\bdes
\label{DefnDecomp}
Given a subspace pair $(\ModelSet, \BsetPerp)$ such that $\ModelSet
\subseteq \Bset$, we say that a norm $\Plainreg$ is \mbox{$(\ModelSet,
  \BsetPerp)$-\emph{decomposable}} if
\begin{align}
\Reg{\alpha + \beta} & = \Reg{\alpha} + \Reg{\beta} \qquad \mbox{for
all $\alpha \in \ModelSet$ and $\beta \in \BsetPerp$.}
\end{align}
\edes
\noindent To gain some intuition for this definition, note that by
triangle inequality, we always have the bound $\Reg{\alpha + \beta}
\leq \Reg{\alpha} + \Reg{\beta}$.  For a decomposable regularizer,
this inequality always holds with equality.  Thus, given a fixed
vector $\alpha \in \ModelSet$, the key property of any decomposable
regularizer is that it affords the \emph{maximum penalization} of any
deviation $\beta \in \BsetPerp$. \\

For a given error norm $\norm{\cdot}$, its interaction with the
regularizer $\Plainreg$ plays an important role in our results.  In
particular, we have the following:
\bdes[Subspace compatibility] 
\label{DefnSubspaceCompat}
Given the regularizer $\Reg{\cdot}$ and a norm $\norm{\cdot}$, the
associated \emph{subspace compatibility} is given by
\begin{align}
\label{EqnDefnSubspaceCompat}
\Compat(\Bset) & \defn \sup_{\theta \in \BsetSmall \backslash \{0\}}
\frac{\Reg{\theta}}{\norm{\theta}} \qquad \mbox{when $\Bset \neq
  \{0\}$}, \qquad \mbox{and $\Compat(\{0\}) \defn 0$.}
\end{align}

\edes
\noindent The quantity $\Compat(\Bset)$ corresponds to the Lipschitz
constant of the norm $\Regplain$ with respect to $\norm{\cdot}$, when
restricted to the subspace $\Bset$.


\subsection{Some illustrative examples}

We now describe some particular examples of $M$-estimators with
decomposable regularizers, and discuss the form of the projected
gradient updates as well as RSC/RSM conditions.  We cover two main
families of examples: log-linear models with sparsity constraints and
$\ell_1$-regularization (Section~\ref{SecLogLinear}), and matrix
regression problems with nuclear norm regularization
(Section~\ref{SecMatrixReg}).

\subsubsection{Sparse log-linear models and $\ell_1$-regularization}
\label{SecLogLinear}
Suppose that each sample $Z_i$ consists of a scalar-vector pair $(y_i,
x_i) \in \real \times \real^\pdim$, corresponding to the scalar
response $y_i \in \mathcal{Y}$ associated with a vector of predictors
$x_i \in \real^\pdim$.  A log-linear model with canonical link
function assumes that the response $y_i$ is linked to the covariate
vector $x_i$ via a conditional distribution of the form $\mprob(y_i
\mid x_i; \thetastar, \sigma) \propto \exp \biggr \{ \frac{y_i \,
  \inprod{\thetastar}{x_i} -
  \LinkFun(\inprod{\thetastar}{x_i})}{c(\sigma)} \biggr \}$, where
$c(\sigma)$ is a known quantity, $\LinkFun(\cdot)$ is the log-partition
function to normalize the density, and $\thetastar \in \real^\pdim$ is
an unknown regression vector.  In many applications, the regression
vector $\thetastar$ is relatively sparse, so that it is natural to
impose an $\ell_1$-constraint.  Computing the maximum likelihood
estimate subject to such a constraint involves solving the convex
program\footnote{The link function $\LinkFun$ is convex since it is
  the log-partition function of a canonical exponential family.}
\begin{align}
\label{EqnLasso}
\thetahat & \in \arg \min_{\thetapar \in \Parset} \; \Big \{
\underbrace{ \frac{1}{\numobs} \sum_{i=1}^\numobs \big \{
  \LinkFun(\inprod{\theta}{x_i}) - y_i \, \inprod{\theta}{x_i} \big
  \}}_{\LossN(\thetapar; \Zdata)}\Big\} \quad \mbox{such that
  $\|\thetapar\|_1 \leq \radcon$,}
\end{align}
with $x_i \in \real^\usedim$ as its $i^{th}$ row.  We refer to this
estimator as the log-linear Lasso; it is a special case of the
$M$-estimator~\eqref{EqnMestCon}, with the loss function
$\LossN(\thetapar; \Zdata) = \frac{1}{\numobs} \sum_{i=1}^\numobs \big
\{ \LinkFun(\inprod{\theta}{x_i}) - y_i \, \inprod{\theta}{x_i} \big
\}$ and the regularizer $\Reg{\thetapar} = \|\thetapar\|_1 =
\sum_{j=1}^\pdim |\thetapar_j|$.

Ordinary linear regression is the special case of the log-linear
setting with $\LinkFun(t) = t^2/2$ and $\Parset = \real^\pdim$, and in
this case, the estimator~\eqref{EqnLasso} corresponds to ordinary
least-squares version of Lasso~\cite{Chen98,Tibshirani96}. Other forms
of log-linear Lasso that are of interest include logistic regression,
Poisson regression, and multinomial regression.

\paragraph{Projected gradient updates:}  Computing the gradient of 
the log-linear loss from equation~\eqref{EqnLasso} is straightforward:
we have \mbox{$\nabla \LossN(\theta) = \frac{1}{\numobs}
  \sum_{i=1}^{\numobs} x_i \big \{ \LinkFun'(\inprod{\theta}{x_i}) -
  y_i \big \}$,} and the update~\eqref{EqnProjVersion} corresponds to
the Euclidean projection of the vector $\iter{t} - \frac{1}{\Lsmooth}
\nabla \LossN(\iter{t})$ onto the $\ell_1$-ball of radius $\radcon$.
It is well-known that this projection can be characterized in terms of
soft-thresholding, and that the projected
update~\eqref{EqnProjVersion} can be computed easily.  We refer the
reader to Duchi et al.~\cite{DuchiSSSiCh08} for an efficient
implementation requiring $\order(\pdim)$ operations.

\paragraph{Composite gradient updates:} The composite gradient update
for this problem amounts to solving
\begin{align*}
  \iter{t+1} = \arg\min_{\|\theta\|_1 \leq
    \radlag}\left\{\inprod{\theta}{\nabla\LossN(\theta)} +
  \frac{\Lsmooth}{2}\|\theta - \iter{t}\|_2^2 + \regpar\|\theta\|_1
  \right\}. 
\end{align*}
The update can be computed by two soft-thresholding operations. The
first step is soft thresolding the vector $\iter{t} -
\frac{1}{\Lsmooth} \nabla \LossN(\iter{t})$ at a level $\regpar$. If
the resulting vector has $\ell_1$-norm greater than $\radlag$, then we
project on to the $\ell_1$-ball just like before. Overall, the
complexity of the update is still $\order(\pdim)$ as before.

\paragraph{Decomposability of $\ell_1$-norm:}
We now illustrate how the $\ell_1$-norm is decomposable with respect
to appropriately chosen subspaces.  For any subset \mbox{$\Sset
  \subseteq \{1, 2, \ldots, \pdim \}$}, consider the subspace
\begin{align} 
\label{EqnModelSparse}
\ModelSet(\Sset) & \defn \big \{\TmpVarA \in \real^\pdim \, \mid \,
\TmpVarA_j = 0 \quad \mbox{for all $j \notin S$} \},
\end{align}
corresponding to all vectors supported only on $S$.  Defining
$\Bset(\Sset) = \ModelSet(\Sset)$, its orthogonal complement (with
respect to the usual Euclidean inner product) is given by
\begin{align}
\label{EqnBsetSparse}
\BsetPerp(\Sset) = \ModelSet^\perp(\Sset) & = \big \{ \TmpVarB \in
\real^\pdim \, \mid \, \TmpVarB_j = 0 \quad \mbox{for all $j \in S$}
\big \}.
\end{align}
To establish the decomposability of the $\ell_1$-norm with respect to
the pair $(\ModelSet(\Sset), \BsetPerp(\Sset))$, note that any
$\TmpVarA \in \ModelSet(\Sset)$ can be written in the partitioned form
$\TmpVarA = (\TmpVarA_S, 0_{\Sbar})$, where $\TmpVarA_S \in
\real^{\kdim}$ and $0_{\Sbar} \in \real^{\pdim - \kdim}$ is a vector
of zeros.  Similarly, any vector $\TmpVarB \in \BsetPerp(\Sset)$ has
the partitioned representation $(0_\Sset, \TmpVarB_\Sbar)$.  With
these representations, we have the decomposition 
\begin{align*}
\|\TmpVarA + \TmpVarB \|_1 & = \| (\TmpVarA_\Sset, 0) + (0,
\TmpVarB_\Sbar) \|_1 \; = \; \| \TmpVarA \|_1 + \| \TmpVarB \|_1.
\end{align*}
Consequently, for any subset $\Sset$, the $\ell_1$-norm is
decomposable with respect to the pairs $(\ModelSet(\Sset),
\ModelSetPerp(\Sset))$.

In analogy to the $\ell_1$-norm, various types of group-sparse norms
are also decomposable with respect to non-trivial subspace pairs.  We
refer the reader to the paper~\cite{NegRavWaiYu09} for further
discussion and examples of such decomposable norms.

\paragraph{RSC/RSM conditions:} 

A calculation using the mean-value theorem shows that for the loss
function~\eqref{EqnLasso}, the error in the first-order Taylor series,
as previously defined in equation~\eqref{EqnTayError}, can be written
as
\begin{align*}
  \Tay(\thetapar;\thetaparodd) & = \frac{1}{\numobs}
  \sum_{i=1}^\numobs \LinkFun'' \big(\inprod{\thetacon}{x_i} \big) \;
  \big(\inprod{x_i}{\thetapar - \thetaparodd} \big)^2
\end{align*}
where $\thetacon = t \thetapar + (1-t) \thetaparodd$ for some $t \in
[0,1]$.  When $\numobs < \pdim$, then we can always find pairs
$\thetapar \neq \thetaparodd$ such that $\inprod{x_i}{\thetapar -
  \thetaparodd} = 0$ for all $i = 1, 2, \ldots, \numobs$, showing that
the objective function can never be strongly convex.  On the other
hand, restricted strong convexity for log-linear models requires only
that there exist positive numbers $(\lossrsc, \FUNLOW(\LossN))$ such
that
\begin{align}
\label{EqnLogLinearRSC}
\frac{1}{\numobs} \sum_{i=1}^\numobs \LinkFun''
\big(\inprod{\thetacon}{x_i} \big) \; \big(\inprod{x_i}{\thetapar -
  \thetaparodd} \big)^2 & \geq \frac{\lossrsc}{2} \,
\plainnorm{\thetapar - \thetaparodd}^2 - \FUNLOW(\LossN) \;
\RegSq{\thetapar - \thetaparodd} \qquad \mbox{for all $\thetapar,
  \thetaparodd \in \PARNEW$},
\end{align}
where $\PARNEW \defn \Omega \cap \Ball_2(R)$ is the intersection of
the parameter space $\Omega$ with a Euclidean ball of some fixed
radius $R$ around zero.  This restriction is essential because for
many generalized linear models, the Hessian function $\LinkFun''$
approaches zero as its argument diverges.  For instance, for the
logistic function $\LinkFun(t) = \log(1 + \exp(t))$, we have
$\LinkFun''(t) = \exp(t)/[1 + \exp(t)]^2$, which tends to zero as $t
\rightarrow +\infty$.  Restricted smoothness imposes an analogous
upper bound on the Taylor error. For a broad class of log-linear
models, such bounds hold with tolerance $\FUNLOW(\LossN)$ and
$\FUNUP(\LossN)$ of the order $\sqrt{\frac{\log \pdim}{\numobs}}$.
Further details on such results are provided in the corollaries to
follow our main theorem. A detailed discussion of RSC for exponential
families in statistical problems can be found in the
paper~\cite{NegRavWaiYu09}.

In order to ensure RSC/RSM conditions on the iterates $\iter{t}$ of
the updates~\eqref{EqnAlgProj} or~\eqref{EqnAlgReg}, we also need to
ensure that $\iter{t} \in \PARNEW$. This can be done by defining
$\LossN^{'} = \LossN + \ind_{\PARNEW}(\thetapar)$, where
$\ind_{\PARNEW}(\thetapar)$ is zero when $\thetapar \in \PARNEW$ and
$\infty$ otherwise. This is equivalent to projection on the
intersection of $\ell_1$-ball with $\PARNEW$ in the
updates~\eqref{EqnAlgProj} and~\eqref{EqnAlgReg} and can be done
efficiently with Dykstra's algorithm~\cite{Dykstra85}, for instance,
as long as the individual projections are efficient.

In the special case of linear regression, we have $\LinkFun''(t) = 1$
for all $t \in \R$, so that the lower bound~\eqref{EqnLogLinearRSC}
involves only the Gram matrix $X^T X/\numobs$.  (Here $X \in
\real^{\numobs \times \pdim}$ is the usual design matrix, with $x_i
\in \real^\pdim$ as its $i^{th}$ row.)  For linear regression and
$\ell_1$-regularization, the RSC condition is equivalent to the lower
bound
\begin{align}
\label{EqnLassoRE}
\frac{\|X (\thetapar - \thetaparodd)\|_2^2}{\numobs} & \geq
\frac{\lossrsc}{2} \|\thetapar - \thetaparodd\|_2^2 - \FUNLOW(\LossN)
\; \|\thetapar - \thetaparodd\|_1^2\qquad \mbox{for all $\thetapar,
  \thetaparodd \in \Omega$.}
\end{align}
Such a condition corresponds to a variant of the restricted eigenvalue
(RE) conditions that have been studied in the
literature~\cite{BiRiTsy08,GeerBuhl09}.  Such RE conditions are
significantly milder than the restricted isometry property; we refer
the reader to van de Geer and Buhlmann~\cite{GeerBuhl09} for an
in-depth comparison of different RE conditions.  From past work, the
condition~\eqref{EqnLassoRE} is satisfied with high probability for a
broad classes of anisotropic random design
matrices~\cite{RasWaiYu10,RudZho11}, and parts of our analysis make
use of this fact.

\subsubsection{Matrices and nuclear norm regularization}
\label{SecMatrixReg}

We now discuss a general class of matrix regression problems that
falls within our framework.  Consider the space of $\pdima \times
\pdimb$ matrices endowed with the trace inner product $\tracer{A}{B}
\defn \trace(A^T B)$.  In order to ease notation, we define $\pdim
\defn \min \{ \pdima, \pdimb \}$.  Let $\ThetaStar \in \real^{\pdima
  \times \pdimb}$ be an unknown matrix and suppose that for $i = 1, 2,
\ldots, \numobs$, we observe a scalar-matrix pair \mbox{$Z_i = (y_i,
  X_i) \in \real \times \real^{\pdima \times \pdimb}$} linked to
$\ThetaStar$ via the linear model
\begin{align}
\label{EqnMatrixObs}
y_i & = \tracer{X_i}{\ThetaStar} + w_i, \qquad \mbox{for $i = 1, 2,
  \ldots, \numobs$},
\end{align}
where $w_i$ is an additive observation noise. In many contexts, it is
natural to assume that $\ThetaStar$ is exactly low-rank, or
approximately so, meaning that it is well-approximated by a matrix of
low rank.  In such settings, a number of authors
(e.g.,~\cite{Fa02,RohTsy10,NegWai09}) have studied the $M$-estimator
\begin{align}
\label{EqnSDPMatEst}
\ThetaHat & \in \arg \min_{ \Theta \in \real^{\pdima \times \pdimb}}
\Big \{ \frac{1}{2 \numobs} \, \sum_{i=1}^\numobs \big(y_i -
\tracer{X_i}{\Theta} \big)^2 \Big \} \quad \mbox{such that
  $\nuclear{\Theta} \leq \radcon$,}
\end{align}
or the corresponding regularized version. Here the \emph{nuclear or
  trace norm} is given by \mbox{$\nuclear{\Theta} \defn \sum
  \limits_{j=1}^{\pdim} \singval_j(\Theta)$,} corresponding to the sum
of the singular values. This optimization problem is an instance of a
semidefinite program.  As discussed in more detail in
Section~\ref{SecLowRank}, there are various applications in which this
estimator and variants thereof have proven useful.

\paragraph{Form of projected gradient descent:}
For the M-estimator~\eqref{EqnSDPMatEst}, the projected gradient updates
take a very simple form---namely 
\begin{align}
\label{EqnMatrixIter}
\Iter{t+1} & = \Pi \Big( \Iter{t} - \frac{1}{\Lsmooth}
\frac{\sum_{i=1}^\numobs \big( y_i - \tracer{X_i}{\Iter{t}} \big) \,
  X_i}{\numobs} \Big ),
\end{align}
where $\Pi$ denotes Euclidean projection onto the nuclear norm ball
\mbox{$\Ball_1(\radcon) = \{ \Theta \in \real^{\pdima \times \pdimb}
  \, \mid \, \nuclear{\Theta} \leq \radcon \}$.}  This nuclear norm
projection can be obtained by first computing the singular value
decomposition (SVD), and then projecting the vector of singular values
onto the $\ell_1$-ball.  The latter step can be achieved by the fast
projection algorithms discussed earlier, and there are various methods
for fast computation of SVDs. The composite gradient update also has a
simple form, requiring at most two singular value thresholding
operations as was the case for linear regression.

\paragraph{Decomposability of nuclear norm:} We now define 
matrix subspaces for which the nuclear norm is decomposable.  Given a
target matrix $\ThetaStar$---that is, a quantity to be
estimated---consider its singular value decomposition
\mbox{$\ThetaStar = U D V^T$,} where the matrix $D \in \real^{\mdim
  \times \mdim}$ is diagonal, with the ordered singular values of
$\ThetaStar$ along its diagonal, and $\mdim \defn \min\{\mdima,
\mdimb\}$.  For an integer $\rdim \in \{1, 2, \ldots, \mdim \}$, let
$U^\rdim \in \real^{\mdim \times \rdim}$ denote the matrix formed by
the top $\rdim$ left singular vectors of $\ThetaStar$ in its columns,
and we define the matrix $V^\rdim$ in a similar fashion.  Using $\col$
to denote the column span of a matrix, we then define the
subspaces\footnote{ Note that the model space $\ModelSet(U^\rdim,
  V^\rdim)$ is \emph{not equal} to $\Bset(U^\rdim, V^\rdim)$.
  Nonetheless, as required by Definition~\ref{DefnDecomp}, we do have
  the inclusion \mbox{$\ModelSet(U^\rdim, V^\rdim) \subseteq
    \Bset(U^\rdim, V^\rdim)$.}}
\begin{subequations}
\begin{align}
\ModelSet(U^\rdim, V^\rdim) & \defn \big \{ \Theta \in \real^{\mdima
  \times \mdimb} \, \mid \, \col(\Theta^T) \subseteq \col(V^r), \: \:
  \col(\Theta) \subseteq \col(U^r) \big \}, \quad \mbox{and} \\
\BsetPerp(U^\rdim, V^\rdim) & \defn \big \{ \Theta \in \real^{\mdima
  \times \mdimb} \, \mid \, \col(\Theta^T) \subseteq
  (\col(V^\rdim))^\perp, \; \: \col(\Theta) \subseteq
  (\col(U^\rdim))^\perp \big \}.
  \end{align}
\end{subequations}
Finally, let us verify the decomposability of the nuclear norm .  By
construction, any pair of matrices $\Theta \in \ModelSet(U^\rdim,
V^\rdim)$ and $\Gamma \in \BsetPerp(U^\rdim , V^\rdim)$ have
orthogonal row and column spaces, which implies the required
decomposability condition---namely $\matsnorm{\Theta +\Gamma}{1} =
\matsnorm{\Theta}{1} + \matsnorm{\Gamma}{1}$.

In some special cases such as matrix completion or matrix
decomposition that we describe in the sequel, $\PARNEW$ will involve
an additional bound on the entries of $\ThetaStar$ as well as the
iterates $\Iter{t}$ to establish RSC/RSM conditions. This can be done
by augmenting the loss with an indicator of the constraint and using
cyclic projections for computing the updates as mentioned earlier in
Example~\ref{SecLogLinear}.


\section{Main results and some consequences}
\label{SecMain}

We are now equipped to state the two main results of our paper, and
discuss some of their consequences.  We illustrate its application to
several statistical models, including sparse regression
(Section~\ref{SecSparseVec}), matrix estimation with rank constraints
(Section~\ref{SecLowRank}), and matrix decomposition problems
(Section~\ref{SecMatDecomp}).


\subsection{Geometric convergence}

Recall that the projected gradient algorithm~\eqref{EqnAlgProj} is
well-suited to solving an $M$-estimation problem in its constrained
form, whereas the composite gradient algorithm~\eqref{EqnAlgReg} is
appropriate for a regularized problem.  Accordingly, let $\thetahat$
be any optimal solution to the constrained problem~\eqref{EqnMestCon},
or the regularized problem~\eqref{EqnMestReg}, and let
$\{\iter{t}\}_{t=0}^\infty$ be a sequence of iterates generated by
generated by the projected gradient updates~\eqref{EqnAlgProj}, or the
the composite gradient updates~\eqref{EqnAlgReg}, respectively.  Of
primary interest to us in this paper are bounds on the
\emph{optimization error}, which can be measured either in terms of
the error vector $\DelHatIt{t} \defn \iter{t} - \thetahat$, or the
difference between the cost of $\iter{t}$ and the optimal cost defined
by $\thetahat$.  In this section, we state two main results
----Theorems~\ref{ThmMainCon} and~\ref{ThmMainLag}---corresponding to
the constrained and regularized cases respectively.  In addition to
the optimization error previously discussed, both of these results
involve the \emph{statistical error} $\DeltaStar \defn \thetahat -
\thetastar$ between the optimum $\thetahat$ and the nominal parameter
$\thetastar$.  At a high level, these results guarantee that under the
RSC/RSM conditions, the optimization error shrinks geometrically, with
a contraction coefficient that depends on the the loss function
$\LossN$ via the parameters $(\lossrsc, \FUNLOW(\LossN))$ and
$(\Lsmooth, \FUNUP(\LossN))$.  An interesting feature is that the
contraction occurs only up to a certain tolerance $\epsilon^2$
depending on these same parameters, and the statistical error.
However, as we discuss, for many statistical problems of interest, we
can show that this tolerance $\epsilon^2$ is of a lower order than the
intrinsic statistical error, and hence can be neglected from the
statistical point of view.  Consequently, our theory gives an explicit
upper bound on the number of iterations required to solve an
$M$-estimation problem up to the statistical precision.

\paragraph{Convergence rates for projected gradient:}

We now provide the notation necessary for a precise statement of this
claim.  Our main result actually involves a family of upper bounds on
the optimization error, one for each pair $(\ModelSet, \BsetPerp)$ of
$\Regplain$-decomposable subspaces (see
Definition~\ref{DefnDecomp}). As will be clarified in the sequel, this
subspace choice can be optimized for different models so as to obtain
the tightest possible bounds.  For a given pair $(\ModelSet,
\BsetPerp)$ such that $16 \Compat^2(\Bset) \FUNUP(\LossN) < \Lsmooth$,
let us define the \emph{contraction coefficient}
\begin{align}
\label{EqnDefnContrac}
\contrac(\LossN; \Bset) & \defn \Big \{ 1 - \frac{\lossrsc}{
  \Lsmooth} + \frac{16 \Compat^2(\Bset) \big(\FUNUP(\LossN) +
  \FUNLOW(\LossN) \big)}{\Lsmooth} \Big \} \; \Big \{ 1 - 
\frac{16
  \Compat^2(\Bset) \FUNUP(\LossN)}{\Lsmooth} \Big \}^{-1}.
\end{align}
In addition, we define the \emph{tolerance parameter}
\begin{align}
\label{EqnDefnHackStatErr}
\HACKSTATERR & \defn \frac{32 \big( \FUNUP(\LossN) + \FUNLOW(\LossN)
 \big) \; \big(2 \Reg{\BigProjModelPerp{\thetastar}} + \Compat(\Bset)
 \norm{\DeltaStar} + 2 \Reg{\DeltaStar} \big)^2 }{\Lsmooth},
\end{align}
where $\DeltaStar = \thetahat - \thetastar$ is the statistical error,
and $\BigProjModelPerp{\thetastar}$ denotes the Euclidean projection
of $\thetastar$ onto the subspace $\ModelSetPerp$.\\

\noindent In terms of these two ingredients, we now state our first
main result:
\btheos
\label{ThmMainCon}
Suppose that the loss function $\LossN$ satisfies the RSC/RSM
condition with parameters $(\lossrsc, \FUNLOW(\LossN))$ and
$(\Lsmooth, \FUNUP(\LossN))$ respectively.  Let $(\ModelSet, \Bset)$
be any $\Regplain$-decomposable pair of subspaces such that $\ModelSet
\subseteq \Bset$ and $0 < \contrac \equiv \contrac(\LossN, \Bset) < 1$.
Then for any optimum $\thetahat$ of the problem~\eqref{EqnMestCon} for
which the constraint is active, we have
\begin{align}
\label{EqnMainConRate}
\plainnorm{\iter{t+1} - \thetahat}^2 & \leq \contrac^t \,
\plainnorm{\iter{0} - \thetahat}^2 + \frac{\HACKSTATERR}{1-\contrac}
\qquad \mbox{for all iterations $t = 0, 1, 2, \ldots$.}
\end{align}
\etheos

\paragraph{Remarks:}
Theorem~\ref{ThmMainCon} actually provides a family of upper bounds,
one for each $\Regplain$-decomposable pair $(\ModelSet, \Bset)$ such
that $0 < \contrac \equiv \contrac(\LossN, \Bset) < 1$.  This
condition is always satisfied by setting $\Bset$ equal to the trivial
subspace $\{0\}$: indeed, by definition~\eqref{EqnDefnSubspaceCompat}
of the subspace compatibility, we have $\Compat(\Bset) = 0$, and hence
$\contrac(\LossN; \{0\}) = \big(1 - \frac{\lossrsc}{\Lsmooth} \big)
< 1$.  Although this choice of $\Bset$ minimizes the contraction
coefficient, it will lead\footnote{Indeed, the setting $\ModelPerp =
  \R^\pdim$ means that the term $\Reg{\BigProjModelPerp{\thetastar}} =
  \Reg{\thetastar}$ appears in the tolerance; this quantity is far
  larger than statistical precision.} to a very large tolerance
parameter $\HACKSTATERR$.
A more typical application of Theorem~\ref{ThmMainCon} involves
non-trivial choices of the subspace $\Bset$. 

\begin{figure}[h]
\begin{center}
\begin{tabular}{ccc}
\psfrag{#t0#}{$\DelHatIt{0}$}
\psfrag{#t1#}{$\DelHatIt{1}$}
\psfrag{#tt#}{$\DelHatIt{t}$}
\psfrag{#that#}{$0$}
\psfrag{#e#}{$\epsilon$}
\widgraph{.42\textwidth}{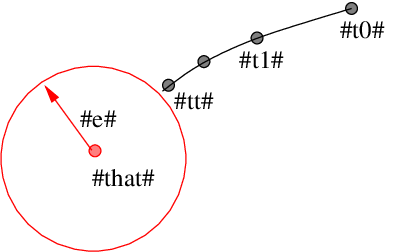} & &
\psfrag{#t0#}{$\DelHatIt{0}$}
\psfrag{#t1#}{$\DelHatIt{1}$}
\psfrag{#tt#}{$\DelHatIt{t}$}
\psfrag{#that#}{$0$}
\psfrag{#e#}{$\epsilon$}
\psfrag{#se#}{$\plainnorm{\DeltaStar}$}
\psfrag{#tstar#}{$\DeltaStar$}
\widgraph{.34\textwidth}{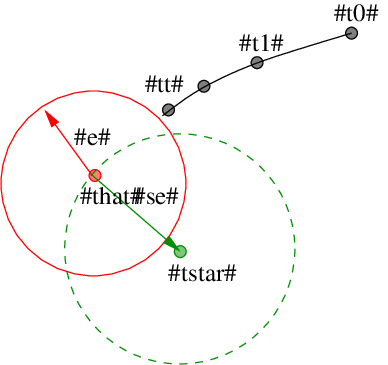} \\
(a) & & (b)
\end{tabular}
\caption{(a) Generic illustration of Theorem~\ref{ThmMainCon}.  The
  optimization error \mbox{$\DelHatIt{t} = \iter{t} - \thetahat$} is
  guaranteed to decrease geometrically with coefficient $\contrac \in
  (0,1)$, up to the tolerance $\epsilon^2 = \HACKSTATERR$, represented
  by the circle.  (b) Relation between the optimization tolerance
  $\HACKSTATERR$ (solid circle) and the statistical precision
  $\plainnorm{\DeltaStar} = \plainnorm{\thetastar - \thetahat}$
  (dotted circle). In many settings, we have $\HACKSTATERR \ll
  \plainnorm{\DeltaStar}^2$, so that convergence is guaranteed up to a
  tolerance lower than statistical precision.}
\label{FigThmPicture}
\end{center}
\end{figure}

The bound~\eqref{EqnMainConRate} guarantees that the optimization
error decreases geometrically, with contraction factor $\contrac \in
(0,1)$, up to a certain tolerance proportional to $\HACKSTATERR$, as
illustrated in Figure~\ref{FigThmPicture}(a).  The contraction factor
$\contrac$ approaches the $1-\lossrsc/\Lsmooth$ as the number of
samples grows. The appearance of the ratio $\lossrsc/\Lsmooth$ is
natural since it measures the conditioning of the objective function;
more specifically, it is essentially a restricted condition number of
the Hessian matrix.  On the other hand, the tolerance parameter
$\epsilon$ depends on the choice of decomposable subspaces, the
parameters of the RSC/RSM conditions, and the statistical error
$\DeltaStar = \thetahat - \thetastar$ (see
equation~\eqref{EqnDefnHackStatErr}).  In the corollaries of
Theorem~\ref{ThmMainCon} to follow, we show that the subspaces can
often be chosen such that $\HACKSTATERR = o(\plainnorm{\thetahat -
  \thetastar}^2)$.  Consequently, the bound~\eqref{EqnMainConRate}
guarantees geometric convergence up to a tolerance \emph{smaller than
  statistical precision}, as illustrated in
Figure~\ref{FigThmPicture}(b). This is sensible, since in statistical
settings, there is no point to optimizing beyond the statistical
precision.\\

%

The result of Theorem~\ref{ThmMainCon} takes a simpler form when there
is a subspace $\ModelSet$ that includes $\thetastar$, and the
$\Regplain$-ball radius is chosen such that $\radcon \leq
\Reg{\thetastar}$.  In this case, by appropriately controlling the
error term, we can establish that it is of lower order than the
statistical precision ---namely, the squared difference
$\norm{\thetahat - \thetastar}^2$ between an optimal solution
$\thetahat$ to the convex program~\eqref{EqnMestCon}, and the unknown
parameter $\thetastar$.

\bcors
\label{CorGeneric}
In addition to the conditions of Theorem~\ref{ThmMainCon}, suppose
that $\thetastar \in \ModelSet$ and \mbox{$\radcon \leq
  \Reg{\thetastar}$.}  Then as long as $\Compat^2(\Bset)
\big(\FUNUP(\LossN) + \FUNLOW(\LossN) \big) = o(1)$, we have
\begin{align}
\plainnorm{\iter{t+1} - \thetahat}^2 & \leq \contrac^t \,
\plainnorm{\iter{0} - \thetahat}^2 + o \big(\plainnorm{\thetahat -
  \thetastar}^2 \big) \qquad \mbox{for all iterations $t = 0, 1, 2,
  \ldots$.}
\end{align}
\ecors
\noindent Thus, Corollary~\ref{CorGeneric} guarantees that the
optimization error decreases geometrically, with contraction factor
$\contrac$, up to a tolerance that is of strictly lower order than the
statistical precision $\plainnorm{\thetahat - \thetastar}^2$.  As will
be clarified in several examples to follow, the condition
$\Compat^2(\Bset) \big(\FUNUP(\LossN) + \FUNLOW(\LossN) \big) = o(1)$
is satisfied for many statistical models, including sparse linear
regression and low-rank matrix regression. This result is illustrated
in Figure~\ref{FigThmPicture}(b), where the solid circle represents
the optimization tolerance, and the dotted circle represents the
statistical precision.  In the results to follow, we will quantify the
term $o \big(\plainnorm{\thetahat - \thetastar}^2 \big)$ in a more
precise manner for different statistical models. \\

\paragraph{Convergence rates for composite gradient:}
We now present our main result for the composite gradient
iterates~\eqref{EqnAlgReg} that are suitable for the Lagrangian-based
estimator~\eqref{EqnMestReg}.  As before, our analysis yields a range
of bounds indexed by subspace pairs $(\ModelSet, \BsetPerp)$ that are
$\Regplain$-decomposable.  For any subspace $\Bset$ such that $64
\FUNLOW(\LossN) \Compat^2(\Bset) < \lossrsc$, we define
\emph{effective RSC coefficient} as
\begin{align}
\label{EqnDefnUpdateRSC}
\uplossrsc & \defn \lossrsc - 64 \FUNLOW(\LossN) \Compat^2(\Bset).
\end{align}
This coefficient accounts for the residual amount of strong convexity
after accounting for the lower tolerance terms.  In addition, we
define the \emph{compound contraction coefficient} as
\begin{align}
\label{EqnDefnContracLag}
\contrac(\LossN; \Bset) & \defn \left \{ 1 - \frac{\uplossrsc}{ 4
  \Lsmooth} + \frac{64 \Compat^2(\Bset) \FUNUP(\LossN)}{\uplossrsc}
\right \} \; \slopa(\Bset)
\end{align}
where $\slopa(\Bset) \; \defn \; \big(1-\frac{64 \FUNUP(\LossN)
  \Compat^2(\BsetSmall)}{\uplossrsc} \big)^{-1}$, and $\DeltaStar =
\betahatlag - \thetastar$ is the statistical error
vector\footnote{When the context is clear, we remind the reader that
  we drop the subscript $\regpar$ on the parameter $\thetahat$.}  for
a specific choice of $\radlag$ and $\regpar$.  As before, the
coefficient $\contrac$ measures the geometric rate of convergence for
the algorithm.  Finally, we define the \emph{compound tolerance
  parameter}
\begin{align}
\label{EqnDefnHackStatErrLag}
\HACKSTATERR & \defn 8 \, \slopa(\Bset) \, \slopb(\Bset) \left ( 6
\Compat(\Bset) \norm{\DeltaStar} + 8
\Reg{\BigProjModelPerp{\thetastar}} \right )^2,
\end{align}
where $\slopb(\Bset) \defn 2 \left (\frac{\uplossrsc}{4 \Lsmooth} +
\frac{128 \FUNUP(\LossN) \Compat^2(\BsetSmall)}{\uplossrsc} \right )
\FUNLOW(\LossN) + 8 \FUNUP(\LossN) + 2 \FUNLOWALL$.  As with our previous result, the
tolerance parameter determines the radius up to which geometric
convergence can be attained.

Recall that the regularized problem~\eqref{EqnMestReg} involves both a
regularization weight $\regpar$, and a constraint radius $\radlag$.
Our theory requires that the constraint radius is chosen such that
$\radlag \geq \Reg{\thetastar}$, which ensures that $\thetastar$ is
feasible.  In addition, the regularization parameter should be chosen
to satisfy the constraint
\begin{align}
\label{proplagbounds}
\regpar & \geq 2 \Regdual{\grad \LossN(\betastar)},
\end{align}
where $\Plainregdual$ is the dual norm of the regularizer.  This
constraint is known to play an important role in proving bounds on the
statistical error of regularized $M$-estimators (see the
paper~\cite{NegRavWaiYu09} and references therein for further
details).  Recalling the definition~\eqref{EqnMestReg} of the overall
objective function $\composite_\numobs(\thetapar)$, the following
result provides bounds on the \emph{excess loss}
$\composite_\numobs(\iter{t}) - \composite_\numobs(\thetahatlag)$.
\btheos
\label{ThmMainLag}
Consider the optimization problem~\eqref{EqnMestReg} for a radius
$\radlag$ such that $\thetastar$ is feasible, and a regularization
parameter $\regpar$ satisfying the bound~\eqref{proplagbounds}, and
suppose that the loss function $\LossN$ satisfies the RSC/RSM
condition with parameters $(\lossrsc, \FUNLOW(\LossN))$ and
$(\Lsmooth, \FUNUP(\LossN))$ respectively.  Let $(\ModelSet,
\BsetPerp)$ be any $\Regplain$-decomposable pair such that
\begin{align}
\label{EqnFranceAnnoy}
\contrac \; \equiv \; \contrac(\LossN, \Bset) \in [0,1), \quad
  \mbox{and} \quad \frac{32 \, \radlag }{1-\contrac(\LossN;\Bset)}
  \slopa(\Bset) \slopb(\Bset) \; \leq \; \regpar.
\end{align}
Then for any tolerance parameter \mbox{$\delta^2 \geq
  \frac{\HACKSTATERR}{(1-\contrac)}$,} we have
\begin{align}
\label{EqnMainConRateLag}
\composite_\numobs(\iter{t}) - \composite_\numobs(\thetahatlag) \leq
\delta^2 \quad \mbox{for all} \quad t \; \geq \;
\frac{2\log\frac{\composite_\numobs(\iter{0}) -
    \composite_\numobs(\thetahatlag)}{\delta^2}}{\log(1/\contrac)}
+
\log_2\log_2\bigg(\frac{\radlag\regpar}{\delta^2}\bigg)
\bigg(1+\frac{\log 2}{\log(1/\contrac)}\bigg).
\end{align}
\etheos

\paragraph{Remarks:}
Note that the bound~\eqref{EqnMainConRateLag} guarantees the excess
loss \mbox{$\composite_\numobs(\iter{t}) -
  \composite_\numobs(\thetahat)$} decays geometrically up to any
squared error $\delta^2$ larger than the compound
tolerance~\eqref{EqnDefnHackStatErrLag}.  Moreover, the RSC condition
also allows us to translate this bound on objective values to a bound
on the optimization error $\iter{t} - \thetahat$.  In particular, for
any iterate $\iter{t}$ such that $\composite_\numobs(\iter{t}) -
\composite_\numobs(\thetahat) \leq \delta^2$, we are guaranteed that
\begin{equation}
  \label{EqnParErrorLag}
  \norm{\iter{t} - \thetahatlag}^2 \leq \frac{2 \delta^2}{\uplossrsc}
  + \frac{16 \delta^2 \FUNLOW(\LossN)}{\uplossrsc\regpar^2} + \frac{4
    \FUNLOW(\LossN) (6\Compat(\Bset) +
    8\Reg{\BigProjModelPerp{\thetastar}})^2}{\uplossrsc}.
\end{equation}
In conjunction with Theorem~\ref{ThmMainLag}, we see that it suffices
to take a number of steps that is logarithmic in the inverse tolerance
$(1/\delta)$, again showing a geometric rate of convergence.

Whereas Theorem~\ref{ThmMainCon} requires setting the radius so that
the constraint is active, Theorem~\ref{ThmMainLag} has only a very
mild constraint on the radius $\radlag$, namely that it be large
enough such that $\radlag \geq \Reg{\thetastar}$.  The reason for this
much milder requirement is that the additive regularization with
weight $\regpar$ suffices to constrain the solution, whereas the extra
side constraint is only needed to ensure good behavior of the
optimization algorithm in the first few iterations.  The
regularization parameter $\regpar$ must satisfy the so-called dual
norm condition~\eqref{proplagbounds}, which has appeared in past
literature on statistical estimation, and is well-characterized for a
broad range of statistical models (e.g., see the
paper~\cite{NegRavWaiYu09} and references therein). \\

\paragraph{Step-size setting:} It seems that the
updates~\eqref{EqnAlgProj} and~\eqref{EqnAlgReg} need to know the
smoothness bound $\Lsmooth$ in order to set the step-size for gradient
updates. However, we can use the same doubling trick as described in
Algorithm (3.1) of Nesterov~\cite{Nesterov07}. At each step, we check
if the smoothness upper bound holds at the current iterate relative to
the previous one. If the condition does not hold, we double our
estimate of $\Lsmooth$ and resume. This guarantees a geometric
convergence with a contraction factor worse at most by a factor of 2,
compared to the knowledge of $\Lsmooth$. We refer the reader to
Nesterov~\cite{Nesterov07} for details. 

The following subsections are devoted to the development of some
consequences of Theorems~\ref{ThmMainCon} and~\ref{ThmMainLag} and
Corollary~\ref{CorGeneric} for some specific statistical models, among
them sparse linear regression with $\ell_1$-regularization, and matrix
regression with nuclear norm regularization.  In contrast to the
entirely deterministic arguments that underlie the
Theorems~\ref{ThmMainCon} and~\ref{ThmMainLag}, these corollaries
involve probabilistic arguments, more specifically in order to
establish that the RSC and RSM properties hold with high probability.


\subsection{Sparse vector regression} 
\label{SecSparseVec}

Recall from Section~\ref{SecLogLinear} the observation model for sparse
linear regression.  In a variety of applications,
it is natural to assume that $\betaopt$ is sparse. For a parameter
$\qpar \in [0,1]$ and radius $\radq > 0$, let us define the
$\ell_\qpar$ ``ball''
\begin{align}
\Ball_\qpar(\radq) & \defn \big \{ \thetapar \in \real^\pdim \, \mid \,
\sum_{j=1}^\pdim |\beta_j|^\qpar \leq \radq \big \}.
\end{align}
Note that $\qpar = 0$ corresponds to the case of ``hard sparsity'',
for which any vector $\beta \in \Ball_0(R_0)$ is supported on a set of
cardinality at most $R_0$.  For $\qpar \in (0,1]$, membership in the
set $\Ball_\qpar(\radq)$ enforces a decay rate on the ordered
coefficients, thereby modelling approximate sparsity.  In order to
estimate the unknown regression vector $\thetastar \in
\Ball_\qpar(R_\qpar)$, we consider the least-squares Lasso estimator
from Section~\ref{SecLogLinear}, based on the quadratic loss function
$\loss(\thetapar; Z_1^\numobs) \defn \frac{1}{2\numobs} \|y - \Xmat
\thetapar\|_2^2$, where $\Xmat \in \real^{\numobs \times \pdim}$ is
the design matrix.  In order to state a concrete result, we consider a
random design matrix $\Xmat$, in which each row $x_i \in \real^\pdim$
is drawn i.i.d. from a $N(0, \CovMat)$ distribution, where $\CovMat$
is a positive definite covariance matrix.  We refer to this as the
\emph{$\CovMat$-ensemble of random design matrices}, and use
$\lammax(\CovMat)$ and $\lammin(\CovMat)$ to refer the maximum and
minimum eigenvalues of $\CovMat$ respectively, and $\rhovec(\CovMat)
\defn \max \limits_{j=1, 2, \ldots, \pdim} \CovMat_{jj}$ for the
maximum variance.  We also assume that the observation noise is
zero-mean and sub-Gaussian with parameter $\noisevar^2$.

\paragraph{Guarantees for constrained Lasso:}

Our convergence rate on the optimization error $\iter{t} - \thetahat$
is stated in terms of the contraction coefficient
\begin{align}
\label{EqnDefnContracSparse}
\contrac & \defn \Big \{ 1 - \frac{\lammin(\CovMat)}{4
  \lammax(\CovMat)} + \phin(\CovMat) \Big \} \; \Big \{ 1 -
\phin(\CovMat) \Big \}^{-1},
\end{align}
where we have adopted the shorthand
\begin{align}
\label{EqnPhinSparse}
\phin(\CovMat) \defn \left\{\begin{array}{cc}\frac{\PLAINCON_0
\rhovec(\CovMat)}{\lammax(\CovMat)} \; \radq \: \big(\frac{\log
\pdim}{\numobs}\big)^{1-\qpar/2}&\mbox{for $q > 0$}\\ 
\frac{\PLAINCON_0
\rhovec(\CovMat)}{\lammax(\CovMat)} \; s \: \big(\frac{\log
\pdim}{\numobs}\big)&\mbox{for $q = 0$}
\end{array}\right. , \qquad \mbox{for a numerical
constant $\PLAINCON_0$,}
\end{align}
We assume that $\phin(\CovMat)$ is small enough to ensure that
$\contrac \in (0,1)$; in terms of the sample size, this amounts to a
condition of the form $\numobs = \Omega(\radq^{1/(1-\qpar/2)} \log
\pdim)$.  Such a scaling is sensible, since it is known from minimax
theory on sparse linear regression~\cite{RasWaiYu09} to be necessary
for any method to be statistically consistent over the
$\ell_\qpar$-ball.  \\

\noindent With this set-up, we have the following consequence of
Theorem~\ref{ThmMainCon}:
\bcors[Sparse vector recovery]
\label{CorSparse}
Under conditions of Theorem~\ref{ThmMainCon}, suppose that we solve
the constrained Lasso with \mbox{$\radcon \leq \|\thetastar\|_1$.}
\begin{enumerate}
\item[(a)] \emph{Exact sparsity:} If $\thetastar$ is supported on a
subset of cardinality $\kdim$, then with probability at least \mbox{$1
- \exp(-\PLAINCON_1 \log \pdim)$,} the iterates~\eqref{EqnAlgProj}
with $\Lsmooth = 2 \lammax(\CovMat)$ satisfy
\begin{align}
\label{EqnSparseBoundZero}
\|\iter{t} - \betahat\|^2_2 & \leq \contrac^t \| \iter{0} -
\betahat\|_2^2 + \PLAINCON_2 \; \phin(\CovMat) \; \|\thetahat -
\thetastar\|_2^2 \qquad \mbox{for all $t = 0, 1, 2, \ldots$}.
\end{align}
\item[(b)] \emph{Weak sparsity:} Suppose that $\thetastar \in
  \Ball_\qpar(\radq)$ for some $\qpar \in (0,1]$.  Then with
    probability at least \mbox{$1 - \exp(-\PLAINCON_1 \log \pdim)$,}
    the iterates~\eqref{EqnAlgProj} with $\Lsmooth = 2
    \lammax(\CovMat)$ satisfy
\begin{align}
\label{EqnSparseBoundWeak}
\|\iter{t} - \thetahat\|_2^2 & \leq \contrac^t \, \|\iter{0} -
\thetahat\|_2^2 + \PLAINCON_2 \; \phin(\CovMat) \; \biggr \{ \radq
\big(\frac{\log \pdim}{\numobs} \big)^{1-\qpar/2} +
\|\thetahat - \thetastar\|_2^2 \biggr \}.
\end{align}

\end{enumerate}
\ecors

We provide the proof of Corollary~\ref{CorSparse} in
Section~\ref{SecProofCorSparse}.  Here we compare part (a), which
deals with the special case of exactly sparse vectors, to some past
work that has established convergence guarantees for optimization
algorithms for sparse linear regression.  Certain methods are known to
converge at sublinear rates (e.g.,~\cite{becker09nesta}), more
specifically at the rate $\order(1/t^2)$.  The geometric rate of
convergence guaranteed by Corollary~\ref{CorSparse} is exponentially
faster.  Other work on sparse regression has provided geometric rates
of convergence that hold once the iterates are close to the
optimum~\cite{BreLor08,HalWotZha08}, or geometric convergence up to
the noise level $\noisevar^2$ using various methods, including greedy
methods~\cite{TroppGil07} and thresholded gradient
methods~\cite{GarKha09}. In contrast, Corollary~\ref{CorSparse}
guarantees geometric convergence for all iterates up to a precision
below that of statistical error.  For these problems, the statistical
error $\frac{ \noisevar^2 \spindex \log \pdim}{\numobs}$ is typically
much smaller than the noise variance $\noisevar^2$, and decreases as
the sample size is increased.

In addition, Corollary~\ref{CorSparse} also applies to the case of
approximately sparse vectors, lying within the set
$\Ball_\qpar(\radq)$ for $\qpar \in (0,1]$.  There are some important
  differences between the case of exact sparsity
  (Corollary~\ref{CorSparse}(a)) and that of approximate sparsity
  (Corollary~\ref{CorSparse}(b)).  Part (a) guarantees geometric
  convergence to a tolerance depending only on the statistical error
  $\|\thetahat - \thetastar\|_2$.  In contrast, the second result also
  has the additional term $\radq \big( \frac{\log \pdim}{\numobs}
  \big)^{1-\qpar/2}$. This second term arises due to the statistical
  non-identifiability of linear regression over the $\ell_\qpar$-ball,
  and it is no larger than $\norm{\thetahat - \thetastar}_2^2$ with
  high probability.  This assertion follows from known
  results~\cite{RasWaiYu09} about minimax rates for linear regression
  over $\ell_\qpar$-balls; these unimprovable rates include a term of
  this order.

\paragraph{Guarantees for regularized Lasso:} 
Using similar methods, we can also use Theorem~\ref{ThmMainLag} to
obtain an analogous guarantee for the regularized Lasso estimator.
Here focus only on the case of exact sparsity, although the result
extends to approximate sparsity in a similar fashion.  Letting
$\plaincon_i, i = 0, 1, 2, 3, 4$ be universal positive constants, we
define the modified curvature constant \mbox{$\uplossrsc \defn
  \lossrsc - \plaincon_0 \; \frac{\spindex \, \log
    \pdim}{\numobs}\rhovec(\CovMat)$.}  Our results assume that $
\numobs = \Omega( \spindex \log \pdim)$, a condition known to be
necessary for statistical consistency, so that $\uplossrsc > 0$.  The
contraction factor then takes the form
\begin{align*}
\contrac \defn \big \{ 1 - \frac{\lammin(\CovMat)}{16
  \lammax(\CovMat)} + \plaincon_1 \phin(\CovMat) \big\} \; \big \{1 -
\plaincon_2 \phin(\CovMat) \big \}^{-1}, \quad \mbox{where} \quad
\phin(\CovMat) = \frac{\rhovec(\CovMat)}{\uplossrsc} \; \frac{\spindex
  \, \log \pdim}{\numobs}.
\end{align*}
The tolerance factor in the optimization is given by 
\begin{align}
\label{SparseMinError}
\CRIT^2 & \defn \frac{5 + \plaincon_2 \phin(\CovMat)} {1-\plaincon_3
  \phin(\CovMat)} \; \frac{\rhovec(\CovMat) \; \spindex \log
  \pdim}{\numobs} \|\thetastar - \thetahat\|_2^2,
\end{align}
where $\thetastar \in \real^\pdim$ is the unknown regression vector,
and $\thetahat$ is any optimal solution. With this notation, we have
the following corollary.
\bcors[Regularized Lasso]
\label{CorSparseLag}
Under conditions of Theorem~\ref{ThmMainLag}, suppose that we solve
the regularized Lasso with $\regpar = 6 \sqrt{\frac{ \noisevar \log
    \pdim}{\numobs}}$, and that $\thetastar$ is supported on a subset
of cardinality at most $\spindex$. Suppose that we have the condition
\begin{align}
  64 \radlag \frac{\log \pdim}{\numobs}\; \frac{5 +
    \frac{\uplossrsc}{4\Lsmooth} + \frac{64 \spindex \log\pdim
      /\numobs} {\uplossrsc}} {\frac{\uplossrsc}{4\Lsmooth} -
    \frac{128 \spindex \log\pdim /\numobs} {\uplossrsc}} \leq
  \regpar. 
  \label{eqn:condition-regsparse}
\end{align}
Then with probability at least
\mbox{$1 - \exp(-\PLAINCON_4 \log \pdim)$,} for any $\delta^2 \geq
\CRIT^2$, for any optimum $\thetahatlag$, we have
\begin{align*}
  \|\iter{t} - \thetahatlag\|_2^2 & \leq \delta^2 \qquad \mbox{for all
    iterations $t \; \geq \;
    \big(\log\frac{\composite_\numobs(\iter{0}) -
      \composite_\numobs(\thetahatlag)}{\delta^2}\big) /
    \big(\log\frac{1}{\contrac} \big)$.}
\end{align*}
\ecors
\noindent As with Corollary~\ref{CorSparse}(a), this result guarantees
that $\order(\log(1/\CRIT^2))$ iterations are sufficient to obtain an
iterate $\iter{t}$ that is within squared error $\order(\CRIT^2)$ of
any optimum $\thetahatlag$. The
condition~\eqref{eqn:condition-regsparse} is the specialization of
Equation~\ref{EqnFranceAnnoy} to the sparse linear regression problem,
and imposes an upper bound on admissible settings of $\radlag$ for our
theory. Moreover, whenever $\frac{\spindex \log \pdim}{\numobs} =
o(1)$---a condition that is required for statistical consistency of
\emph{any method}---the optimization tolerance $\CRIT^2$ is of lower
order than the statistical error $\|\thetastar - \theta\|_2^2$.

\subsection{Matrix regression with rank constraints}
\label{SecLowRank}

We now turn to estimation of matrices under various types of ``soft''
rank constraints.  Recall the model of matrix regression from
Section~\ref{SecMatrixReg}, and the $M$-estimator based on
least-squares regularized with the nuclear norm~\eqref{EqnSDPMatEst}.
So as to reduce notational overhead, here we specialize to square
matrices $\ThetaStar \in \real^{\mdim \times \mdim}$, so that our
observations are of the form
\begin{align}
\label{EqnMatrixTwo}
y_i & = \tracer{X_i}{\ThetaStar} + w_i, \quad \mbox{for $i = 1, 2,
  \ldots, \numobs$,}
\end{align}
where $X_i \in \real^{\mdim \times \mdim}$ is a matrix of covariates,
and $w_i \sim N(0, \noisevar^2)$ is Gaussian noise.  As discussed in
Section~\ref{SecMatrixReg}, the nuclear norm $\Reg{\Theta} =
\nuclear{\Theta} = \sum_{j=1}^\mdim \singval_j(\Theta)$ is
decomposable with respect to appropriately chosen matrix subspaces,
and we exploit this fact heavily in our analysis.

We model the behavior of both exactly and approximately low-rank
matrices by enforcing a sparsity condition on the vector
$\singval(\Theta) = \begin{bmatrix} \singval_1(\Theta) &
  \singval_2(\Theta) & \cdots & \singval_\pdim(\Theta) \end{bmatrix}$
of singular values.  In particular, for a parameter $\qpar \in [0,1]$,
we define the $\ell_\qpar$-``ball'' of matrices
\begin{align}
\label{EqnMatrixEllq}
\Ball_\qpar(\radq) \defn \big \{ \Theta \in \real^{\mdim \times \mdim}
\, \mid \, \sum_{j=1}^\mdim |\singval_j(\Theta)|^\qpar \leq \radq \big
\}.
\end{align}
Note that if $\qpar = 0$, then $\Ball_0(R_0)$ consists of the set of
all matrices with rank at most $\rdim = R_0$.  On the other hand, for
$\qpar \in (0,1]$, the set $\Ball_\qpar(\radq)$ contains matrices of
  all ranks, but enforces a relatively fast rate of decay on the
  singular values.

\subsubsection{Bounds for matrix compressed sensing}
\label{SecMatCompress}

We begin by considering the compressed sensing version of matrix
regression, a model first introduced by Recht et
al.~\cite{RecFazPar10}, and later studied by other authors
(e.g.,~\cite{LeeBres09, NegWai09}).  In this model, the observation
matrices $X_i \in \real^{\mdim \times \mdim}$ are dense and drawn from
some random ensemble.  The simplest example is the standard Gaussian
ensemble, in which each entry of $X_i$ is drawn i.i.d. as standard
normal $N(0,1)$.  Note that $X_i$ is a dense matrix in general; this
in an important contrast with the matrix completion setting to follow
shortly.

Here we consider a more general ensemble of random matrices $X_i$, in
which each matrix \mbox{$X_i \in \real^{\mdim \times \mdim}$} is drawn
i.i.d. from a zero-mean normal distribution in $\real^{\mdim^2}$ with
covariance matrix $\CovMat \in \real^{\mdim^2 \times \mdim^2}$.  The
setting $\CovMat = I_{\mdim^2 \times \mdim^2}$ recovers the standard
Gaussian ensemble studied in past work.  As usual, we let
$\lammax(\CovMat)$ and $\lammin(\CovMat)$ define the maximum and
minimum eigenvalues of $\CovMat$, and we define $\rhomat(\CovMat) =
\sup_{\|u\|_2 = 1} \sup_{\|v\|_2 = 1} \var \big( \tracer{X}{u v^T}
\big)$, corresponding to the maximal variance of $X$ when projected
onto rank one matrices.  For the identity ensemble, we have
$\rhomat(I) = 1$.

We now state a result on the convergence of the
updates~\eqref{EqnMatrixIter} when applied to a statistical problem
involving a matrix $\ThetaStar \in \Ball_\qpar(\radq)$.  The
convergence rate depends on the contraction coefficient
\begin{align*}
\contrac & \defn \Big \{ 1 - \frac{\lammin(\CovMat)}{4
  \lammax(\CovMat)} + \phin(\CovMat) \Big \} \; \Big \{ 1 -
\phin(\CovMat) \Big \}^{-1},
\end{align*}
where $\phin(\CovMat) \defn \frac{\PLAINCON_1
  \rhomat(\CovMat)}{\lammax(\CovMat)} \; \radq \big (
\frac{\mdim}{\numobs} \big)^{1 - \qpar/2}$ for some universal constant
$\PLAINCON_1$.  In the case $\qpar = 0$, corresponding to matrices
with rank at most $r$, note that we have $R_0 = r$.  With this
notation, we have the following convergence guarantee:\\

\bcors[Low-rank matrix recovery]
\label{CorLowRank}
Under conditions of Theorem~\ref{ThmMainCon}, consider the
semidefinite program~\eqref{EqnSDPMatEst} with $\radcon 
\leq \nuclear{\ThetaStar}$, and suppose that we apply the projected
gradient updates~\eqref{EqnMatrixIter} with $\Lsmooth = 2
\lammax(\CovMat)$.
\begin{enumerate}
\item[(a)] \emph{Exactly low-rank:} In the case $\qpar = 0$, if 
  $\ThetaStar$ has rank $\rdim < \pdim$, then with probability at
  least \mbox{$1 - \exp(-\PLAINCON_0 \mdim)$,} the
  iterates~\eqref{EqnMatrixIter} satisfy the bound
\begin{align}
\label{EqnMatLowRankZero}
\frob{\Iter{t} - \ThetaHat}^2 & \leq \contrac^t \frob{\Iter{0} -
  \ThetaHat}^2 + \PLAINCON_2 \; \phin(\CovMat) \; \frob{\ThetaHat -
  \ThetaStar}^2 \qquad \mbox{for all $t = 0, 1, 2, \ldots$}.
\end{align}
\item[(b)] \emph{Approximately low-rank:} If $\ThetaStar \in
  \Ball_\qpar(\radq)$ for some $\qpar \in (0,1]$, then with
    probability at least \mbox{$1 - \exp(-\PLAINCON_0 \mdim)$,} the
    iterates~\eqref{EqnMatrixIter} satisfy
\begin{align}
\label{EqnMatLowRankWeak}
\frob{\Iter{t} - \ThetaHat}^2 & \leq \contrac^t \, \frob{\Iter{0} -
  \ThetaHat}^2 + \PLAINCON_2 \phin(\CovMat) \; \biggr \{ \radq \bigg(
\frac{\mdim}{\numobs} \bigg)^{1-\qpar/2} + \frob{\ThetaHat -
  \ThetaStar}^2 \biggr \},
\end{align}
\end{enumerate}
\ecors
Although quantitative aspects of the rates are different,
Corollary~\ref{CorLowRank} is analogous to Corollary~\ref{CorSparse}.
For the case of exactly low rank matrices (part (a)), geometric
convergence is guaranteed up to a tolerance involving the statistical
error $\frob{\ThetaHat - \ThetaStar}^2$.  For the case of
approximately low rank matrices (part (b)), the tolerance term
involves an additional factor of $\radq \big( \frac{\mdim}{\numobs}
\big)^{1-\qpar/2}$.  Again, from known results on minimax rates for
matrix estimation~\cite{RohTsy10}, this term is known to be of
comparable or lower order than the quantity $\frob{\ThetaHat -
  \ThetaStar}^2$.  As before, it is also possible to derive an
analogous corollary of Theorem~\ref{ThmMainLag} for estimating
low-rank matrices; in the interests of space, we leave such a
development to the reader.


\subsubsection{Bounds for matrix completion}
\label{SecMatComplete}

In this model, observation $y_i$ is a noisy version of a randomly
selected entry $\ThetaStar_{a(i), b(i)}$ of the unknown matrix
$\ThetaStar$.  Applications of this matrix completion problem include
collaborative filtering~\cite{SreAloJaa05}, where the rows of the
matrix $\ThetaStar$ correspond to users, and the columns correspond to
items (e.g., movies in the Netflix database), and the entry
$\ThetaStar_{ab}$ corresponds to user's $a$ rating of item $b$.  Given
observations of only a subset of the entries of $\ThetaStar$, the goal
is to fill in, or complete the matrix, thereby making recommendations
of movies that a given user has not yet seen.

Matrix completion can be viewed as a particular case of the matrix
regression model~\eqref{EqnMatrixObs}, in particular by setting $X_i =
E_{a(i) b(i)}$, corresponding to the matrix with a single one in
position $(a(i), b(i))$, and zeroes in all other positions.  Note that
these observation matrices are extremely sparse, in contrast to the
compressed sensing model.  Nuclear-norm based estimators for matrix
completion are known to have good statistical properties
(e.g.,~\cite{CanRec08,Recht09,SreAloJaa05,NegWai10b}).  Here we
consider the $M$-estimator
\begin{align}
\label{EqnMestNetflix}
\ThetaHat & \in \arg \min_{ \Theta \in \Parset} \; \frac{1}{2 \numobs}
\sum_{i=1}^\numobs \big(y_i - \Theta_{a(i) b(i)} \big)^2 \quad
\mbox{such that $\nuclear{\Theta} \leq \radcon$},
\end{align}
where $\Parset = \{ \Theta \in \real^{\mdim \times \mdim} \, \mid \,
\|\Theta\|_\infty \leq \frac{\ANNOY}{\mdim} \}$ is the set of matrices
with bounded elementwise $\ell_\infty$ norm.  This constraint
eliminates matrices that are overly ``spiky'' (i.e., concentrate too
much of their mass in a single position); as discussed in the
paper~\cite{NegWai10b}, such spikiness control is necessary in order to
bound the non-identifiable component of the matrix completion model.

\bcors[Matrix completion]
\label{CorMatComp}
Under the conditions of Theorem~\ref{ThmMainCon}, suppose that
$\ThetaStar \in \Ball_\qpar(\radq)$, and that we solve the
program~\eqref{EqnMestNetflix} with $\radcon \leq
\nuclear{\ThetaStar}$.  As long as $\numobs > \PLAINCON_0
\radq^{1/(1-\qpar/2)} \; \mdim \log \mdim$ for a sufficiently large
constant $\PLAINCON_0$, then with probability at least $1 - \exp(-
\PLAINCON_1 \mdim \log \mdim)$, there is a contraction coefficient
$\contrac_t \in (0,1)$ that decreases with $t$ such that for all
iterations $t = 0, 1, 2, \ldots$,
\begin{align}
\label{EqnMatComp}
\frob{\Iter{t+1} - \ThetaHat}^2 & \leq \contrac_t^t \; \frob{\Iter{0} -
  \ThetaHat}^2 + \PLAINCON_2 \, \Big \{ \radq \big(\frac{ \,
  \spike^2 \mdim \log \mdim}{\numobs} \big)^{1-\qpar/2} +
\frob{\ThetaHat - \ThetaStar}^2 \Big \}.
\end{align}

\ecors

In some cases, the bound on $\|\Theta\|_\infty$ in the
algorithm~\eqref{EqnMestNetflix} might be unknown, or
undesirable. While this constraint is necessary in
general~\cite{NegWai10b}, it can be avoided if more information such
as the sampling distribution (that is, the distribution of $X_i$) is
known and used to construct the estimator. In this case, Koltchinskii
et al.~\cite{KoltchinskiiLoTs2011} show error bounds on a nuclear norm
penalized estimator without requiring $\ell_\infty$ bound on
$\ThetaHat$. 

Again a similar corollary of Theorem~\ref{ThmMainLag} can be derived
by combining the proof of Corollary~\ref{CorMatComp} with that of
Theorem~\ref{ThmMainLag}. An interesting aspect of this problem is
that the condition~\ref{EqnFranceAnnoy}(b) takes the form $\regpar >
\frac{\plaincon\alpha\sqrt{\mdim\log\mdim/\numobs}}{1-\contrac}$,
where $\alpha$ is a bound on $\|\Theta\|_\infty$. This condition is
independent of $\radlag$, and hence, given a sample size as stated in
the corollary, the algorithm always converges geometrically for any
radius $\radlag \geq \nuclear{\ThetaStar}$.


\subsection{Matrix decomposition problems}
\label{SecMatDecomp}

In recent years, various researchers have studied methods for solving
the problem of matrix decomposition
(e.g.,~\cite{Chand09,CandesLiMaWr2009,XuCaSa2010,AgarwalNegWai11,
  HsuKaZh2010}).  The basic problem has the following form: given a
pair of unknown matrices $\ThetaStar$ and $\GammaStar$, both lying in
$\real^{\mdima \times \mdimb}$, suppose that we observe a third matrix
specified by the model $Y = \ThetaStar + \GammaStar + W$, where $W \in
\real^{\mdima \times \mdimb}$ represents observation noise.  Typically
the matrix $\ThetaStar$ is assumed to be low-rank, and some
low-dimensional structural constraint is assumed on the matrix
$\GammaStar$. For example, the papers~\cite{Chand09,
  CandesLiMaWr2009,HsuKaZh2010} consider the setting in which
$\GammaStar$ is sparse, while Xu et al.~\cite{XuCaSa2010} consider a
column-sparse model, in which only a few of the columns of
$\GammaStar$ have non-zero entries.  In order to illustrate the
application of our general result to this setting, here we consider
the low-rank plus column-sparse framework~\cite{XuCaSa2010}.  (We note
that since the $\ell_1$-norm is decomposable, similar results can
easily be derived for the low-rank plus entrywise-sparse setting as
well.)

Since $\ThetaStar$ is assumed to be low-rank, as before we use the
nuclear norm $\nuclear{\Theta}$ as a regularizer (see
Section~\ref{SecMatrixReg}).  We assume that the unknown matrix
$\GammaStar \in \real^{\mdima \times \mdimb}$ is column-sparse, say
with at most $\spindex < \mdimb$ non-zero columns.  A suitable convex
regularizer for this matrix structure is based on the \emph{columnwise
  $(1,2)$-norm}, given by
\begin{align}
\mynorm{\Gamma}{1}{2} & \defn \sum_{j=1}^\mdimb \|\Gamma_j\|_2,
\end{align}
where $\Gamma_j \in \real^{\mdima}$ denotes the $j^{th}$ column of
$\Gamma$.  Note also that the dual norm is given by the
\emph{elementwise $(\infty,2)$-norm} $\mynorm{\Gamma}{\infty}{2} =
\max_{j=1, \ldots, \mdimb} \|\Gamma_j\|_2$, corresponding to the
maximum $\ell_2$-norm over columns.

In order to estimate the unknown pair $(\ThetaStar, \GammaStar)$, we
consider the $M$-estimator
\begin{align}
  \label{eqn:matdecompobj}
  (\ThetaHat, \GammaHat) & \defn \arg\min_{\Theta, \Gamma} \frob{Y -
    \Theta - \Gamma}^2\quad\mbox{such that}~~\nuclear{\Theta} \leq
  \radcon_\Theta, ~~ \mynorm{\Gamma}{1}{2} \leq
  \radcon_\Gamma~\mbox{and}~\mynorm{\Theta}{\infty}{2} \leq
  \frac{\spike}{\sqrt{\mdimb}}  
\end{align}

The first two constraints restrict $\Theta$ and $\Gamma$ to a nuclear
norm ball of radius $\radcon_\Theta$ and a $(1,2)$-norm ball of radius
$\radcon_\Gamma$, respectively.  The final constraint controls the
``spikiness'' of the low-rank component $\Theta$, as measured in the
$(\infty,2)$-norm, corresponding to the maximum $\ell_2$-norm over the
columns.  As with the elementwise $\ell_\infty$-bound for matrix
completion, this additional constraint is required in order to limit
the non-identifiability in matrix decomposition.  (See the
paper~\cite{AgarwalNegWai11} for more discussion of
non-identifiability issues in matrix decomposition.)\\

With this set-up, consider the projected gradient algorithm when
applied to the matrix decomposition problem: it generates a sequence
of matrix pairs $(\Theta^t, \Gamma^t)$ for $t = 0, 1, 2, \ldots$, and
the optimization error is characterized in terms of the matrices
$\DelHatIt{t}_\Theta \defn \Theta^t - \ThetaHat$ and
\mbox{$\DelHatIt{t}_\Gamma \defn \Gamma^t - \GammaHat$}.  Finally, we
measure the optimization error at time $t$ in terms of the squared
Frobenius error \mbox{$e^2(\DelHatIt{t}_\Theta, \DelHatIt{t}_\Gamma)
  \defn \frob{\DelHatIt{t}_\Theta}^2 + \frob{\DelHatIt{t}_\Gamma}^2$,}
summed across both the low-rank and column-sparse components.

\bcors[Matrix decomposition]
\label{CorMatDecomp}
Under the conditions of Theorem~\ref{ThmMainCon}, suppose that
\mbox{$\mynorm{\ThetaStar}{\infty}{2} \leq
  \frac{\spike}{\sqrt{\mdimb}}$} and $\GammaStar$ has at most
$\spindex$ non-zero columns.  If we solve the convex
program~\eqref{eqn:matdecompobj} with $\radcon_\Theta \leq
\nuclear{\ThetaStar}$ and $\radcon_\Gamma \leq
\mynorm{\GammaStar}{1}{2}$, then for all iterations $t = 0, 1, 2,
\ldots$,
\begin{align*}
e^2(\DelHatIt{t}_\Theta, \DelHatIt{t}_\Gamma) & \leq
\left(\frac{3}{4}\right)^t \;e^2(\DelHatIt{0}_\Theta,
\DelHatIt{0}_\Gamma) + \PLAINCON \; \biggr(\frob{\GammaHat -
  \GammaStar}^2 + \spike^2 \frac{\spindex}{\mdimb}\biggr).
\end{align*}
\ecors
This corollary has some unusual aspects, relative to the previous
corollaries.  First of all, in contrast to the previous results, the
guarantee is a deterministic one (as opposed to holding with high
probability).  More specifically, the RSC/RSM conditions hold
deterministic sense, which should be contrasted with the high
probability statements given in
Corollaries~\ref{CorSparse}-\ref{CorMatComp}. Consequently, the
effective conditioning of the problem does not depend on sample size
and we are guaranteed geometric convergence at a fixed rate,
independent of sample size. The additional tolerance term is
completely independent of the rank of $\ThetaStar$ and only depends on
the column-sparsity of $\GammaStar$.


\section{Simulation results}
\label{SecSimulations}

In this section, we provide some experimental results that confirm the
accuracy of our theoretical results, in particular showing excellent
agreement with the linear rates predicted by our theory.  In addition,
the rates of convergence slow down for smaller sample sizes, which
lead to problems with relatively poor conditioning.  In all the
simulations reported below, we plot the log error $\plainnorm{\iter{t}
  - \thetahat}$ between the iterate $\iter{t}$ at time $t$ versus the
final solution $\thetahat$.  Each curve provides the results averaged
over five random trials, according to the ensembles which we now
describe.

\subsection{Sparse regression} 

We begin by considering the linear regression model $y = X \thetastar
+ w$ where $\thetastar$ is the unknown regression vector belonging to
the set $\Ball_\qpar(\radq)$, and i.i.d. observation noise $w_i \sim
N(0,0.25)$. We consider a family of ensembles for the random design
matrix $\Xmat \in \real^{\numobs \times \usedim}$.  In particular, we
construct $\Xmat$ by generating each row $x_i \in \real^{\usedim}$
independently according to following procedure.  Let $z_1, \ldots,
z_\numobs$ be an i.i.d. sequence of $N(0,1)$ variables, and fix some
correlation parameter $\toep \in [0,1)$.  We first initialize by
  setting $x_{i,1} = z_1/\sqrt{1-\toep^2}$, and then generate the
  remaining entries by applying the recursive update $x_{i,t+1} =
  \toep x_{i,t} + z_t$ for $t = 1, 2, \ldots, \usedim -1$, so that
  $x_i \in \real^{\usedim}$ is a zero-mean Gaussian random vector. It
  can be verified that all the eigenvalues of $\CovMat = \cov(x_i)$
  lie within the interval $[\frac{1}{(1+\toep)^2},
    \frac{2}{(1-\toep)^2(1+\toep)}]$, so that $\Sigma$ has a a finite
  condition number for all $\toep \in [0,1)$.  At one extreme, for
    $\toep = 0$, the matrix $\Sigma$ is the identity, and so has
    condition number equal to $1$.  As $\toep \rightarrow 1$, the
    matrix $\Sigma$ becomes progressively more ill-conditioned, with a
    condition number that is very large for $\toep$ close to one.  As
    a consequence, although incoherence conditions like the restricted
    isometry property can be satisfied when $\toep = 0$, they will
    fail to be satisfied (w.h.p.)  once $\toep$ is large enough.

For this random ensemble of problems, we have investigated convergence
rates for a wide range of dimensions $\usedim$ and radii $R_\qpar$.
Since the results are relatively uniform across the choice of these
parameters, here we report results for dimension $\usedim = 20,000$,
and radius $R_q = \lceil (\log \usedim)^2 \rceil$.  In the case $\qpar
= 0$, the radius $R_0 = \spindex$ corresponds to the sparsity
level. The per iteration cost in this case is
$\order(\numobs\usedim)$.  In order to reveal dependence of
convergence rates on sample size, we study a range of the form
$\numobs = \lceil \PREFACT \; \spindex \log \pdim \rceil$, where the
\emph{order parameter} $\PREFACT > 0$ is varied.

\begin{figure}[h!]
  \centering
  \begin{tabular}{ccc}
    \includegraphics[scale=0.33]{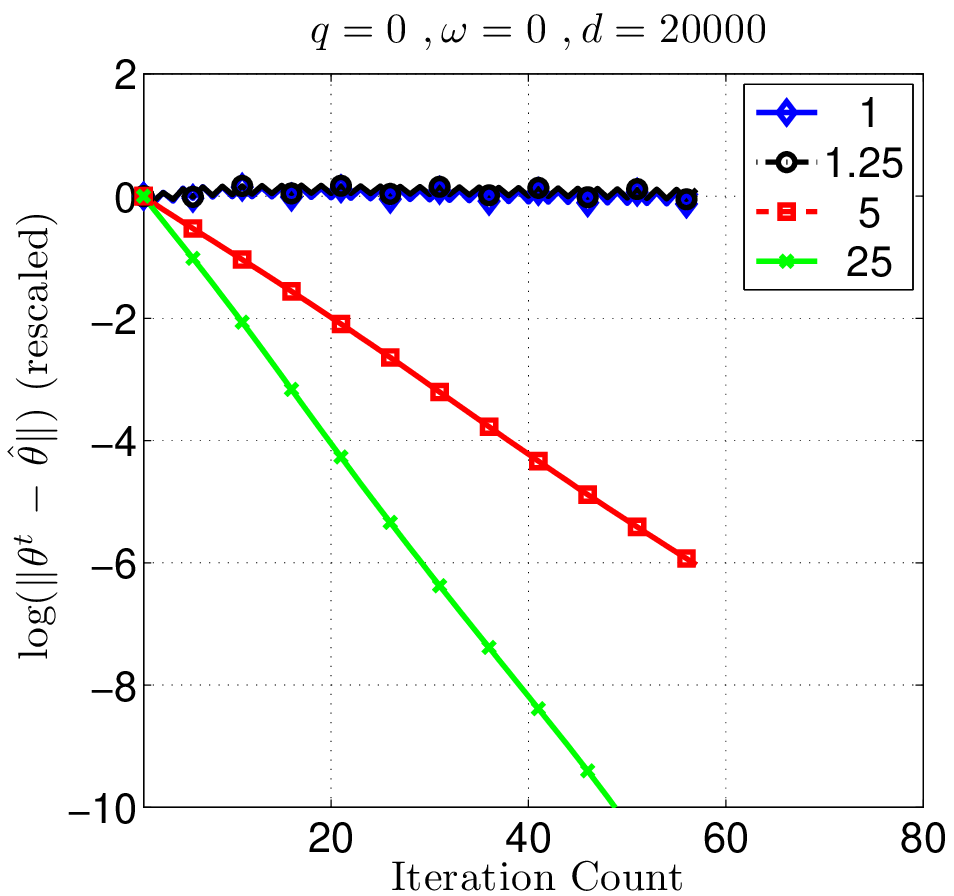}
    & 
    \includegraphics[scale=0.33]{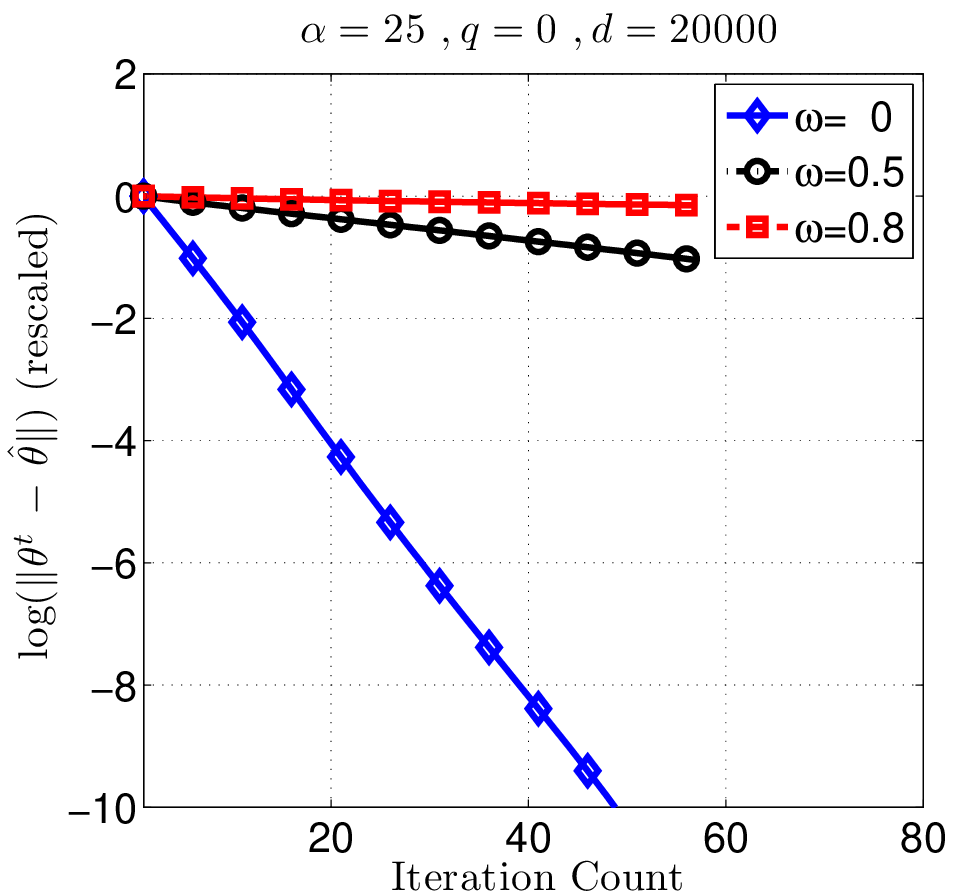}
    &
    \includegraphics[scale=0.33]{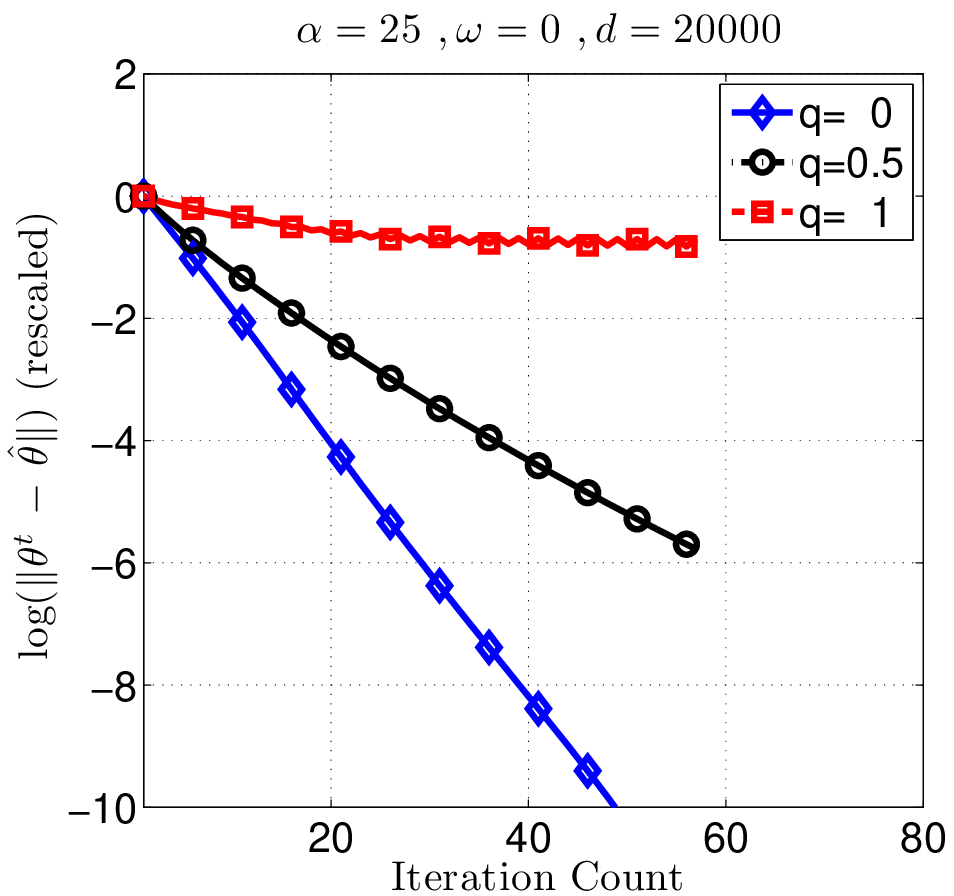} \\
(a) & (b) & (c)
  \end{tabular}
  \caption{\label{fig:sparselinconvergence} Plot of the log of the
    optimization error $\log(\norm{\iter{t} - \betahat}_2)$ in the
    sparse linear regression problem, rescaled so the plots start at
    0.  In this problem, $\pdim = 20000$, $\spindex = \lceil \log\pdim
    \rceil$, $\numobs = \alpha\spindex\log\pdim$. Plot (a) shows
    convergence for the exact sparse case with $\qpar=0$ and $\Sigma =
    I$ (i.e. $\toep = 0$). In panel (b), we observe how convergence
    rates change as the correlation parameter $\toep$ is varied for $q
    = 0$ and $\alpha = 25$. Plot (c) shows the convergence rates when
    $\toep = 0$, $\alpha = 25$ and $\qpar$ is varied.}
\end{figure}

Our first experiment is based on taking the correlation parameter
$\toep = 0$, and the $\ell_\qpar$-ball parameter $\qpar = 0$,
corresponding to exact sparsity.  We then measure convergence rates
for sample sizes specified by $\PREFACT \in \{1,1.25,5,25\}$.  As
shown by the results plotted in panel (a) of
Figure~\ref{fig:sparselinconvergence}, projected gradient descent
fails to converge for $\PREFACT = 1$ or $\PREFACT = 1.25$; in both these
cases, the sample size $\numobs$ is too small for the RSC and RSM
conditions to hold, so that a constant step size leads to oscillatory
behavior in the algorithm.  In contrast, once the order parameter
$\PREFACT$ becomes large enough to ensure that the RSC/RSM conditions
hold (w.h.p.), we observe a geometric convergence of the error
$\norm{\iter{t} - \betahat}_2$. Moreover the convergence rate is faster for
$\PREFACT = 25$ compared to $\PREFACT = 5$, since the RSC/RSM
constants are better with larger sample size.  Such behavior is in
agreement with the conclusions of Corollary~\ref{CorSparse}, which
predicts that the the convergence rate should improve as the number of
samples $\numobs$ is increased.

On the other hand, Corollary~\ref{CorSparse} also predicts that
convergence rates should be slower when the condition number of
$\Sigma$ is worse.  In order to test this prediction, we again studied
an exactly sparse problem ($\qpar = 0$), this time with the fixed
sample size $\numobs = \lceil 25\spindex\log\pdim\rceil$, and we
varied the correlation parameter $\toep \in \{0, 0.5, 0.8\}$. As shown
in panel (b) of Figure~\ref{fig:sparselinconvergence}, the convergence
rates slow down as the correlation parameter is increased and for the
case of extremely high correlation of $\toep = 0.8$, the optimization
error curve is almost flat---the method makes very slow progress in
this case.

A third prediction of Corollary~\ref{CorSparse} is that the
convergence of projected gradient descent should become slower as the
sparsity parameter $\qpar$ is varied between exact sparsity ($\qpar =
0$), and the least sparse case ($\qpar =1$).  (In particular, note for
$\numobs > \log \usedim$, the quantity $\phin$ from
equation~\eqref{EqnPhinSparse} is monotonically increasing with
$\qpar$.)  Panel (c) of Figure~\ref{fig:sparselinconvergence} shows
convergence rates for the fixed sample size $\numobs =
25\spindex\log\pdim$ and correlation parameter $\toep = 0$, and with
the sparsity parameter $\qpar \in \{0, 0.5, 1.0\}$.  As expected, the
convergence rate slows down as $\qpar$ increases from $0$ to $1$.
Corollary~\ref{CorSparse} further captures how the contraction factor
changes as the problem parameters $(\spindex, \pdim, \numobs)$ are
varied. In particular, it predicts that as we change the triplet
simultaneously, while holding the ratio $\alpha = \spindex \log
\pdim/\numobs$ constant, the convergence rate should stay the same. We
recall that this phenomenon was indeed demonstrated in
Figure~\ref{FigDim} in Section~\ref{SecIntro}.

\subsection{Low-rank matrix estimation}

We also performed experiments with two different versions of low-rank
matrix regression.  Our simulations applied to instances of the
observation model $y_i = \tracer{X_i}{\ThetaStar} + w_i$, for $i = 1,
2, \ldots, \numobs$, where $\ThetaStar \in \real^{200 \times 200}$ is
a fixed unknown matrix, $X_i \in \real^{200 \times 200}$ is a matrix
of covariates, and \mbox{$w_i \sim N(0, 0.25)$} is observation noise.
In analogy to the sparse vector problem, we performed simulations with
the matrix $\ThetaStar$ belonging to the set $\Ball_\qpar(\radq)$ of
approximately low-rank matrices, as previously defined in
equation~\eqref{EqnMatrixEllq} for $\qpar \in [0,1]$.  The case $\qpar
= 0$ corresponds to the set of matrices with rank at most $r = R_0$,
whereas the case $\qpar = 1$ corresponds to the ball of matrices with
nuclear norm at most $R_1$.

\begin{figure}[h!]
  \begin{center}
    \begin{tabular}{ccc}
      \widgraph{.4\textwidth}{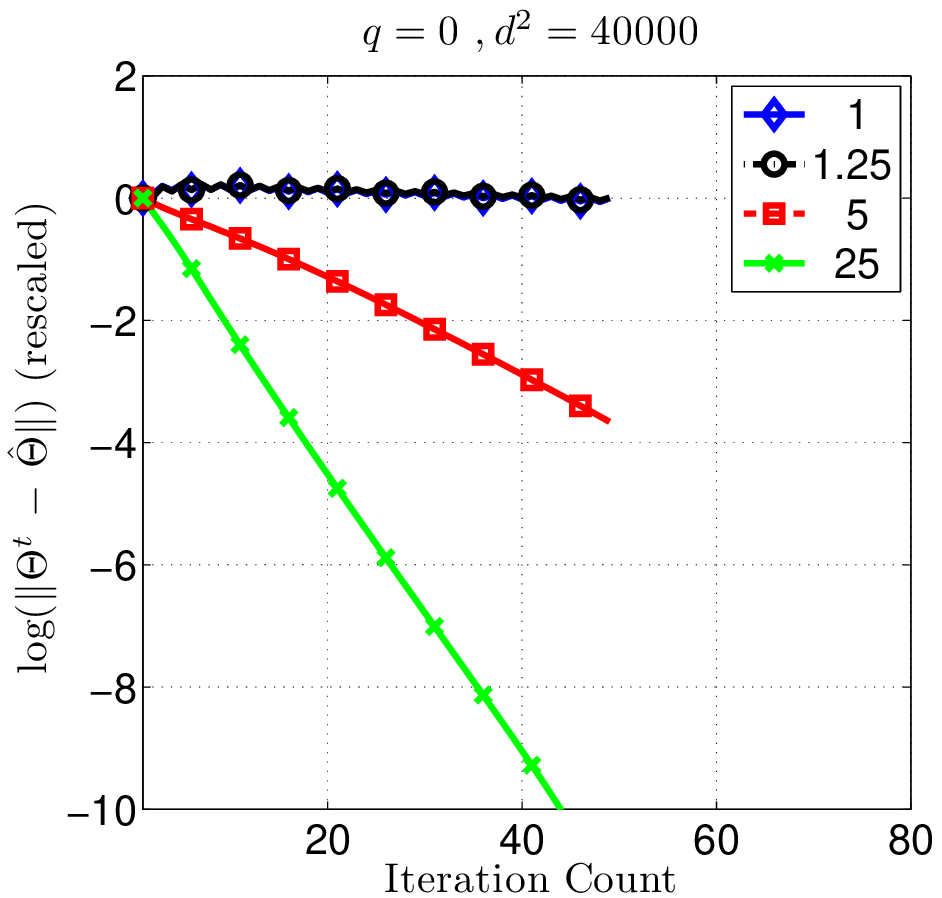}
      & \hspace*{.2in} &
      \widgraph{.4\textwidth}{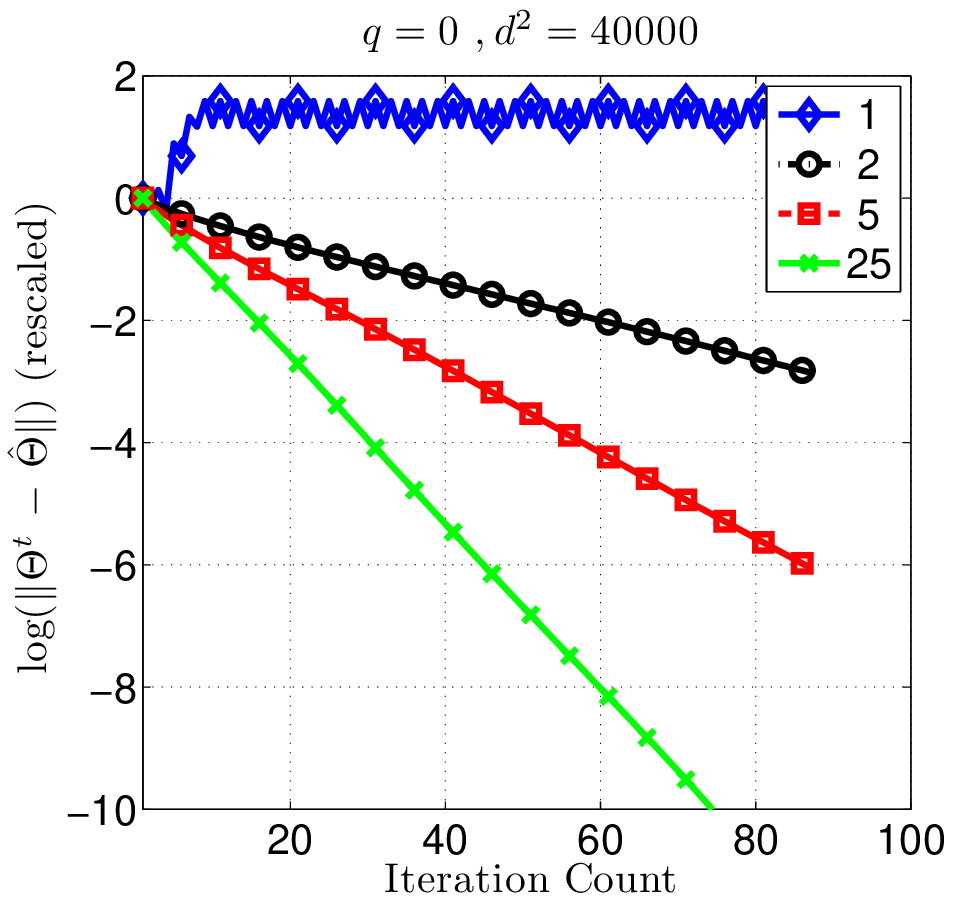}
      \\
(a) & & (b)
 \end{tabular}
 \caption{(a) Plot of log Frobenius error $\log(\frobnorm{\Theta^t -
     \ThetaHat})$ versus number of iterations in matrix compressed
   sensing for a matrix size $\mdim = 200$ with rank $R_0 = 5$, and
   sample sizes $\numobs = \alpha R_0 \mdim$.  For \mbox{$\alpha \in
     \{1, 1.25\}$,} the algorithm oscillates, whereas geometric
   convergence is obtained for $\alpha \in \{5,25\}$, consistent with
   the theoretical prediction.  (b) Plot of log Frobenius error
   $\log(\frobnorm{\Theta^t - \ThetaHat})$ versus number of iterations
   in matrix completion with $\mdim = 200$, $R_0 = 5$, and $\numobs =
   \alpha R_o \mdim \log(\mdim)$ with $\alpha \in \{1,2,5,25\}$.  For
   $\alpha \in \{2,5,25\}$ the algorithm enjoys geometric
   convergence.}
\label{fig:matrixconvergence}
  \end{center}
\end{figure}

In our first set of matrix experiments, we considered the matrix
version of compressed sensing~\cite{Recht09}, in which each matrix
$X_i \in \real^{200 \times 200}$ is randomly formed with
i.i.d. $N(0,1)$ entries, as described in Section~\ref{SecMatCompress}.
In the case $\qpar = 0$, we formed a matrix $\ThetaStar \in \real^{200
  \times 200}$ with rank $R_0 = 5$, and performed simulations over the
sample sizes $\numobs = \PREFACT R_0 \, \mdim$, with the parameter
$\PREFACT \in \{1, 1.25, 5, 25\}$. The per iteration cost in this case
is $\order(\numobs \mdim^2)$. As seen in panel (a) of
Figure~\ref{fig:matrixconvergence}, the projected gradient descent
method exhibits behavior that is qualitatively similar to that for the
sparse linear regression problem.  More specifically, it fails to
converge when the sample size (as reflected by the order parameter
$\PREFACT$) is too small, and converges geometrically with a
progressively faster rate as $\PREFACT$ is increased.  We have also
observed similar types of scaling as the matrix sparsity parameter is
increased from $\qpar = 0$ to $\qpar = 1$.

In our second set of matrix experiments, we studied the behavior of
projected gradient descent for the problem of matrix completion, as
described in Section~\ref{SecMatComplete}.  For this problem, we again
studied matrices of dimension $\mdim = 200$ and rank $R_0 = 5$, and we
varied the sample size as \mbox{$\numobs = \PREFACT \: R_0 \: \mdim
  \log \mdim$} for $\PREFACT \in \{1,2,5,25\}$. As shown in panel (b)
of Figure~\ref{fig:matrixconvergence}, projected gradient descent for
matrix completion also enjoys geometric convergence for $\PREFACT$
large enough.


\section{Proofs}
\label{SecProofs}

In this section, we provide the proofs of our results. Recall that we
use $\DelHatIt{t} \defn \iter{t} - \thetahat$ to denote the
optimization error, and $\Delta^* = \thetahat - \thetastar$ to denote
the statistical error.  For future reference, we point out a slight
weakening of restricted strong convexity (RSC), useful for obtaining
parts of our results.  As the proofs to follow reveal, it is only
necessary to enforce an RSC condition of the form
  \begin{equation}
    \Tay(\iter{t}; \thetahat) \geq \frac{\lossrsc}{2} \,
    \plainnorm{\iter{t} - \thetahat}^2 - \FUNLOW(\LossN) \;
    \RegSq{\iter{t} - \thetahat} - \delta^2,
    \label{eqn:RSCgeneral}
  \end{equation}
which is milder than the original RSC condition~\eqref{EqnRSC}, in
that it applies only to differences of the form $\iter{t} -
\thetahat$, and allows for additional slack $\delta$.  We make use of
this refined notion in the proofs of various results to follow.

With this relaxed RSC condition and the same RSM condition as before,
our proof shows that
\begin{align}
    \label{eqn:thmmaingeneral}
    \plainnorm{\iter{t+1} - \thetahat}^2 & \leq \contrac^t \,
    \plainnorm{\iter{0} - \thetahat}^2 + \frac{\HACKSTATERR +
      2\delta^2/\Lsmooth}{1-\contrac} \qquad \mbox{for all iterations
      $t = 0, 1, 2, \ldots$.}
  \end{align}
Note that this result reduces to the previous statement when $\delta =
0$.  This extension of Theorem~\ref{ThmMainCon} is used in the proofs
of Corollaries~\ref{CorMatComp} and~\ref{CorMatDecomp}.

We will assume without loss of generality that all the iterates lie in
the subset $\PARNEW$ of $\Parset$. This can be ensured by augmenting
the loss with the indicator of $\PARNEW$ or equivalently performing
projections on the set $\PARNEW \cap \ball{\Regplain}{\radcon}$ as
mentioned earlier.

\subsection{Proof of Theorem~\ref{ThmMainCon}}

Recall that Theorem~\ref{ThmMainCon} concerns the constrained
problem~\eqref{EqnMestCon}.  The proof is based on two technical
lemmas.  The first lemma guarantees that at each iteration $t = 0, 1,
2, \ldots$, the optimization error $\DelHatIt{t} = \iter{t} -
\thetahat$ belongs to an interesting constraint set defined by the
regularizer.
\blems
\label{LemMUFCvsMCBoring}
Let $\thetahat$ be any optimum of the constrained
problem~\eqref{EqnMestCon} for which $\Reg{\thetahat} = \radcon$.  Then
for any iteration $t = 1, 2, \ldots$ and for any
$\Plainreg$-decomposable subspace pair $(\ModelSet, \BsetPerp)$, the
optimization error $\DelHatIt{t} \defn \iter{t} - \thetahat$ belongs
to the set
\begin{align}
\label{EqnMUFCvsMCBoring}
\KeySet(\ModelSet; \Bset; \thetastar) & \defn \biggr \{ \Delta \in
\Parset \, \mid \Reg{\Delta} \leq 2 \, \Compat(\Bset) \, \norm{\Delta}
+ 2 \Reg{\BigProjModelPerp{\thetastar}} + 2 \Reg{\DeltaStar} +
\Compat(\Bset) \norm{\DeltaStar} \biggr \}.
\end{align}
\elems

\noindent The proof of this lemma, provided in
Appendix~\ref{AppLemMUFCvsMCBoring}, exploits the decomposability of
the regularizer in an essential way. \\

The structure of the set~\eqref{EqnMUFCvsMCBoring} takes a simpler
form in the special case when $\ModelSet$ is chosen to contain
$\thetastar$ and $\Bset = \ModelSet$.  In this case, we have
$\Reg{\BigProjModelPerp{\thetastar}} = 0$, and hence the optimization
error $\DelHatIt{t}$ satisfies the inequality
\begin{align}
\label{EqnNonCone}
\Reg{\DelHatIt{t}} & \leq 2 \, \Compat(\ModelSet) \, \big \{
\norm{\DelHatIt{t}} + \norm{\DeltaStar} \big \} + 2 \Reg{\DeltaStar}.
\end{align}
An inequality of this type, when combined with the definitions of
RSC/RSM, allows us to establish the curvature conditions required to
prove globally geometric rates of convergence. \\

\noindent We now state a second lemma under the more general RSC
condition~\eqref{eqn:RSCgeneral}:
\blems
\label{LemResGrad}
Under the RSC condition~\eqref{eqn:RSCgeneral} and RSM
condition~\eqref{EqnRSM}, for all $t = 0,1,2, \ldots$, we have
\begin{multline}
\label{EqnResGrad}
\Lsmooth \, \inprod{\iter{t} - \iter{t+1}}{\iter{t} - \betahat} \\
\geq \Big \{ \frac{\Lsmooth}{2} \plainnorm{\iter{t} - \iter{t+1}}^2 -
\FUNUP(\LossN) \RegSq{\iter{t+1} -\iter{t}} \Big \} + \Big \{
\frac{\lossrsc}{2} \plainnorm{\iter{t} - \betahat}^2 - \FUNLOW(\LossN)
\, \RegSq{\iter{t} - \thetahat} - \delta^2 \Big \}.
\end{multline}
\elems 
The proof of this lemma, provided in Appendix~\ref{AppLemResGrad},
follows along the lines of the intermediate result within Theorem
2.2.8 of Nesterov~\cite{Nesterov04}, but with some care required to
handle the additional terms that arise in our weakened forms of strong
convexity and smoothness.

\noindent \\

\vsmall

Using these auxiliary results, let us now complete the the proof of
Theorem~\ref{ThmMainCon}.  We first note the elementary relation
\begin{align}
\label{EqnElementary}
\plainnorm{\iter{t+1} - \betahat}^2 & = \plainnorm{\iter{t} - \betahat
  - \iter{t} + \iter{t+1}}^2 \; = \plainnorm{\iter{t} - \betahat}^2 +
\plainnorm{\iter{t} - \iter{t+1}}^2 - 2 \inprod{\iter{t} -
  \betahat}{\iter{t} - \iter{t+1}}.
\end{align}
We now use Lemma~\ref{LemResGrad} and the more general form of
RSC~\eqref{eqn:RSCgeneral} to control the cross-term, thereby
obtaining the upper bound
\begin{align*}
\plainnorm{\iter{t+1} - \betahat}^2 & \leq \plainnorm{\iter{t} -
  \betahat}^2 - \frac{\lossrsc}{\Lsmooth} \plainnorm{\iter{t} -
  \betahat}^2 + \frac{2 \FUNUP(\LossN)}{\Lsmooth} \RegSq{\iter{t+1} -
  \iter{t}} + \frac{2 \FUNLOW(\LossN)}{\Lsmooth} \RegSq{\iter{t} -
  \thetahat}  + \frac{2\delta^2}{\Lsmooth}\\
& = \big(1 - \frac{\lossrsc}{\Lsmooth} \big) \plainnorm{\iter{t} -
\betahat}^2 + \frac{2 \FUNUP(\LossN)}{\Lsmooth} 
\RegSq{\iter{t+1} -
\iter{t}} + \frac{2 \FUNLOW(\LossN)}{\Lsmooth} \RegSq{\iter{t} -
\thetahat} + \frac{2\delta^2}{\Lsmooth}.
\end{align*}
We now observe that by triangle inequality and the Cauchy-Schwarz
inequality,
\begin{align*}
\RegSq{\iter{t+1} - \iter{t}} & \leq \big (\Reg{ \iter{t+1} -
  \thetahat} + \Reg{\thetahat - \iter{t}} \big)^2 \; \leq \; 2
\RegSq{\iter{t+1} - \thetahat} + 2 \RegSq{\iter{t} - \thetahat}.
\end{align*}
Recall the definition of the optimization error $\DelHatIt{t} \defn
\iter{t} - \thetahat$, we have the upper bound
\begin{align}
\label{EqnDelicate}
\plainnorm{\DelHatIt{t+1}}^2 & \leq \big(1 - \frac{\lossrsc}{\Lsmooth}
\big) \plainnorm{\DelHatIt{t}}^2 + \frac{4 \FUNUP(\LossN)}{\Lsmooth}
\RegSq{\DelHatIt{t+1}} + \frac{4 \FUNUP(\LossN) + 2
\FUNLOW(\LossN)}{\Lsmooth} \RegSq{\DelHatIt{t}} +
\frac{2\delta^2}{\Lsmooth}. 
\end{align}

We now apply Lemma~\ref{LemMUFCvsMCBoring} to control the terms
involving $\Regplain^2$.  In terms of squared quantities, the
inequality~\eqref{EqnMUFCvsMCBoring} implies that
\begin{align*}
\RegSq{\DelHatIt{t}} & \leq 4 \, \Compat^2(\Bset^\perp) \,
\norm{\DelHatIt{t}}^2 + 2 \HACKERRSQ \qquad \mbox{for all $t = 0, 1,
2, \ldots$,}
\end{align*}
where we recall that $\Compat^2(\BsetPerp)$ is the subspace
compatibility~\eqref{EqnDefnSubspaceCompat} and $\HACKERRSQ$
accumulates all the residual terms.  Applying this bound twice---once
for $t$ and once for $t+1$---and substituting into
equation~\eqref{EqnDelicate} yields that $\big \{ 1 - \frac{16
  \Compat^2(\BsetPerp) \FUNUP(\LossN)}{\Lsmooth} \big \}
\plainnorm{\DelIt{t+1}}^2$ is upper bounded by
\begin{align*}
\Big \{ 1 - \frac{\lossrsc}{\Lsmooth} + \frac{16 \Compat^2(\Bset^\perp)
  \big(\FUNUP(\LossN) + \FUNLOW(\LossN) \big)}{\Lsmooth} \Big \}
\plainnorm{\DelIt{t}}^2 + \frac{16 \big( \FUNUP(\LossN) +
  \FUNLOW(\LossN) \big) \HACKERRSQ}{\Lsmooth} +
\frac{2\delta^2}{\Lsmooth}.
\end{align*}
Under the assumptions of Theorem~\ref{ThmMainCon}, we are guaranteed
 that $\frac{16 \Compat^2(\Bset^\perp) \FUNUP(\LossN)}{\Lsmooth} < 1/2$, and
 so we can re-arrange this inequality into the form
\begin{align}
\label{EqnBasic}
\plainnorm{\DelIt{t+1}}^2 & \leq \contrac \, \plainnorm{\DelIt{t}}^2 +
\HACKSTATERR + \frac{2\delta^2}{\Lsmooth}
\end{align}
where $\contrac$ and $\HACKSTATERR$ were previously defined in
equations~\eqref{EqnDefnContrac} and~\eqref{EqnDefnHackStatErr}
respectively.  Iterating this recursion yields
\begin{align*}
\plainnorm{\DelIt{t+1}}^2 & \leq \contrac^t \, \plainnorm{\DelIt{0}}^2
+ \biggr(\HACKSTATERR   +
\frac{2\delta^2}{\Lsmooth}\biggr)\,\big(\sum_{j=0}^t \contrac^j
\big). 
\end{align*}
The assumptions of Theorem~\ref{ThmMainCon} guarantee that $\contrac
\in (0,1)$, so that summing the geometric series yields the
claim~\eqref{EqnMainConRate}.


\subsection{Proof of Theorem~\ref{ThmMainLag}}

The Lagrangian version of the optimization program is based on solving
the convex program~\eqref{EqnMestReg}, with the objective function
$\composite(\thetapar) = \LossN(\thetapar) + \regpar \Reg{\thetapar}$.
Our proof is based on analyzing the error $\composite(\iter{t}) -
\composite(\thetahat)$ as measured in terms of this objective
function.  It requires two technical lemmas, both of which are stated
in terms of a given tolerance $\objerror > 0$, and an integer $T > 0$
such that
\begin{align}
\label{EqnAssCondition} 
\obj(\iter{t}) - \obj(\thetahat) & \leq \objerror \qquad \mbox{for all
  $t \geq T$.}
\end{align}
Our first
technical lemma is analogous to Lemma~\ref{LemMUFCvsMCBoring}, and
restricts the optimization error $\DelHatIt{t} = \iter{t} - \thetahat$
to a cone-like set.
\blems[Iterated Cone Bound (ICB)]
\label{LemICB}
Let $\thetahat$ be any optimum of the regularized
$M$-estimator~\eqref{EqnMestReg}.  Under
condition~\eqref{EqnAssCondition} with parameters $(T, \objerror)$,
for any iteration $t \geq T$ and for any $\Plainreg$-decomposable
subspace pair $(\ModelSet, \BsetPerp)$, the optimization error
$\DelHatIt{t} \defn \iter{t} - \thetahat$ satisfies
\begin{align}
  \label{EqnICB}
  \Reg{\DelHatIt{t}} & \leq 4 \Compat(\Bset) \norm{\DelHatIt{t}} + 8
  \Compat(\Bset) \norm{\DeltaStar} + 8
  \Reg{\BigProjModelPerp{\thetastar}} + 2
  \min\left(\frac{\objerror}{\regpar},\radlag\right)
\end{align}
\elems
Our next lemma guarantees sufficient decrease of the objective value
difference $\composite(\iter{t}) - \composite(\thetahat)$.
Lemma~\ref{LemICB} plays a crucial role in its proof.  Recall the
definition~\eqref{EqnDefnContracLag} of the compound contraction
coefficient $\contrac(\LossN; \Bset)$, defined in terms of the related
quantities $\slopa(\Bset)$ and $\slopb(\Bset)$.  Throughout the proof,
we drop the arguments of $\contrac$, $\slopa$ and $\slopb$ so as to
ease notation.
\blems
\label{LemLagDecrease}
Under the RSC~\eqref{eqn:RSCgeneral} and RSM
conditions~\eqref{EqnRSM}, as well as
assumption~\eqref{EqnAssCondition} with \mbox{parameters} $(\objerror,
T)$, for all $t \geq T$, we have
\begin{align*}
  \obj(\iter{t}) - \obj(\thetahat) & \leq \contrac^{t-T}
  (\obj(\iter{T}) - \obj(\thetahat)) + \frac{2}{1-\contrac}
  \slopa(\ModelSet) \; \slopb(\ModelSet) ( \coneslack^2 + \staterr^2
  ),
\end{align*}
where $\coneslack \defn 2 \min(\objerror/\regpar,\radlag)$ and
$\staterr \defn 8 \Compat(\Bset) \norm{\DeltaStar} + 8
\Reg{\BigProjModelPerp{\thetastar}}$.
\elems

We are now in a position to prove our main theorem, in particular via
a recursive application of Lemma~\ref{LemLagDecrease}.  At a high
level, we divide the iterations $t = 0, 1, 2, \ldots$ into a series of
disjoint epochs $[T_k, T_{k+1})$ with $0 = T_0 \leq T_1 \leq T_2 \leq \cdots$.
  Moreover, we define an associated sequence of tolerances $\tol_0 >
  \tol_1 > \cdots$ such that at the end of epoch $[T_{k-1}, T_k)$, the
    optimization error has been reduced to $\tol_k$.  Our analysis
    guarantees that $\composite(\iter{t}) - \composite(\thetahat) \leq
    \tol_k$ for all $t \geq T_k$, allowing us to apply
    Lemma~\ref{LemLagDecrease} with smaller and smaller values of
    $\objerror$ until it reduces to the statistical error $\staterr$.

At the first iteration, we have no a priori bound on the error
$\objerror_0 = \obj(\iter{0}) - \obj(\thetahat)$. However, since
Lemma~\ref{LemLagDecrease} involves the quantity $\coneslack =
\min(\tol/\regpar, \radlag)$, we may still apply it\footnote{It is for
  precisely this reason that our regularized $M$-estimator includes
  the additional side-constraint defined in terms of $\radlag$.}  at
the first epoch with $\coneslack_0 = \radlag$ and $T_0 = 0$.  In this
way, we conclude that for all $t \geq 0$,
\begin{align*}
  \obj(\iter{t}) - \obj(\thetahat) & \leq \contrac^{t} (\obj(\iter{0})
  - \obj(\thetahat)) + \frac{2}{1-\contrac} \slopa \slopb ( \radlag^2
  + \staterr^2 ).
\end{align*}

Now since the contraction coefficient $\contrac \in (0,1)$, for all
iterations $t \geq T_1 \defn (\lceil \log(2 \,
\objerror_0/\objerror_1)/\log(1/\contrac) \rceil)_+$, we are
guaranteed that
\begin{align*}
  \obj(\iter{t}) - \obj(\thetahat) & \leq \underbrace{\frac{4 \,
      \slopa \slopb}{1-\contrac} (\radlag^2 +
    \staterr^2)}_{\objerror_1} \; \leq \frac{8 \slopa
    \slopb}{1-\contrac} \max(\radlag^2,\staterr^2).
\end{align*}

This same argument can now be applied in a recursive manner.  Suppose
that for some $k \geq 1$, we are given a pair $(\objerror_k, T_k)$
such that condition~\eqref{EqnAssCondition} holds.  An application of
Lemma~\ref{LemLagDecrease} yields the bound
\begin{align*}
  \obj(\iter{t}) - \obj(\thetahat) & \leq \contrac^{t-T_k}
  (\obj(\iter{T_k}) - \obj(\thetahat)) + \frac{2 \, \slopa \slopb
  }{1-\contrac} ( \coneslack_k^2 + \staterr^2 ) \qquad \mbox{for all
    $t \geq T_k$.}
\end{align*}
We now define $\objerror_{k+1} \defn \frac{4 \, \slopa
  \slopb}{1-\contrac} (\coneslack_k^2 + \staterr^2)$.  Once again,
since $\contrac < 1$ by assumption, we can choose $T_{k+1} \defn
\lceil \log (2 \objerror_{k}/\objerror_{k+1}) / \log(1/\contrac)
\rceil + T_k$, thereby ensuring that for all $t \geq T_{k+1}$, we have
\begin{align*}
\obj(\iter{t}) - \obj(\thetahat) & \leq \frac{8 \slopa
  \slopb}{1-\contrac} \max( \coneslack_k^2, \staterr^2 ).
\end{align*}
In this way, we arrive at recursive inequalities involving the
tolerances $\{\objerror_k\}_{k=0}^\infty$ and time steps
$\{T_k\}_{k=0}^\infty$---namely
\begin{subequations}
\begin{align}
\label{EqnObjDecrease}
\objerror_{k+1} & \leq \frac{8 \, \slopa \slopb}{1-\contrac}
\max(\coneslack_k^2 , \staterr^2), \qquad \mbox{where $\coneslack_k =
  2 \, \min \{\objerror_{k}/\regpar,\radlag \}$, and} \\
\label{EqnTimeRecurse}
  T_k & \leq k + \frac{\log(2^k
    \objerror_0/\objerror_k)}{\log(1/\contrac)}.
\end{align}
\end{subequations}
Now we claim that the recursion~\eqref{EqnObjDecrease} can be
unwrapped so as to show that
\begin{align}
  \label{EqnObjErrRecur}
  \objerror_{k+1} & \leq \frac{\objerror_k}{4^{2^{k-1}}} \quad
  \mbox{and} \quad \frac{\objerror_{k+1}}{\regpar} \; \leq
  \frac{\radlag}{4^{2^k}} \qquad \mbox{for all $k = 1, 2, \ldots$.}
\end{align}
Taking these statements as given for the moment, let us now show how
they can be used to upper bound the smallest $k$ such that
$\objerror_k \leq \delta^2$.  If we are in the first epoch, the claim
of the theorem is straightforward from
equation~\eqref{EqnObjDecrease}. If not, we first use the
recursion~\eqref{EqnObjErrRecur} to upper bound the number of epochs
needed and then use the inequality~\eqref{EqnTimeRecurse} to obtain
the stated result on the total number of iterations needed. Using the
second inequality in the recursion~\eqref{EqnObjErrRecur}, we see that
it is sufficient to ensure that $\frac{\radlag \regpar}{4^{2^{k-1}}}
\; \leq \; \delta^2$.  Rearranging this inequality, we find that the
error drops below $\delta^2$ after at most
\begin{equation*}
  k_\delta \; \geq \; \log \left ( \log \left ( \frac{\radlag
    \regpar}{\delta^2} \right ) / \log(4) \right )/\log(2) + 1 =
  \log_2\log_2\left(\frac{\radlag\regpar}{\delta^2}\right) 
\end{equation*}
\noindent
epochs.  Combining the above bound on $k_\delta$ with the
recursion~\ref{EqnTimeRecurse}, we conclude that the inequality
$\obj(\iter{t}) - \obj(\thetahat) \leq \delta^2$ is guaranteed to hold
for all iterations
\begin{align*}
  t & \geq k_\delta\left(1 + \frac{\log 2}{\log(1/\contrac)}\right) +
  \frac{\log\frac{\objerror_0}{\delta^2}}{\log(1/\contrac)},
\end{align*}
which is the desired result. \\

\noindent It remains to prove the recursion~\eqref{EqnObjErrRecur},
which we do via induction on the index $k$.  We begin with base case
$k = 1$.  Recalling the setting of $\objerror_1$ and our assumption on
$\regpar$ in the theorem statement~\eqref{EqnFranceAnnoy}, we are
guaranteed that $\objerror_1/\regpar \leq \radlag/4$, so that
$\coneslack_1 \leq \coneslack_0 = \radlag$.  By applying
equation~\eqref{EqnObjDecrease} with $\coneslack_1 =
2\objerror_1/\regpar$ and assuming $\coneslack_1 \geq \staterr$, we
obtain
\begin{align}
\label{eqn:secondepochbound}
\objerror_2 \; \leq \; 
\frac{32\slopa\slopb\objerror_1^2}{(1-\contrac)\regpar^2} \;
\stackrel{(i)}{\leq} \;
\frac{32\slopa\slopb\radlag\objerror_1}{(1-\contrac)4\regpar}
\; \stackrel{(ii)}{\leq} \; \frac{\objerror_1}{4},
\end{align}
where step (i) uses the fact that $\frac{\objerror_1}{\regpar} \leq
\frac{\radlag}{4}$, and step (ii) uses the
condition~\eqref{EqnFranceAnnoy} on $\regpar$.  We have thus verified
the first inequality~\eqref{EqnObjErrRecur} for $k = 1$.  Turning to
the second inequality in the statement~\eqref{EqnObjErrRecur}, using
equation~\ref{eqn:secondepochbound}, we have
\begin{align*}
  \frac{\objerror_2}{\regpar} \; \leq \; \frac{\objerror_1}{4\regpar}
  \; \stackrel{(iii)}{\leq} \; \frac{\radlag}{16},
\end{align*}
where step (iii) follows from the assumption~\eqref{EqnFranceAnnoy} on
$\regpar$.  Turning to the inductive step, we again assume that
$2\objerror_k/\regpar \geq \staterr$ and obtain from 
inequality~\eqref{EqnObjDecrease} 
\begin{align*}
  \objerror_{k+1} \leq
  \frac{32\slopa\slopb\objerror_k^2}{(1-\contrac)\regpar^2} 
  \stackrel{(iv)}{\leq}
  \frac{32\slopa\slopb\objerror_k\radlag}{(1-\contrac)\regpar4^{2^{k-1}}}
  \stackrel{(v)}{\leq} 
  \frac{\objerror_k}{4^{2^{k-1}}}. 
\end{align*}
Here step (iv) uses the second inequality of the inductive
hypothesis~\eqref{EqnObjErrRecur} and step (v) is a consequence of the
condition on $\regpar$ as before. The second part of the induction is
similarly established, completing the proof.


\subsection{Proof of Corollary~\ref{CorGeneric}}

In order to prove this claim, we must show that $\HACKSTATERR$, as
defined in equation~\eqref{EqnDefnHackStatErr}, is of order lower than
$\Exs[\norm{\thetahat - \thetastar}^2] = \Exs[\norm{\DeltaStar}^2]$.
We make use of the following lemma, proved in
Appendix~\ref{AppLemConeOpt}:
\blems
\label{LemConeOpt}

If $\radcon \leq \Reg{\thetastar}$, then for any solution $\thetahat$
of the constrained problem~\eqref{EqnMestCon} and any
$\Regplain$-decomposable subspace pair $(\ModelSet, \Bset^\perp)$, the
statistical error $\DeltaStar = \thetahat - \thetastar$ satisfies the
inequality
\begin{align}
\label{EqnConeOpt}
\Reg{\DeltaStar} & \leq 2 \Compat(\Bset^\perp) \norm{\DeltaStar} +
\Reg{\BigProjModelPerp{\thetastar}}.
\end{align}
\elems

Using this lemma, we can complete the proof of
 Corollary~\ref{CorGeneric}.  Recalling the
 form~\eqref{EqnDefnHackStatErr}, under the condition $\thetastar \in
 \ModelSet$, we have
\begin{align*}
\HACKSTATERR & \defn \frac{32 \big( \FUNUP(\LossN) + \FUNLOW(\LossN)
  \big) \; \big(2 \Reg{\DeltaStar} + \Compat(\Bset^\perp) \norm{\DeltaStar}
  \big)^2 }{\Lsmooth}.
\end{align*}
Using the assumption
$\frac{(\FUNUP(\LossN) + \FUNLOW(\LossN)) \Compat^2(\Bset^\perp)}{\Lsmooth}
= o(1)$, it suffices to show that $\Reg{\DeltaStar} \leq
2 \Compat(\Bset^\perp) \norm{\DeltaStar}$.  Since Corollary~\ref{CorGeneric}
assumes that $\thetastar \in \ModelSet$ and hence that
$\BigProjModelPerp{\thetastar} = 0$, Lemma~\ref{LemConeOpt} implies
that $\Reg{\DeltaStar} \leq 2 \Compat(\Bset^\perp) \norm{\DeltaStar}$, as
required.


\subsection{Proofs of Corollaries~\ref{CorSparse} and~\ref{CorSparseLag}}
\label{SecProofCorSparse}

The central challenge in proving this result is verifying that
suitable forms of the RSC and RSM conditions hold with sufficiently
small parameters $\FUNLOW(\LossN)$ and $\FUNUP(\LossN)$.

\blems
\label{LemGarvesh}
Define the maximum variance $\rhovec(\CovMat) \defn \max \limits_{j=1, 2,
  \ldots, \pdim} \CovMat_{jj}$. Under the conditions of
Corollary~\ref{CorSparse}, there are universal positive constants
$(\PLAINCON_0, \PLAINCON_1)$ such that for all $\Delta \in
\real^\pdim$, we have
\begin{subequations}
\begin{align}
\label{EqnGarveshLower}
\frac{ \|\Xmat \Delta \|_2^2}{\numobs} & \geq \; \frac{1}{2}
\|\CovMatSqrt \Delta\|_2^2 - \PLAINCON_1 \rhovec(\CovMat) \frac{\log
  \pdim}{\numobs} \; \|\Delta\|_1^2, \qquad \mbox{and} \\
\label{EqnGarveshUpper}
\frac{ \|\Xmat \Delta \|_2^2}{\numobs} & \leq \; 2 \| \CovMatSqrt \Delta
\|_2^2 + \PLAINCON_1 \rhovec(\CovMat) \frac{\log \pdim}{\numobs} \;
\|\Delta\|_1^2,
\end{align}
\end{subequations}
with probability at least $1 - \exp(-\PLAINCON_0 \, \numobs)$.
\elems
\noindent Note that this lemma implies that the RSC and RSM conditions
both hold with high probability, in particular with parameters 
\begin{align*}
\lossrsc = \frac{1}{2} \lammin(\CovMat), & \mbox{ and } \quad
\FUNLOW(\LossN) = \PLAINCON_1 \rhovec(\CovMat) \frac{\log
  \pdim}{\numobs}, \qquad \mbox{for RSC, and} \\
\Lsmooth = 2 \lammax(\CovMat) & \mbox{ and } \quad
\FUNUP(\LossN) = \PLAINCON_1 \rhovec(\CovMat) \frac{\log
    \pdim}{\numobs} \qquad \mbox{for RSM.}
\end{align*}
This lemma has been proved by Raskutti et al.~\cite{RasWaiYu10} for
obtaining minimax rates in sparse linear regression.

Let us first prove Corollary~\ref{CorSparse} in the special case of
hard sparsity ($\qpar = 0$), in which $\thetastar$ is supported on a
subset $\Sset$ of cardinality $\kdim$.  Let us define the model
subspace \mbox{$\ModelSet \defn \big \{ \theta \in \real^\pdim \, \mid \,
\theta_j = 0 \quad \mbox{for all $j \notin \Sset$} \big \}$}, so that
$\thetastar \in \ModelSet$.  Recall from Section~\ref{SecLogLinear}
that the $\ell_1$-norm is decomposable with respect to $\ModelSet$ and
$\ModelSetPerp$; as a consequence, we may also set $\Bset^\perp = \ModelSet$
in the definitions~\eqref{EqnDefnContrac}
and~\eqref{EqnDefnHackStatErr}.  By
definition~\eqref{EqnDefnSubspaceCompat} of the subspace compatibility
between with $\ell_1$-norm as the regularizer, and $\ell_2$-norm as
the error norm, we have $\Compat^2(\ModelSet) = \kdim$.  Using the
settings of $\FUNLOW(\LossN)$ and $\FUNUP(\LossN)$ guaranteed by
Lemma~\ref{LemGarvesh} and substituting into
equation~\eqref{EqnDefnContrac}, we obtain a contraction coefficient
\begin{align}
\contrac(\CovMat) & \defn \Big \{ 1 - \frac{\lammin(\CovMat)}{4
  \lammax(\CovMat)} + \phin(\CovMat) \Big \} \; \Big \{ 1 -
\phin(\CovMat) \Big \}^{-1},
\end{align}
where $\phin(\CovMat) \defn \frac{\PLAINCON_2
  \rhovec(\CovMat)}{\lammax(\CovMat)} \; \frac{\kdim \log
  \pdim}{\numobs}$ for some universal constant $\PLAINCON_2$.  A
similar calculation shows that the tolerance term takes the form
\begin{align*}
\HACKSTATERR & \leq \PLAINCON_3 \; \phin(\CovMat) \Big \{
\frac{\|\DeltaStar\|_1^2}{\kdim} + \|\DeltaStar\|_2^2 \Big \} \qquad
\mbox{for some constant $\PLAINCON_3$.}
\end{align*}
Since $\radcon \leq \|\thetastar\|_1$, then Lemma~\ref{LemConeOpt} (as
exploited in the proof of Corollary~\ref{CorGeneric}) shows that
\mbox{$\|\DeltaStar\|^2_1 \leq 4 \kdim \|\DeltaStar\|_2^2$,} and hence
that $\HACKSTATERR \leq \PLAINCON_3 \; \phin(\CovMat) \;
\|\DeltaStar\|_2^2$.  This completes the proof of the
claim~\eqref{EqnSparseBoundZero} for $\qpar = 0$. \\

We now turn to the case $\qpar \in (0,1]$, for which we bound the term
$\HACKSTATERR$ using a slightly different choice of the subspace pair
$\ModelSet$ and $\BsetPerp$.  For a truncation level $\mutrun > 0$ to be
chosen, define the set $\Stau \defn \big \{ j \in \{1, 2, \ldots,
\pdim \} \, \mid \, |\thetastar_j| > \mutrun \big \}$, and define the
associated subspaces $\ModelSet = \ModelSet(\Stau)$ and $\BsetPerp =
\ModelSetPerp(\Stau)$.  By combining Lemma~\ref{LemConeOpt} and the
definition~\eqref{EqnDefnHackStatErr} of $\HACKSTATERR$, for any pair
$(\ModelSet(\Stau), \ModelSetPerp(\Stau))$, we have
\begin{align*}
\epsilon^2(\DeltaStar; \ModelSet, \ModelSetPerp) & \leq
\frac{\PLAINCON \, \rhovec(\CovMat) }{\lammax(\CovMat)} \, \frac{\log
\pdim}{\numobs} \big(\|\BigProjModelPerp{\thetastar}\|_1 +
\sqrt{|\Stau|} \, \|\DeltaStar\|_2 \big)^2,
\end{align*}
where to simplify notation, we have omitted the dependence of
$\ModelSet$ and $\ModelSetPerp$ on $\Stau$.  We now choose the
threshold $\mutrun$ optimally, so as to trade-off the term
$\|\BigProjModelPerp{\thetastar}\|_1$, which decreases as $\mutrun$
increases, with the term $\sqrt{\Stau} \|\DeltaStar\|_2$, which
increases as $\mutrun$ increases.

By definition of $\ModelPerp(\Stau)$, we have
\begin{align*}
\|\BigProjModelPerp{\thetastar}\|_1 & = \sum_{j \notin \Stau}
|\thetastar_j| \; = \; \mutrun \, \sum_{j \notin \Stau}
\frac{|\thetastar_j|}{\mutrun} \; \leq \; \mutrun \, \sum_{j \notin \Stau}
\biggl(\frac{|\thetastar_j|}{\mutrun} \biggr)^\qpar,
\end{align*}
where the inequality holds since $|\thetastar_j| \leq \mutrun$ for all $j
\notin \Stau$.  Now since $\thetastar \in \Ball_\qpar(\radq)$, we
conclude that
\begin{align}
\label{EqnTrunUpper}
\|\BigProjModelPerp{\thetastar}\|_1 & \leq \mutrun^{1-\qpar} \sum_{j
\notin \Stau} |\thetastar_j|^\qpar \; \leq \mutrun^{1-\qpar} \radq.
\end{align}
On the other hand, again using the inclusion $\thetastar \in
\Ball_\qpar(\radq)$, we have $\radq \geq \sum_{j \in \Stau}
|\thetastar_j|^\qpar \; \geq \; |\Stau| \, \mutrun^\qpar$ which
implies that $|\Stau| \leq \mutrun^{-\qpar} \radq$.  By combining this
bound with inequality~\eqref{EqnTrunUpper}, we obtain the upper bound
\begin{align*}
\epsilon^2(\DeltaStar; \ModelSet, \ModelSetPerp) & \leq
\frac{\PLAINCON \, \rhovec(\CovMat) }{\lammax(\CovMat)} \, \frac{\log
  \pdim}{\numobs} \big(\mutrun^{2-2\qpar} \radq^2 + \mutrun^{-\qpar}
\radq \|\DeltaStar\|_2^2 \big) \; = \; \frac{\PLAINCON \,
  \rhovec(\CovMat) }{\lammax(\CovMat)} \, \frac{\log \pdim}{\numobs}
\mutrun^{-\qpar} \radq \big(\mutrun^{2-\qpar} \radq +
\|\DeltaStar\|_2^2 \big).
\end{align*}
Setting $\mutrun^2 = \, \frac{\log \pdim}{\numobs}$
then yields
\begin{align*}
\epsilon^2(\DeltaStar; \ModelSet, \ModelSetPerp) & \leq
\phin(\CovMat) \; \biggr \{ \radq \big( \frac{\log
  \pdim}{\numobs} \big)^{1-\qpar/2} + \|\DeltaStar\|_2^2 \biggr \},
\qquad \mbox{where $\phin(\CovMat) \defn \frac{\PLAINCON
    \rhovec(\CovMat)}{\lammax(\CovMat)} \radq \big( \frac{\log
    \pdim}{\numobs} \big)^{1-\qpar/2}$.}
\end{align*}

Finally, let us verify the stated form of the contraction coefficient.
For the given subspace $\Bset^\perp = \ModelSet(\Stau)$ and choice of
$\mutrun$, we have $\Compat^2(\Bset^\perp) = |\Stau| \leq \mutrun^{-\qpar}
\radq$.  From Lemma~\ref{LemGarvesh}, we have
\begin{align*}
16 \Compat^2(\Bset^\perp) \frac{\FUNLOW(\LossN) +
  \FUNUP(\LossN)}{\Lsmooth} & \leq \phin(\CovMat),
\end{align*}
and hence, by definition~\eqref{EqnDefnContrac} of the contraction
coefficient,
\begin{align*}
\contrac & \leq \Big \{ 1 - \frac{\lossrsc}{2 \Lsmooth} +
\phin(\CovMat) \Big \} \; \Big \{ 1 -
\phin(\CovMat) \Big \}^{-1}.
\end{align*}

For proving Corollary~\ref{CorSparseLag}, we observe that the stated
settings $\uplossrsc$, $\phin(\CovMat)$ and $\contrac$ follow directly
from Lemma~\ref{LemGarvesh}. The bound for
condition~\ref{ThmMainLag}(a) follows from a standard argument about
the suprema of $\pdim$ independent Gaussians with variance
$\noisevar$.

\subsection{Proof of Corollary~\ref{CorLowRank}}

This proof is analogous to that of Corollary~\ref{CorSparse}, but
appropriately adapted to the matrix setting.  We first state a lemma
that allows us to establish appropriate forms of the RSC/RSM
conditions.  Recall that we are studying an instance of matrix
regression with random design, where the vectorized form $\vect(X)$ of
each matrix is drawn from a $N(0, \CovMat)$ distribution, where
$\CovMat \in \real^{\mdim^2 \times \mdim^2}$ is some covariance
matrix.  In order to state this result, let us define the quantity
\begin{align}
\label{EqnDefnRhoMat}
\rhomat(\CovMat) & \defn \sup_{\|u\|_2 = 1, \; \|v\|_2 = 1} \var(u^T X
v), \quad \mbox{where $\vect(X) \sim N(0, \CovMat)$.}
\end{align}
\blems
\label{LemSahand}
Under the conditions of Corollary~\ref{CorLowRank}, there are
universal positive constants $(\PLAINCON_0, \PLAINCON_1)$ such that
\begin{subequations}
\begin{align}
\label{EqnOrdMatRSC}
\frac{ \|\XopN(\Delta) \|_2^2}{\numobs} & \geq \; \frac{1}{2} \;
\lammin(\CovMat) \, \frob{\Delta}^2 - \PLAINCON_1 \rhomat(\CovMat)
\frac{\mdim}{\numobs} \; \nuclear{\Delta}^2, \qquad \mbox{and} \\
\label{EqnOrdMatRSM}
\frac{ \|\XopN(\Delta) \|_2^2}{\numobs} & \leq \; 2 \,
\lammax(\CovMat) \; \frob{\Delta}^2 - \PLAINCON_1 \; \rhomat(\CovMat)
\; \frac{\mdim}{\numobs} \; \nuclear{\Delta}^2, \qquad \mbox{for all
$\Delta \in \real^{\mdim \times \mdim}$.}
\end{align}
\end{subequations}
with probability at least $1 - \exp(-\PLAINCON_0 \, \numobs)$.
\elems
\noindent Given the quadratic nature of the least-squares loss, the
bound~\eqref{EqnOrdMatRSC} implies that the RSC condition holds with
$\lossrsc = \frac{1}{2} \lammin(\CovMat)$ and \mbox{$\FUNLOW(\LossN) =
  \PLAINCON_1 \rhomat(\CovMat) \frac{\mdim}{\numobs}$,} whereas the
bound~\eqref{EqnOrdMatRSM} implies that the RSM condition holds with
$\Lsmooth = 2 \lammax(\CovMat)$ and \mbox{$\FUNUP(\LossN) =
  \PLAINCON_1 \rhomat(\CovMat) \frac{\mdim}{\numobs}$.}

We now prove Corollary~\ref{CorLowRank} in the special case of exactly
low rank matrices ($\qpar = 0$), in which $\ThetaStar$ has some rank
$\rdim \leq \mdim$.  Given the singular value decomposition
$\ThetaStar = U D V^T$, let $U^\rdim$ and $V^\rdim$ be the $\mdim
\times \rdim$ matrices whose columns correspond to the $\rdim$
non-zero (left and right, respectively) singular vectors of
$\ThetaStar$.  As in Section~\ref{SecMatrixReg}, define the subspace of
matrices
\begin{align}
\label{EqnBefore}
\ModelSet(U^r, V^r) & \defn \big \{ \Theta \in \real^{\mdim \times
\mdim} \, \mid \, \col(\Theta) \subseteq U^r \mbox{ and } \row(\Theta)
\subseteq V^r \big \},
\end{align}
as well as the associated set $\BsetPerp(U^r, V^r)$.  Note that
$\ThetaStar \in \ModelSet$ by construction, and moreover (as discussed
in Section~\ref{SecMatrixReg}, the nuclear norm is decomposable with
respect to the pair $(\ModelSet, \BsetPerp)$.

By definition~\eqref{EqnDefnSubspaceCompat} of the subspace
compatibility with nuclear norm as the regularizer and Frobenius norm
as the error norm, we have $\Compat^2(\ModelSet) = \rdim$.  Using the
settings of $\FUNLOW(\LossN)$ and $\FUNUP(\LossN)$ guaranteed by
Lemma~\ref{LemSahand} and substituting into
equation~\eqref{EqnDefnContrac}, we obtain a contraction coefficient
\begin{align}
\contrac(\CovMat) & \defn \Big \{ 1 - \frac{\lammin(\CovMat)}{4
  \lammax(\CovMat)} + \phin(\CovMat) \Big \} \; \Big \{ 1 -
\phin(\CovMat) \Big \}^{-1},
\end{align}
where $\phin(\CovMat) \defn \frac{\PLAINCON_2
  \rhomat(\CovMat)}{\lammax(\CovMat)} \; \frac{\rdim \mdim}{\numobs}$
  for some universal constant $\PLAINCON_2$.  A similar calculation
  shows that the tolerance term takes the form
\begin{align*}
\HACKSTATERR & \leq \PLAINCON_3 \; \phin(\CovMat) \Big \{
\frac{\nuclear{\DeltaStar}^2}{\rdim} + \frob{\DeltaStar}^2 \Big \}
\qquad \mbox{for some constant $\PLAINCON_3$.}
\end{align*}
Since $\radcon \leq \nuclear{\ThetaStar}$ by assumption,
 Lemma~\ref{LemConeOpt} (as exploited in the proof of
 Corollary~\ref{CorGeneric}) shows that $\nuclear{\DeltaStar}^2 \leq 4
 \rdim \frob{\DeltaStar}^2$, and hence that
\begin{align*}
\HACKSTATERR & \leq \PLAINCON_3 \; \phin(\CovMat) \;
\frob{\DeltaStar}^2,
\end{align*}
which show the claim~\eqref{EqnMatLowRankZero} for $\qpar = 0$.

We now turn to the case $\qpar \in (0,1]$; as in the proof of this
case for Corollary~\ref{CorSparse}, we bound $\HACKSTATERR$ using a
slightly different choice of the subspace pair.  Recall our notation
$\singval_1(\ThetaStar) \geq \singval_2(\ThetaStar) \geq \cdots \geq
\singval_\mdim(\ThetaStar) \geq 0$ for the ordered singular values of
$\ThetaStar$.  For a threshold $\mutrun$ to be chosen, define $\Stau =
\big \{ j \in \{1,2, \ldots, \pdim \} \, \mid \, \singval_j(\ThetaStar)
> \mutrun\big \}$, and $U(\Stau) \in \real^{\mdim \times |\Stau|}$ be the
matrix of left singular vectors indexed by $\Stau$, with the matrix
$V(\Stau)$ defined similarly.  We then define the subspace
$\ModelSet(\Stau) \defn \ModelSet(U(\Stau), V(\Stau))$ in an analogous
fashion to equation~\eqref{EqnBefore}, as well as the subspace
$\BsetPerp(\Stau)$.

Now by a combination of Lemma~\ref{LemConeOpt} and the
definition~\eqref{EqnDefnHackStatErr} of $\HACKSTATERR$, for any pair
$(\ModelSet(\Stau), \BsetPerp(\Stau))$, we have
\begin{align*}
\epsilon^2(\DeltaStar; \ModelSet, \BsetPerp) & \leq \frac{\PLAINCON \,
\rhomat(\CovMat) }{\lammax(\CovMat)} \, \frac{\mdim}{\numobs} \big(
\sum_{j \notin \Stau} \singval_j(\ThetaStar) + \sqrt{|\Stau|} \,
\frob{\DeltaStar} \big)^2,
\end{align*}
where to simplify notation, we have omitted the dependence of
$\ModelSet$ and $\ModelSetPerp$ on $\Stau$.  As in the proof of
Corollary~\ref{CorSparse}, we now choose the threshold $\mutrun$
optimally, so as to trade-off the term $\sum_{j \notin \Stau}
\singval_j(\ThetaStar)$ with its competitor $\sqrt{|\Stau|} \,
\frob{\DeltaStar}$.  Exploiting the fact that $\ThetaStar \in
\Ball_\qpar(\radq)$ and following the same steps as the proof of
Corollary~\ref{CorSparse} yields the bound
\begin{align*}
\epsilon^2(\DeltaStar; \ModelSet, \BsetPerp) & \leq
 \frac{\PLAINCON
\, \rhomat(\CovMat) }{\lammax(\CovMat)} \, \frac{\mdim}{\numobs}
\big(\mutrun^{2-2\qpar} \radq^2 + \mutrun^{-\qpar} \radq \frob{\DeltaStar}^2
\big).
\end{align*}
Setting $\mutrun^2 =   \frac{\mdim}{\numobs}$ then yields
\begin{align*}
\epsilon^2(\DeltaStar; \ModelSet, \BsetPerp) & \leq
 \phin(\CovMat) \; \biggr \{ \radq \big( 
 \frac{\mdim}{\numobs} \big)^{1-\qpar/2} + \frob{\DeltaStar}^2 \biggr
 \},
\end{align*}
as claimed.  The stated form of the contraction coefficient can be
verified by a calculation analogous to the proof of
Corollary~\ref{CorSparse}.


\subsection{Proof of Corollary~\ref{CorMatComp}}

In this case, we let $\Xop_\numobs: \real^{\mdim \times \mdim}
\rightarrow \real^\numobs$ be the operator defined by the model of
random signed matrix sampling~\cite{NegWai10b}.  As previously
argued, establishing the RSM/RSC property amounts to obtaining a form
of uniform control over $\frac{\|\XopN(\Theta)\|_2^2}{\numobs}$.  More
specifically, from the proof of Theorem~\ref{ThmMainCon}, we see that
it suffices to have a form of RSC for the difference $\DelHatIt{t} =
\Iter{t} - \ThetaHat$, and a form of RSM for the difference
$\Iter{t+1} - \Iter{t}$.  The following two lemmas summarize these
claims:
\blems
\label{LemMCRSC}
There is a constant $\PLAINCON$ such that for all iterations $t =
0,1,2, \ldots$ and integers \mbox{$\rdim = 1, 2, \ldots, \mdim-1$,}
with probability at least $1 - \exp(-\mdim \log \mdim)$,
\begin{align}
\label{EqnMCRSC}
\frac{\|\XopN(\DelHatIt{t})\|_2^2}{\numobs} & \geq \frac{1}{2}
\frob{\DelHatIt{t}}^2 - \underbrace{\PLAINCON \spike \sqrt{\frac{\rdim
      \, \mdim \log \mdim}{\numobs}} \Big \{ \frac{ \sum_{j=r+1}^\mdim
    \singval_j(\ThetaStar)}{\sqrt{\rdim}} + \spike \sqrt{\frac{\rdim
      \mdim \log \mdim}{\numobs}} + \frob{\DeltaStar} \Big
  \}}_{\DEL_\ell(r)}.
\end{align}
\elems

\blems
\label{LemMCRSM}
There is a constant $\PLAINCON$ such that for all iterations $t =
0,1,2, \ldots$ and integers \mbox{$\rdim = 1, 2, \ldots, \mdim-1$,}
with probability at least $1 - \exp(-\mdim \log \mdim)$, the
difference $\SHORT \defn \Iter{t+1} - \Iter{t}$ satisfies the
inequality $\frac{\|\XopN(\SHORT)\|_2^2}{\numobs} \leq 2
\frob{\SHORT}^2 + \DEL_u(r)$, where
\begin{align*}
\DEL_u(r) & \defn \PLAINCON \spike \sqrt{\frac{\rdim \mdim \log
    \mdim}{\numobs}} \Big \{ \frac{\sum_{j=r+1}^\mdim
  \singval_j(\ThetaStar)}{\sqrt{\rdim}} + \spike \sqrt{\frac{\rdim
    \mdim \log \mdim}{\numobs}} + \frob{\DeltaStar} +
\frob{\DelHatIt{t}} + \frob{\DelHatIt{t+1}} \Big \}.
\end{align*}
\elems

\vtiny

We can now complete the proof of Corollary~\ref{CorMatComp} by a minor
modification of the proof of Theorem~\ref{ThmMainCon}.  Recalling the
elementary relation~\eqref{EqnElementary}, we have
\begin{align*}
\frob{\Iter{t+1} - \ThetaHat}^2 & = \frob{\Iter{t} - \ThetaHat}^2
+ \frob{\Iter{t} - \Iter{t+1}}^2 - 2 \tracer{\Iter{t} -
  \ThetaHat}{\Iter{t} - \Iter{t+1}}.
\end{align*}
From the proof of Lemma~\ref{LemResGrad}, we see that the combination
of Lemma~\ref{LemMCRSC} and~\ref{LemMCRSM} (with $\lossrsc =
\frac{1}{2}$ and $\Lsmooth = 2$) imply that
\begin{align*}
2 \tracer{\Iter{t} - \Iter{t+1}}{\Iter{t} - \ThetaHat} & \geq
\frob{\Iter{t} - \Iter{t+1}}^2 + \frac{1}{4} \frob{\Iter{t} -
\ThetaHat}^2 - \DEL_u(r) - \DEL_\ell(r)
\end{align*}
and hence that
\begin{align*}
\frob{\DelHatIt{t+1}}^2 & \leq \frac{3}{4} \frob{\DelHatIt{t}}^2 +
\DEL_\ell(r) + \DEL_u(r).
\end{align*}
We substitute the forms of $\DEL_\ell(r)$ and $\DEL_u(r)$ given in
Lemmas~\ref{LemMCRSC} and~\ref{LemMCRSM} respectively; performing some
algebra then yields
\begin{align*}
\Big \{ 1 - \frac{\PLAINCON \, \spike \sqrt{\frac{ \rdim \mdim \log
      \mdim}{\numobs}}}{\frob{\DelHatIt{t+1}}} \Big \}
\frob{\DelHatIt{t+1}}^2 & \leq \Big \{ \frac{3}{4} + \frac{\PLAINCON
  \spike \, \sqrt{\frac{\rdim \mdim \log
      \mdim}{\numobs}}}{\frob{\DelHatIt{t}}} \Big \}
\frob{\DelHatIt{t}}^2 + \PLAINCON' \; \DEL_\ell(r).
\end{align*}
Consequently, as long as $\min \{ \frob{\DelHatIt{t}}^2, \;
\frob{\DelHatIt{t+1}}^2 \} \geq \PLAINCON_3 \spike \frac{\rdim \mdim
  \log \mdim}{\numobs}$ for a sufficiently large constant
$\PLAINCON_3$, we are guaranteed the existence of some $\contrac_t \in
(0,1)$ decreasing with $t$ such that
\begin{align}
\label{EqnMCContrac}
\frob{\DelHatIt{t+1}}^2 & \leq \contrac \frob{\DelHatIt{t}}^2 +
\PLAINCON' \DEL_\ell(r).
\end{align}
Since $\DEL_\ell(r) = \Omega( \frac{\rdim \mdim \log
  \mdim}{\numobs})$, this inequality~\eqref{EqnMCContrac} is valid for
  all $t = 0, 1, 2, \ldots$ as long as $\PLAINCON'$ is sufficiently
  large. Now iterating this bound, we see that 

\begin{align*}
\frob{\DelHatIt{t+1}}^2 & \leq \bigg(\prod_{s=1}^t\contrac_s\bigg)
\frob{\DelHatIt{0}}^2 + \PLAINCON' \, \DEL_\ell(r) \,\bigg(
\contrac_{t} + \contrac_{t}\contrac_{t-1} + \dots + \prod_{s=2}^t
\contrac_s \bigg).
\end{align*}
Since $\contrac_t$ is decreasing in $t$, we observe that the second
term in the above bound is at most 

\begin{equation*}
\PLAINCON' \, \DEL_\ell(r) \,\bigg( \contrac_{t} +
\contrac_{t}\contrac_{t-1} + \dots + \prod_{s=2}^t \contrac_s \bigg)
\leq \PLAINCON' \, \DEL_\ell(r) \,\bigg( \contrac_1 + \contrac_1^2 +
\contrac_1^{t-1} \bigg) \leq \PLAINCON' \, \frac{\DEL_\ell(r)}{1 -
  \contrac_1}.
\end{equation*}
We also define $\barcontrac_t = (\sum_{s=1}^t \contrac_t)/t$. Then the
arithmetic mean-geometric mean inequality yields the upper bound
$\prod_{s=1}^t\contrac_s \leq \barcontrac_t^t$. Combining this with
our earlier upper bound further yields the inequality
\begin{align}
\label{EqnBraga}
\frob{\DelHatIt{t+1}}^2 & \leq \barcontrac_t^t \frob{\DelHatIt{0}}^2 +
\frac{\PLAINCON'}{1-\contrac_1} \; \DEL_\ell(r).
\end{align}

It remains to choose the cut-off $r \in \{1, 2, \ldots, \mdim -1 \}$
so as to minimize the term $\DEL_\ell(r)$.  In particular, when
$\ThetaStar \in \Ball_\qpar(\radq)$, then as shown in the
paper~\cite{NegWai09}, the optimal choice is $\rdim \asymp
\spike^{-\qpar}\radq \big(\frac{\numobs}{\mdim \log \mdim}
\big)^{\qpar/2}$.  Substituting into the inequality~\eqref{EqnBraga}
and performing some algebra yields that there is a universal constant
$\PLAINCON_4$ such that the bound
\begin{align*}
\frob{\DelHatIt{t+1}}^2 & \leq \contrac^t \frob{\DelHatIt{0}}^2 +
\frac{\PLAINCON_4}{1-\contrac} \Big \{ \radq \big(\frac{\spike \mdim
  \log \mdim}{\numobs} \big)^{1-\qpar/2} + \sqrt{\radq
  \big(\frac{\spike \mdim \log \mdim}{\numobs} \big)^{1-\qpar/2}} \;
\frob{\DeltaStar} \Big \}.
\end{align*}
holds. Now by the Cauchy-Schwarz inequality we have
\begin{align*}
\sqrt{\radq \big(\frac{\spike \mdim \log \mdim}{\numobs}
  \big)^{1-\qpar/2}} \; \frob{\DeltaStar} & \leq \frac{1}{2} \radq
\big(\frac{\spike\mdim \log \mdim}{\numobs} \big)^{1-\qpar/2} +
\frac{1}{2} \frob{\DeltaStar}^2,
\end{align*}
and the claimed inequality~\eqref{EqnMatComp} follows.

\subsection{Proof of Corollary~\ref{CorMatDecomp}}

Again the main argument in the proof would be to establish the RSM and
RSC properties for the decomposition problem. We define
$\DelHatIt{t}_\Theta = \Iter{t} - \ThetaHat$ and $\DelHatIt{t}_\Gamma =
\Gamma^t - \GammaHat$. We start with giving a lemma that establishes
RSC for the differences $(\DelHatIt{t}_\Theta, \DelHatIt{t}_\Gamma)$. We
recall that just like noted in the previous section, it suffices to
show RSC only for these differences. Showing RSC/RSM in this example
amounts to analyzing $\frob{\DelHatIt{t}_\Theta +
  \DelHatIt{t}_\Gamma}^2$. We recall that this section assumes that
$\GammaStar$ has only $\spindex$ non-zero columns.

\blems
\label{LemMatDecompRSC}
There is a constant $\PLAINCON$ such that for all iterations $t = 0,
1, 2,\dots$,
\begin{equation}
\frob{\DelHatIt{t}_\Theta + \DelHatIt{t}_\Gamma}^2 \geq
\frac{1}{2}\big(\frob{\DelHatIt{t}_\Theta}^2 +
\frob{\DelHatIt{t}_\Gamma}^2\big) -
\PLAINCON\spike\sqrt{\frac{\spindex}{\mdimb}}\biggr(\frob{\GammaHat -
  \GammaStar} + \spike\sqrt{\frac{\spindex}{\mdimb}}\biggr)
\label{eqn:matdecomprsc}
\end{equation}
\elems 
\noindent This proof of this lemma follows by a straightforward
modification of analogous results in the
paper~\cite{AgarwalNegWai11}.\\

Matrix decomposition has the interesting property that the RSC
condition holds in a deterministic sense (as opposed to with high
probability).  The same deterministic guarantee holds for the RSM
condition; indeed, we have
\begin{align}
  \frob{\DelHatIt{t}_\Delta + \DelHatIt{t}_\Gamma}^2 \leq
  2\big(\frob{\DelHatIt{t}_\Theta}^2 + \frob{\DelHatIt{t}_\Gamma}^2\big),
  \label{eqn:matdecomprsm}
\end{align}
by Cauchy-Schwartz inequality. Now we appeal to the more general form
of Theorem~\ref{ThmMainCon} as stated in
Equation~\ref{eqn:thmmaingeneral}, which gives
\begin{equation*}
\frob{\DelHatIt{t+1}_\Theta}^2 + \frob{\DelHatIt{t+1}_\Gamma}^2 \leq
\left(\frac{3}{4}\right)^t\big( \frob{\DelHatIt{0}_\Theta}^2 +
\frob{\DelHatIt{0}_\Gamma}^2 \big) +
\PLAINCON\sqrt{\frac{\spike\spindex}{\mdimb}}\biggr(\frob{\GammaHat - 
  \GammaStar} + \frac{\spike\spindex}{\mdimb}\biggr).
\end{equation*}
The stated form of the corollary follows by an application of
Cauchy-Schwarz inequality.


\section{Discussion}

In this paper, we have shown that even though high-dimensional
$M$-estimators in statistics are neither strongly convex nor smooth,
simple first-order methods can still enjoy global guarantees of
geometric convergence.  The key insight is that strong convexity and
smoothness need only hold in restricted senses, and moreover, these
conditions are satisfied with high probability for many statistical
models and decomposable regularizers used in practice.  Examples
include sparse linear regression and $\ell_1$-regularization, various
statistical models with group-sparse regularization, matrix regression
with nuclear norm constraints (including matrix completion and
multi-task learning), and matrix decomposition problems.  Overall, our
results highlight some important connections between computation and
statistics: the properties of $M$-estimators favorable for fast rates
in a statistical sense can also be used to establish fast rates for
optimization algorithms.

\paragraph{Acknowledgements:}   
All three authors were partially supported by grants AFOSR-09NL184; in
addition, AA was partially supported by a Microsoft Graduate
Fellowship and Google PhD Fellowship, and SN and MJW acknowledge
funding from NSF-CDI-0941742.  We would like to thank the anonymous
reviewers and associate editor for their helpful comments that helped
to improve the paper, and Bin Yu for inspiring discussions on the
interaction between statistical and optimization error.


\appendix


\section{Auxiliary results for Theorem~\protect \robrefthmcon}

In this appendix, we provide the proofs of various auxiliary lemmas
required in the proof of Theorem~\ref{ThmMainCon}.

\subsection{Proof of Lemma~\ref{LemMUFCvsMCBoring}}
\label{AppLemMUFCvsMCBoring}

Since $\iter{t}$ and $\thetahat$ are both feasible and $\thetahat$
lies on the constraint boundary, we have $\Reg{\iter{t}} \leq
\Reg{\thetahat}$.  Since $\Reg{\thetahat} \leq \Reg{\thetastar} +
\Reg{\thetahat - \thetastar}$ by triangle inequality, we conclude that
\begin{align*}
\Reg{\iter{t}} & \leq \Reg{\thetastar} + \Reg{\DeltaStar}.
\end{align*}
Since $\thetastar = \BigProjModel{\thetastar} +
\BigProjModelPerp{\thetastar}$, a second application of triangle
inequality yields
\begin{align}
\label{EqnRedOne}
\Reg{\iter{t}} & \leq \Reg{\BigProjModel{\thetastar}} +
\Reg{\BigProjModelPerp{\thetastar}} + \Reg{\DeltaStar}.
\end{align}
Now define the difference $\DelIt{t} \defn \iter{t} - \thetastar$.
(Note that this is slightly different from $\DelHatIt{t}$, which is
measured relative to the optimum $\thetahat$.)  With this notation, we
have
\begin{align*}
\Reg{\iter{t}} & = \Regplain \big( \BigProjModel{\thetastar} +
\BigProjModelPerp{\thetastar} + \BigProjBset{\DelIt{t}} +
\BigProjBsetPerp{\DelIt{t}} \big) \\
& \stackrel{(i)}{\geq} \Regplain \big( \BigProjModel{\thetastar} +
\BigProjBsetPerp{\DelIt{t}} \big) - \Regplain \big(
\BigProjModelPerp{\thetastar} + \BigProjBset{\DelIt{t}}\big) \\
& \stackrel{(ii)}{\geq} \Regplain \big( \BigProjModel{\thetastar} +
\BigProjBsetPerp{\DelIt{t}} \big) -
\Reg{\BigProjModelPerp{\thetastar}} - \Reg{ \BigProjBset{\DelIt{t}}},
\end{align*}
where steps (i) and (ii) each use the triangle inequality.  Now by the
decomposability condition, we have $\Regplain \big(
\BigProjModel{\thetastar} + \BigProjBsetPerp{\DelIt{t}} \big) =
\Regplain ( \BigProjModel{\thetastar}) +
\Regplain(\BigProjBsetPerp{\DelIt{t}})$, so that we have shown that
\begin{align*}
 \Reg { \BigProjModel{\thetastar}} + \Reg{\BigProjBsetPerp{\DelIt{t}}}
 - \Reg{\BigProjModelPerp{\thetastar}} - \Reg{\BigProjBset{\DelIt{t}}}
 & \leq \Reg{\iter{t}}.
\end{align*}
Combining this inequality with the earlier bound~\eqref{EqnRedOne}
yields
\begin{align*}
 \Reg{\BigProjModel{\thetastar}} + \Reg{\BigProjBsetPerp{\DelIt{t}}} -
\Reg{\BigProjModelPerp{\thetastar}} - \Reg{ \BigProjBset{\DelIt{t}}} &
\leq \Reg{\BigProjModel{\thetastar}} +
\Reg{\BigProjModelPerp{\thetastar}} + \Reg{\DeltaStar}.
\end{align*}
Re-arranging yields the inequality
\begin{align}
\label{EqnRedClaim}
\Reg{\BigProjBsetPerp{\DelIt{t}}} & \leq \Reg{
 \BigProjBset{\DelIt{t}}} + 2 \Reg{\BigProjModelPerp{\thetastar}} +
 \Reg{\DeltaStar}.
\end{align}

The final step is to translate this inequality into one that applies
to the optimization error $\DelHatIt{t} = \iter{t} - \thetahat$.
Recalling that $\DeltaStar = \thetahat - \thetastar$, we have
$\DelHatIt{t} = \DelIt{t} - \DeltaStar$, and hence
\begin{align}
\label{EqnTri}
\Reg{\DelHatIt{t}} & \leq \Reg{\DelIt{t}} + \Reg{\DeltaStar}, \qquad
\mbox{by triangle inequality.}
\end{align}
In addition, we have
\begin{align*}
\Reg{\DelIt{t}} \; \leq \; \Reg{\BigProjBsetPerp{\DelIt{t}}} + \Reg{
  \BigProjBset{\DelIt{t}}} & \; \stackrel{(i)}{\leq} \; 2 \,
\Reg{\BigProjBset{\DelIt{t}}} + 2 \Reg{\BigProjModelPerp{\thetastar}}
+ \Reg{\DeltaStar} \\
& \stackrel{(ii)}{\leq} 2 \, \Compat(\Bset^\perp)
\norm{\BigProjBset{\DelIt{t}}} + 2 \Reg{\BigProjModelPerp{\thetastar}}
+ \Reg{\DeltaStar},
\end{align*}
where inequality (i) uses the bound~\eqref{EqnRedClaim}, and
inequality (ii) uses the definition~\eqref{EqnDefnSubspaceCompat} of
the subspace compatibility $\Compat$.  Combining with the
inequality~\eqref{EqnTri} yields
\begin{align*}
\Reg{\DelHatIt{t}} & \leq 2 \, \Compat(\Bset^\perp)
\norm{\BigProjBset{\DelIt{t}}} + 2 \Reg{\BigProjModelPerp{\thetastar}}
+ 2 \Reg{\DeltaStar}.
\end{align*}
Since projection onto a subspace is non-expansive, we have
$\norm{\BigProjBset{\DelIt{t}}}
  \leq \norm{\DelIt{t}}$, and hence
\begin{align*}
\norm{\BigProjBset{\DelIt{t}}} & \leq \norm{\DelHatIt{t} + \DeltaStar}
\; \leq \; \norm{\DelHatIt{t}} + \norm{\DeltaStar}.
\end{align*}
Combining the pieces, we obtain the claim~\eqref{EqnMUFCvsMCBoring}.
%


\subsection{Proof of Lemma~\ref{LemResGrad}}
\label{AppLemResGrad}

We start by applying the RSC assumption to the pair $\thetahat$ and
$\iter{t}$, thereby obtaining the lower bound
\begin{align}
  \LossN(\betahat) - \frac{\lossrsc}{2}\plainnorm{\betahat -
    \iter{t}}^2 &\geq \LossN(\iter{t}) + \inprod{\grad
    \LossN(\iter{t})}{\betahat - \iter{t}} - \FUNLOW(\LossN)
    \RegSq{\iter{t} - \thetahat} \nonumber \\
\label{EqnRSCTWO}
& = \LossN(\iter{t}) + \inprod{\grad \LossN(\iter{t})}{\iter{t+1} -
    \iter{t}} + \inprod{\grad \LossN(\iter{t})}{\betahat - \iter{t+1}}
    - \FUNLOW(\LossN) \RegSq{\iter{t} - \thetahat}.
\end{align}
Here the second inequality follows by adding and subtracting terms.

Now for compactness in notation, define $\NEWPHI_t(\thetapar) \defn
\LossN(\iter{t}) + \ip{\grad \LossN(\iter{t})}{\thetapar - \iter{t}} +
\frac{\Lsmooth}{2}\plainnorm{\thetapar - \iter{t}}^2$, and note that
by definition of the algorithm, the iterate $\iter{t+1}$ minimizes
$\NEWPHI_t(\thetapar)$ over the ball $\Ball_\Plainreg(\radcon)$.
Moreover, since $\betahat$ is feasible, the first-order conditions for
optimality imply that \mbox{$\inprod{\grad
    \NEWPHI_t(\iter{t+1})}{\betahat - \iter{t+1}} \geq 0$,} or
equivalently that $\inprod{\grad \LossN(\iter{t}) + \Lsmooth
  (\iter{t+1} - \iter{t})}{\betahat - \iter{t+1}} \; \geq \; 0$.
Applying this inequality to the lower bound~\eqref{EqnRSCTWO}, we find
that
\begin{align}
  \LossN(\betahat) - \frac{\lossrsc}{2}\plainnorm{\betahat -
    \iter{t}}^2 & \geq \LossN(\iter{t}) + \inprod{\grad
    \LossN(\iter{t})}{\iter{t+1} - \iter{t}} + \Lsmooth
    \inprod{\iter{t} - \iter{t+1}}{\betahat - \iter{t+1}} -
    \FUNLOW(\LossN) \RegSq{\iter{t}- \thetahat} \nonumber \\
 &= \NEWPHI_t(\iter{t+1}) - \frac{\Lsmooth}{2}\plainnorm{\iter{t+1} -
    \iter{t}}^2 + \Lsmooth \inprod{\iter{t} - \iter{t+1}}{\betahat -
    \iter{t+1}} - \FUNLOW(\LossN) \RegSq{\iter{t} - \thetahat} \nonumber \\
\label{EqnNearFinal}
& = \NEWPHI_t(\iter{t+1}) + \frac{\Lsmooth}{2}\plainnorm{\iter{t+1} -
  \iter{t}}^2 + \Lsmooth \inprod{\iter{t} - \iter{t+1}}{\betahat -
  \iter{t}} - \FUNLOW(\LossN) \RegSq{\iter{t} - \thetahat},
\end{align}
where the last step follows from adding and subtracting $\iter{t+1}$
in the inner product. 

Now by the RSM condition, we have
\begin{align}
\label{EqnRSMFirst}
\NEWPHI_t(\iter{t+1}) \geq \LossN(\iter{t+1}) - \FUNUP(\LossN)
\RegSq{\iter{t+1} - \iter{t}} \stackrel{(a)}{\geq} \LossN(\betahat) -
\FUNUP(\LossN) \RegSq{\iter{t+1} - \iter{t}},
\end{align}
where inequality (a) follows by the optimality of $\betahat$, and
feasibility of $\iter{t+1}$.  Combining this inequality with the
previous bound~\eqref{EqnNearFinal} yields that $\LossN(\betahat) -
\frac{\lossrsc}{2} \plainnorm{\betahat - \iter{t}}^2$ is lower bounded
by
\begin{align*}
\LossN(\betahat) - \frac{\Lsmooth}{2} \plainnorm{\iter{t+1} -
\iter{t}}^2 + \Lsmooth \inprod{\iter{t} - \iter{t+1}}{\betahat -
\iter{t}} - \FUNLOW(\LossN) \, \RegSq{\iter{t} - \thetahat} -
\FUNUP(\LossN) \RegSq{\iter{t+1} -\iter{t}},
\end{align*}
and the claim~\eqref{EqnResGrad} follows after some simple
algebraic manipulations.
%


\section{Auxiliary results for Theorem~\robrefthmlag}
In this appendix, we prove the two auxiliary lemmas required in the
proof of Theorem~\ref{ThmMainLag}.

\subsection{Proof of Lemma~\ref{LemICB}}

This result is a generalization of an analogous result in Negahban et
al.~\cite{NegRavWaiYu09}, with some changes required so as to adapt
the statement to the optimization setting.  Let $\thetapar$ be any
vector, feasible for the problem~\eqref{EqnMestReg}, that satisfies
the bound
\begin{align}
\label{EqnFrenchTGV}
\obj(\thetapar) & \leq \obj(\thetastar) + \objerror,
\end{align}
and assume that $\regpar \geq 2 \Plainregdual(\nabla
\LossN(\thetastar))$.  We then claim that the error vector $\Delta
\defn \thetapar - \thetastar$ satisfies the inequality
\begin{align}
\label{EqnExtendedCone}
  \Reg{\BigProjBsetPerp{\Delta}} \; \leq \; 3
  \Reg{\BigProjBset{\Delta}} + 4 \Reg{\BigProjModelPerp{\thetastar}} +
  2 \min \big \{ \frac{\objerror}{\regpar}, \radlag \big \}.
\end{align}
For the moment, we take this claim as given, returning later to
verify its validity.

By applying this intermediate claim~\eqref{EqnExtendedCone} in two
different ways, we can complete the proof of Lemma~\ref{LemICB}.
First, we observe that when $\thetapar = \thetahat$, the optimality of
$\thetahat$ and feasibility of $\thetastar$ imply that
assumption~\eqref{EqnFrenchTGV} holds with $\objerror = 0$, and hence
the intermediate claim~\eqref{EqnExtendedCone} implies that the
statistical error $\DeltaStar = \thetastar - \thetahat$ satisfies the
bound
\begin{align}
\label{EqnWhiteOne}
  \Reg{\BigProjBsetPerp{\DeltaStar}} \; \leq \; 3
  \Reg{\BigProjBset{\DeltaStar}} + 4
  \Reg{\BigProjModelPerp{\thetastar}}.
\end{align}
Since $\DeltaStar = \BigProjBset{\DeltaStar} +
\BigProjBsetPerp{\DeltaStar}$, we can write
\begin{align}
\label{EqnWhite}
\Reg{\DeltaStar} & = \Reg{\BigProjBset{\DeltaStar} +
  \BigProjBsetPerp{\DeltaStar}} \; \leq \; 4
\Reg{\BigProjBset{\DeltaStar}} + 4
\Reg{\BigProjModelPerp{\thetastar}},
\end{align}
using the triangle inequality in conjunction with our earlier
bound~\eqref{EqnWhiteOne}.  Similarly, when $\thetapar = \iter{t}$ for
some $t \geq T$, then the given assumptions imply that
condition~\eqref{EqnFrenchTGV} holds with $\objerror > 0$, so that the
intermediate claim (followed by the same argument with triangle
inequality) implies that the error $\DelIt{t} = \iter{t} - \thetastar$
satisfies the bound
\begin{align}
\label{EqnRose}
  \Reg{\DelIt{t}} \; \leq \; 4 \Reg{\BigProjBset{\DelIt{t}}} + 4
  \Reg{\BigProjModelPerp{\thetastar}} + 2 \min \big \{
  \frac{\objerror}{\regpar}, \radlag \big \}.
\end{align}

Now let $\DelHatIt{t} = \iter{t} - \thetahat$ be the optimization
error at time $t$, and observe that we have the decomposition
$\DelHatIt{t} = \DelIt{t} + \DeltaStar$.  Consequently, by triangle
inequality
\begin{align}
\Reg{\DelHatIt{t}} & \leq \Reg{\DelIt{t}} + \Reg{\DeltaStar} \nonumber
\\
& \stackrel{(i)}{\leq} 4 \Big \{ \Reg{\BigProjBset{\DelIt{t}}} +
\Reg{\BigProjBset{\DeltaStar}} \Big \} + 8
\Reg{\BigProjModelPerp{\thetastar}} + 2 \min \big \{
\frac{\objerror}{\regpar}, \: \radlag \big \} \nonumber \\
& \stackrel{(ii)}{\leq} 4 \Compat(\Bset) \; \Big \{
\plainnorm{\BigProjBset{\DelIt{t}}} +
\plainnorm{\BigProjBset{\DeltaStar}} \Big \} + 8
\Reg{\BigProjModelPerp{\thetastar}} + 2 \min \big \{
\frac{\objerror}{\regpar}, \: \radlag \big \} \nonumber \\
\label{EqnAix}
& \stackrel{(iii)}{\leq} 4 \Compat(\Bset) \; \Big \{
\plainnorm{\DelIt{t}} + \plainnorm{\DeltaStar} \Big \} + 8
\Reg{\BigProjModelPerp{\thetastar}} + 2 \min \big \{
\frac{\objerror}{\regpar}, \: \radlag \big \},
\end{align}
where step (i) follows by applying both equation~\eqref{EqnWhite}
and~\eqref{EqnRose}; step (ii) follows from the
definition~\eqref{EqnDefnSubspaceCompat} of the subspace compatibility
that relates the regularizer to the norm $\plainnorm{\cdot}$; and step
(iii) follows from the fact that projection onto a subspace is
non-expansive.  Finally, since $\DelIt{t} = \DelHatIt{t} -
\DeltaStar$, the triangle inequality implies that
$\plainnorm{\DelIt{t}} \leq \plainnorm{\DelHatIt{t}} +
\plainnorm{\DeltaStar}$.  Substituting this upper bound into
inequality~\eqref{EqnAix} completes the proof of Lemma~\ref{LemICB}.\\

It remains to prove the intermediate claim~\eqref{EqnExtendedCone}.
Letting $\thetapar$ be any vector, feasible for the
program~\eqref{EqnMestReg}, and satisfying the
condition~\eqref{EqnFrenchTGV}, and let $\Delta = \thetapar -
\thetastar$ be the associated error vector.  Re-writing the
condition~\eqref{EqnFrenchTGV}, we have
\begin{equation*}
  \LossN(\thetastar + \Delta) + \regpar \Reg{\thetastar + \Delta} \;
  \leq \; \LossN(\thetastar) + \regpar \Reg{\thetastar} + \objerror.
\end{equation*}
Subtracting $\binprod{\nabla \LossN(\thetastar)}{\Delta}$ from each
side and then re-arranging yields the inequality
\begin{align*}
  \LossN(\thetastar + \Delta) - \LossN(\thetastar) - \binprod{\grad
    \LossN (\betastar)}{\Delta} + \regpar \Big \{ \Reg{\thetastar +
    \Delta} - \Reg{\thetastar} \Big \} & \leq \; -\binprod{\grad
    \LossN (\betastar)}{\Delta} + \objerror.
\end{align*}
The convexity of $\LossN$ then implies that $\LossN(\thetastar +
\Delta) - \LossN(\thetastar) - \binprod{\grad \LossN
  (\betastar)}{\Delta} \geq 0$, and hence that
\begin{equation*}
  \regpar \Big \{ \Reg{\thetastar + \Delta} - \Reg{\thetastar} \Big \}
  \; \leq \; -\binprod{\grad \LossN (\betastar)}{\Delta} + \objerror.
\end{equation*}

Applying H\"older's inequality to $\binprod{\grad \LossN
  (\betastar)}{\err}$, as expressed in terms of the dual norms
$\Regplain$ and $\Plainregdual$, yields the upper bound
\begin{align*}
  \regpar \Big \{ \Reg{\thetastar + \Delta} - \Reg{\thetastar} \Big \}
  & \leq \; \Plainregdual(\nabla \LossN(\thetastar)) \; \Reg{\Delta}
  \; + \objerror \; \stackrel{(i)}{\leq} \; \frac{\regpar}{2} \;
  \Reg{\Delta} + \objerror,
\end{align*}
where step (i) uses the fact that $\regpar \geq 2 \Regdual{\grad
  \LossN(\betastar)}$ by assumption.  

For the remainder of the proof, let us introduce the convenient
shorthand $\ProjBset{\err} \defn \BigProjBset{\err}$ and
$\ProjBsetPerp{\err} \defn \BigProjBsetPerp{\err}$, with similar
shorthand for projections involving $\thetastar$.  Making note of the
decomposition $\err = \ProjBset{\err} + \ProjBsetPerp{\err}$, an
application of triangle inequality then yields the upper bound
\begin{align}
\label{EqnMarseille}
   \Reg{\thetastar + \Delta} - \Reg{\thetastar} & \leq \; \frac{1}{2}
   \Big \{ \Reg{\ProjBset{\err}} + \Reg{\ProjBsetPerp{\err}} \Big \} +
   \frac{\objerror}{\regpar},
\end{align}
where we have rescaled both sides by $\regpar > 0$.

It remains to further lower bound the left-hand
side~\eqref{EqnMarseille}.  By triangle inequality, we have
\begin{align}
\label{EqnBread}
-\Reg{\thetastar} & \geq - \Reg{\ProjModel{\thetastar}} -
\Reg{\ProjModelPerp{\thetastar}}.
\end{align}
Let us now write $\thetastar + \Delta = \ProjModel{\thetastar} +
\ProjModelPerp{\thetastar} + \ProjBset{\Delta} +
\ProjBsetPerp{\Delta}$.  Using this representation and triangle
inequality, we have
\begin{align*}
 \Reg{\thetastar + \Delta} & \geq \Reg{\ProjModel{\thetastar} +
   \ProjBsetPerp{\Delta}} - \Reg{\ProjModelPerp{\thetastar} +
   \ProjBset{\Delta}} \; \geq \; 
\Reg{\ProjModel{\thetastar} +
   \ProjBsetPerp{\Delta}} - \Reg{\ProjModelPerp{\thetastar}} -
 \Reg{\ProjBset{\Delta}}.
\end{align*}
Finally, since $\ProjModel{\thetastar} \in \ModelSet$ and
$\ProjBsetPerp{\Delta} \in \BsetPerp$, the decomposability of
  $\Regplain$ implies that $\Reg{\ProjModel{\thetastar} +
    \ProjBsetPerp{\Delta}} = \Reg{\ProjModel{\thetastar}} +
  \Reg{\ProjBsetPerp{\Delta}}$, and hence that
\begin{align}
\label{EqnCamembert}
 \Reg{\thetastar + \Delta} & \geq \Reg{\ProjModel{\thetastar}} +
 \Reg{\ProjBsetPerp{\Delta}} - \Reg{\ProjModelPerp{\thetastar}} -
 \Reg{\ProjBset{\Delta}}.
\end{align}
Adding together equations~\eqref{EqnBread} and~\eqref{EqnCamembert},
we obtain the lower bound
\begin{align}
\label{EqnSandwich}
   \Reg{\thetastar + \Delta} - \Reg{\thetastar} & \geq
   \Reg{\ProjBsetPerp{\Delta}} - 2 \Reg{\ProjModelPerp{\thetastar}} -
   \Reg{\ProjBset{\Delta}}.
\end{align}
Combining this lower bound with the earlier inequality~\eqref{EqnMarseille},
some algebra yields the bound
\begin{align*}
\Reg{\ProjBsetPerp{\Delta}} & \leq 3 \Reg{\ProjBset{\Delta}} + 4
\Reg{\ProjModelPerp{\thetastar}} + 2 \frac{\eta}{\regpar},
\end{align*}
corresponding to the bound~\eqref{EqnExtendedCone} when $\eta/\regpar$
achieves the final minimum.  To obtain the final term involving
$\radlag$ in the bound~\eqref{EqnExtendedCone}, two applications of
triangle inequality yields
\begin{align*}
\Reg{\ProjBsetPerp{\Delta}} & \leq \Reg{\ProjBset{\Delta}} +
\Reg{\Delta} \; \leq \; \Reg{\ProjBset{\Delta}} + 2 \radlag,
\end{align*}
where we have used the fact that $\Reg{\Delta} \leq \Reg{\thetapar} +
\Reg{\thetastar} \leq 2 \radlag$, since both $\thetapar$ and
$\thetastar$ are feasible for the program~\eqref{EqnMestReg}.


\subsection{Proof of Lemma~\ref{LemLagDecrease}}

The proof of this result follows lines similar to the proof of
convergence by Nesterov~\cite{Nesterov07}.  Recall our notation
$\obj(\theta) = \LossN(\theta) + \regpar \Reg{\theta}$, $\DelHatIt{t}
= \iter{t} - \thetahat$, and that $\FunHatIt{t} = \obj(\iter{t}) -
\obj(\thetahat)$.  We begin by proving that under the stated
conditions, a useful version of restricted strong
convexity~\eqref{eqn:RSCgeneral} is in force:
\blems
\label{lemEasyRSC}
Under the assumptions of Lemma~\ref{LemLagDecrease}, we are guaranteed
that
\begin{subequations}
\begin{align}
\label{simpleRSC}
\big \{ \frac{\lossrsc}{2} - 32 \FUNLOWALL \Compat^2(\Bset) \big \}
\plainnorm{\DelHatIt{t}}^2 & \leq 2 \FUNLOWALL \, \ALLSLACK^2 \, + \,
\obj(\iter{t}) - \obj(\thetahat), \quad \mbox{and} \\
\label{noConvex}
\big \{ \frac{\lossrsc}{2} - 32 \FUNLOWALL \Compat^2(\Bset) \big \}
\plainnorm{\DelHatIt{t}}^2 & \leq 2 \, \FUNLOWALL \, \ALLSLACK^2 \, +
\, \Tay(\thetahat;\iter{t}),
\end{align}
\end{subequations}
where $\ALLSLACK \defn \staterr + 2
\min(\frac{\objerror}{\regpar},\radlag)$.
\elems
\noindent See Appendix~\ref{AppLemEasyRSC} for the proof of this
claim.  So as to ease notation in the remainder of the proof, let us
introduce the shorthand
\begin{align}
\label{eqn:composite_iter}
\composite_t(\theta) & \defn \LossN(\iter{t}) + \binprod{\grad
  \LossN(\iter{t})}{\theta - \iter{t}} + \frac{\Lsmooth}{2}
\plainnorm{\theta - \iter{t}}^2 + \regpar\Reg{\theta},
\end{align}
corresponding to the approximation to the regularized loss function
$\composite$ that is minimized at iteration $t$ of the
update~\eqref{EqnAlgReg}.  Since $\iter{t+1}$ minimizes $\composite_t$
over the set $\ball{\Regplain}{\radlag}$, we are guaranteed that
\mbox{$\composite_t(\iter{t+1}) \leq \composite_t(\theta)$} for all
$\theta \in \ball{\Regplain}{\radlag}$.  In particular, for any
$\alpha \in (0,1)$, the vector $\theta_\alpha = \alpha \thetahat +
(1-\alpha)\iter{t}$ lies in the convex set
$\ball{\Regplain}{\radlag}$, so that
\begin{align}
\composite_t(\iter{t+1}) \; \leq \; \composite_t(\theta_\alpha) & =
\LossN(\iter{t}) + \binprod{\grad\LossN(\iter{t})}{\theta_\alpha -
  \iter{t}} + \frac{\Lsmooth}{2}\plainnorm{\theta_\alpha - \iter{t}}^2
+ \regpar\Reg{\theta_\alpha} \nonumber \\
& \stackrel{(i)}{=} \LossN(\iter{t}) +
\binprod{\grad\LossN(\iter{t})}{\alpha \thetahat - \alpha \iter{t}} +
\frac{\Lsmooth\alpha^2}{2}\plainnorm{\thetahat - \iter{t}}^2 +
\regpar\Reg{\theta_\alpha} \nonumber \\
%
& \stackrel{(ii)}{\leq} \LossN(\iter{t}) +
\binprod{\grad\LossN(\iter{t})}{\alpha \thetahat - \alpha \iter{t}} +
\frac{\Lsmooth\alpha^2}{2}\plainnorm{\thetahat - \iter{t}}^2 + \regpar
\alpha \Reg{\thetahat} + \regpar (1-\alpha) \Reg{\iter{t}} \nonumber,
\end{align}
where step (i) follows from substituting the definition of
$\theta_\alpha$, and step (ii) uses the convexity of the
regularizer $\Regplain$.

Now, the stated conditions of the lemma ensure that $\lossrsc/2 - 32
\FUNLOWALL \Compat^2(\Bset) \geq 0$, so that by
equation~\eqref{noConvex}, we have $\LossN(\thetahat) + 2 \FUNLOWALL
\ALLSLACK^2 \geq \LossN(\iter{t}) +
\binprod{\grad\LossN(\iter{t})}{\thetahat - \iter{t}}$.  Substituting
back into our earlier bound yields
\begin{align}
\composite_t(\iter{t+1}) & \leq (1-\alpha) \LossN(\iter{t}) + \alpha
\LossN(\thetahat) + 2 \alpha\FUNLOWALL \ALLSLACK^2 +
\frac{\Lsmooth\alpha^2}{2}\plainnorm{\thetahat - \iter{t}}^2 + \alpha
\regpar \Reg{\thetahat} + (1-\alpha) \regpar \Reg{\iter{t}} \nonumber
\\
\label{eqn:thetaalphabound}
& \stackrel{(iii)}{=} \composite(\iter{t}) -
\alpha(\composite(\iter{t}) - \composite(\thetahat)) + 2 \FUNLOWALL
\ALLSLACK^2 + \frac{\Lsmooth\alpha^2}{2}\plainnorm{\thetahat -
  \iter{t}}^2,
\end{align}
where we have used the definition of $\composite$ and $\alpha \leq 1$
in step (iii).

In order to complete the proof, it remains to relate
$\composite_t(\iter{t+1})$ to $\composite(\iter{t+1})$, which can be
performed by exploiting restricted smoothness.  In particular,
applying the RSM condition at the iterate $\iter{t+1}$ in the
direction $\iter{t}$ yields the upper bound
\begin{equation*}
  \LossN(\iter{t+1}) \leq \LossN(\iter{t}) +
  \binprod{\LossN(\iter{t})}{\iter{t+1} - \iter{t}} +
  \frac{\Lsmooth}{2}\plainnorm{\iter{t+1} - \iter{t}}^2 +
  \FUNUP(\LossN)\Regplain^2(\iter{t+1} - \iter{t}), 
\end{equation*}
so that
\begin{align*}
  \composite(\iter{t+1}) &\leq \LossN(\iter{t}) +
  \binprod{\LossN(\iter{t})}{\iter{t+1} - \iter{t}} +
  \frac{\Lsmooth}{2}\plainnorm{\iter{t+1} - \iter{t}}^2 +
  \FUNUP(\LossN)\Regplain^2(\iter{t+1} - \iter{t}) +
  \regpar\Reg{\iter{t+1}}\\
  &= \composite_t(\iter{t+1}) + \FUNUP(\LossN)\Regplain^2(\iter{t+1} -
  \iter{t}).  
\end{align*}
Combining the above bound with the
inequality~\eqref{eqn:thetaalphabound} and recalling the notation
$\DelHatIt{t} = \iter{t} - \thetahat$, we obtain 
\begin{align}
  \nonumber \composite(\iter{t+1}) &\leq \composite(\iter{t}) -
  \alpha(\composite(\iter{t}) - \composite(\thetahat)) +
  \frac{\Lsmooth\alpha^2}{2}\plainnorm{\thetahat - \iter{t}}^2 +
  \FUNUP(\LossN)\Regplain^2(\iter{t+1} - \iter{t}) + 2 \FUNLOWALL
  \ALLSLACK^2 \\ \nonumber &\stackrel{(iv)}{\leq} \composite(\iter{t})
  - \alpha(\composite(\iter{t}) - \composite(\thetahat)) +
  \frac{\Lsmooth\alpha^2}{2}\plainnorm{\DelHatIt{t}}^2 +
  \FUNUP(\LossN)[\Regplain(\DelHatIt{t+1}) +
    \Regplain(\DelHatIt{t})]^2 + 2 \FUNLOWALL \ALLSLACK^2
  \\ &\stackrel{(v)}{\leq} \composite(\iter{t}) -
  \alpha(\composite(\iter{t}) - \composite(\thetahat)) +
  \frac{\Lsmooth\alpha^2}{2}\plainnorm{\DelHatIt{t}}^2 +
  2\FUNUP(\LossN)(\Regplain^2(\DelHatIt{t+1}) +
  \Regplain^2(\DelHatIt{t})) + 2 \FUNLOWALL \ALLSLACK^2.
  \label{eqn:comptodeltabound}
\end{align}
Here step (iv) uses the fact that $\iter{t} - \iter{t+1} =
\DelHatIt{t} - \DelHatIt{t+1}$ and applies triangle inequality to the
norm $\Regplain$, whereas step (v) follows from Cauchy-Schwarz
inequality. 

Next, combining Lemma~\ref{LemICB} with the Cauchy-Schwarz inequality
inequality yields the upper bound
\begin{equation}
  \Regplain^2(\DelHatIt{t}) \; \leq 32 \Compat^2(\Bset)
  \plainnorm{\DelHatIt{t}}^2 + 2 \ALLSLACK^2
  \label{eqn:lemicbsquared}
\end{equation}
where $\ALLSLACK = \staterr(\ModelSet,\Bset) + 2
\min(\frac{\objerror}{\regpar},\radlag)$, is a constant independent of
$\iter{t}$ and $\staterr(\ModelSet, \Bset)$ was previously defined in
the lemma statement.  Substituting the above bound into
inequality~\eqref{eqn:comptodeltabound} yields that
$\obj(\iter{t+1})$ is at most
\begin{multline}
\label{eqn:compboundalmostthere}
 \obj(\iter{t}) - \alpha (\obj(\iter{t}) - \obj(\thetahat)) +
 \frac{\Lsmooth \alpha^2}{2} \plainnorm{\DelHatIt{t}}^2 + 64
 \FUNUP(\LossN) \Compat^2(\Bset) \plainnorm{\DelHatIt{t+1}}^2
\\ + 64 \FUNUP(\LossN) \Compat^2(\Bset) \plainnorm{\DelHatIt{t}}^2 + 8
\FUNUP(\LossN) \ALLSLACK^2 + 2 \FUNLOWALL \ALLSLACK^2.
\end{multline}

The final step is to translate quantities involving $\DelHatIt{t}$ to
functional values, which may be done using the RSC
condition~\eqref{simpleRSC} from Lemma~\ref{lemEasyRSC}.  In
particular, combining the RSC condition~\eqref{simpleRSC} with the
inequality~\eqref{eqn:compboundalmostthere} yields
\begin{align*}
  \obj(\iter{t+1}) \; \leq \obj(\iter{t}) - \alpha \FunHatIt{t} +
  \frac{\left (\Lsmooth \alpha^2 + 64 \FUNUP(\LossN)
    \Compat^2(\Bset) \right )}{\uplossrsc} (\FunHatIt{t} + 2
  \FUNLOW(\LossN) \ALLSLACK^2) \; +& \\ 
  \frac{64 \FUNUP(\LossN) \Compat^2(\Bset)}{\uplossrsc}(\FunHatIt{t+1}
  + 2 \FUNLOW(\LossN) \ALLSLACK^2) + 8 \FUNUP(\LossN) \ALLSLACK^2 +
  2 \FUNLOWALL \ALLSLACK^2.&
\end{align*}
where we have introduced the shorthand $\uplossrsc \defn \lossrsc - 64
\FUNLOW(\LossN) \Compat^2(\Bset)$.  Recalling the definition of
$\slopb$, adding and subtracting $\obj(\thetahat)$ from both sides,
and choosing $\alpha = \frac{\uplossrsc}{2 \Lsmooth} \in (0,1)$, we
obtain
\begin{equation*}
  \left(1 - \frac{64 \FUNUP(\LossN)
    \Compat^2(\Bset)}{\uplossrsc}\right)\FunHatIt{t+1} \leq \left(1 -
  \frac{\uplossrsc}{4\Lsmooth} + \frac{64 \FUNUP(\LossN)
    \Compat^2(\Bset)}{\uplossrsc}\right)\FunHatIt{t} +
\slopb(\Bset) \ALLSLACK^2.
\end{equation*}
Recalling the definition of the contraction factor $\contrac$ from the
statement of Theorem~\ref{ThmMainLag}, the above expression can be
rewritten as
\begin{equation*}
  \FunHatIt{t+1} \leq \contrac\FunHatIt{t} + \slopb(\Bset)
  \slopa(\Bset) \ALLSLACK^2, \quad \mbox{where $\slopa(\ModelSet) =
    \big \{ 1 - \frac{64 \FUNUP(\LossN) \Compat^2(\Bset)}{\uplossrsc}
    \big \}^{-1}$.}
\end{equation*}
Finally, iterating the above expression yields $\FunHatIt{t} \leq
\contrac^{t-T} \FunHatIt{T} + \frac{\slopa(\Bset)\slopb(\Bset)
  \ALLSLACK^2}{1-\contrac}$, where we have used the condition
$\contrac \in (0,1)$ in order to sum the geometric series, thereby
completing the proof.


\subsection{Proof of Lemma~\ref{lemEasyRSC}}
\label{AppLemEasyRSC}

The key idea to prove the lemma is to use the definition of RSC along
with the iterated cone bound of Lemma~\ref{LemICB} for simplifying the
error terms in RSC. 

Let us first show that condition~\eqref{simpleRSC} holds.  From the
RSC condition assumed in the lemma statement, we have
\begin{align}
  \label{LemEasyHat}
  \LossN(\iter{t}) - \LossN(\thetahat) - \inprod{\nabla \LossN(\thetahat)}{\iter{t}-\thetahat} & \geq \frac{\lossrsc}{2} \,
\plainnorm{\thetahat - \iter{t}}^2 - \FUNLOW(\LossN) \;
\RegSq{\thetahat - \iter{t}}.
\end{align}
From the convexity of $\Regplain$ and definition of the
subdifferential $\partial \Reg{\theta}$, we obtain
\begin{align*}
  \Reg{\iter{t}} - \Reg{\thetahat} - \binprod{\partial
    \Reg{\thetahat}}{\iter{t}-\thetahat} & \geq 0.
\end{align*}
Adding this lower bound with the inequality~\eqref{LemEasyHat} yields
\begin{align*}
\obj(\iter{t}) - \obj(\thetahat) - \inprod{\nabla
  \obj(\thetahat)}{\iter{t}-\thetahat} & \geq \frac{\lossrsc}{2} \,
\plainnorm{\thetahat - \iter{t}}^2 - \FUNLOW(\LossN) \;
\RegSq{\thetahat - \iter{t}},
\end{align*}
where we recall that $\obj(\theta)= \LossN(\theta) + \regpar
\Reg{\theta}$ is our objective function.  By the optimality of
$\thetahat$ and feasibility of $\iter{t}$, we are guaranteed that
\mbox{$\inprod{\nabla \obj(\thetahat)}{\iter{t}-\thetahat} \geq 0$,}
and hence
\begin{align*}
  \obj(\iter{t}) - \obj(\thetahat)  & \geq \frac{\lossrsc}{2} \,
  \plainnorm{\thetahat - \iter{t}}^2 - \FUNLOW(\LossN) \;
\RegSq{\thetahat - \iter{t}} \\
& \stackrel{(i)}{\geq} \frac{\lossrsc}{2} \, \plainnorm{\thetahat -
  \iter{t}}^2 - \FUNLOW(\LossN) \; \ \big \{ 32 \Compat^2(\Bset)
\plainnorm{\thetahat-\iter{t}}^2 + 2 \ALLSLACK^2 \big \}
\end{align*}
where step (i) follows by applying Lemma~\ref{LemICB}.  Some algebra
then yields the claim~\eqref{simpleRSC}. \\

Finally, let us verify the claim~\eqref{noConvex}.  Using the RSC
condition, we have
\begin{align}
\label{LemEasyIter}
\LossN(\thetahat) - \LossN(\iter{t}) - \inprod{\nabla
  \LossN(\iter{t})}{\thetahat - \iter{t}} & \geq \frac{\lossrsc}{2} \,
\plainnorm{\thetahat - \iter{t}}^2 - \FUNLOW(\LossN) \;
\RegSq{\thetahat - \iter{t}}.
\end{align}
As before, applying Lemma~\ref{LemICB} yields
\begin{align*}
\underbrace{\LossN(\thetahat) - \LossN(\iter{t}) - \inprod{\nabla
    \LossN(\iter{t})}{\thetahat - \iter{t}}}_{\Tay(\thetahat;
  \iter{t})} & \geq \frac{\lossrsc}{2} \, \plainnorm{\thetahat -
  \iter{t}}^2 - \FUNLOW(\LossN) \; \left ( 32 \Compat^2(\Bset)
\plainnorm{\thetahat-\iter{t}}^2 + 2 \ALLSLACK^2 \right ), 
\end{align*}
and rearranging the terms and establishes the claim~\eqref{noConvex}.

%

\section{Proof of Lemma~\robreflemcone}
\label{AppLemConeOpt}
Given the condition $\Reg{\thetahat} \leq \radcon \leq
\Reg{\thetastar}$, we have $\Reg{\thetahat} = \Reg{\thetastar +
\DeltaStar} \leq \Reg{\thetastar}$.  By triangle inequality, we have
\begin{align*}
\Reg{\thetastar} & = \Reg{\BigProjModel{\thetastar} +
  \BigProjModelPerp{\thetastar}} \; \leq \;
\Reg{\BigProjModel{\thetastar}} + \Reg{\BigProjModelPerp{\thetastar}}.
\end{align*}
We then write
\begin{align*}
\Reg{\thetastar + \DeltaStar} & = \Reg{\BigProjModel{\thetastar} +
  \BigProjModelPerp{\thetastar} + 
  \BigProjBset{\DeltaStar} + \BigProjBsetPerp{\DeltaStar}} \\
& \stackrel{(i)}{\geq} \Reg{\BigProjModel{\thetastar} +
  \BigProjBsetPerp{\DeltaStar}} - \Reg{\BigProjBset{\DeltaStar}} -
\Reg{\BigProjModelPerp{\thetastar}} \\ 
& \stackrel{(ii)}{=} \Reg{\BigProjModel{\thetastar}} +
\Reg{\BigProjBsetPerp{\DeltaStar}} - \Reg{\BigProjBset{\DeltaStar}} -
\Reg{\BigProjModelPerp{\thetastar}}, 
\end{align*}
where the bound (i) follows by triangle inequality, and step (ii) uses
the decomposability of $\Plainreg$ over the pair $\ModelSet$ and
$\BsetPerp$. By combining this lower bound with the previously
established upper bound
\begin{align*}
\Reg{\thetastar + \DeltaStar} \leq \Reg{\BigProjModel{\thetastar}} +
\Reg{\BigProjModelPerp{\thetastar}},
\end{align*}
we conclude that $\Reg{\BigProjBsetPerp{\DeltaStar}} \leq
\Reg{\BigProjBset{\DeltaStar}} + 2\Reg{\BigProjModelPerp{\thetastar}}$.
Finally, by triangle inequality, we have $\Reg{\DeltaStar} \leq
\Reg{\BigProjBset{\DeltaStar}} + \Reg{\BigProjBsetPerp{\DeltaStar}}$,
and hence
\begin{align*}
\Reg{\DeltaStar} & \leq 2 \Reg{\BigProjBset{\DeltaStar}} +
2\Reg{\BigProjModelPerp{\thetastar}} \\
& \stackrel{(i)}{\leq} 2 \, \Compat(\Bset^\perp)
\norm{\BigProjBset{\DeltaStar}} +
2\Reg{\BigProjModelPerp{\thetastar}} \\
& \stackrel{(ii)}{\leq} 2 \, \Compat(\Bset^\perp) \norm{\DeltaStar}
+ 2\Reg{\BigProjModelPerp{\thetastar}},
\end{align*}
where inequality (i) follows from Definition~\ref{DefnSubspaceCompat}
of the subspace compatibility $\Compat$, and the bound (ii) follows
from non-expansivity of projection onto a subspace.

\section{A general result on Gaussian observation operators}
\label{AppGenGauss}

In this appendix, we state a general result about a Gaussian random
matrices, and show how it can be adapted to prove
Lemmas~\ref{LemGarvesh} and~\ref{LemSahand}.  Let $X \in
\real^{\numobs \times \pdim}$ be a Gaussian random matrix with
i.i.d. rows $x_i \sim N(0, \CovMat)$, where $\CovMat \in \real^{\pdim
\times \pdim}$ is a covariance matrix.  We refer to $X$ as a sample
from the $\CovMat$-Gaussian ensemble. In order to state the result, we
use $\CovMatSqrt$ to denote the symmetric matrix square root.
\bprops
\label{PropGenGauss}
Given a random matrix $\Xmat$ drawn from the $\CovMat$-Gaussian
ensemble, there are universal constants $\PLAINCON_i$, $i = 0, 1$ such
that
\begin{subequations}
\begin{align}
\frac{\|\Xmat \theta\|_2^2}{\numobs} & \geq \frac{1}{2} \|\CovMatSqrt
\theta\|_2^2 - \PLAINCON_1 \frac{(\Exs[\Regdual{x_i}])^2}{\numobs}
\Regplain^2(\theta) \qquad \mbox{and} \\
\frac{\|\Xmat \theta\|_2^2}{\numobs} & \leq 2 \|\CovMatSqrt
\theta\|_2^2 + \PLAINCON_1 \frac{(\Exs[\Regdual{x_i}])^2}{\numobs}
\Regplain^2(\theta) \qquad \mbox{for all $\theta \in \real^\pdim$}
\end{align}
\end{subequations}
with probability greater than $1 - \exp(-\PLAINCON_0 \, \numobs)$.
\eprops

\noindent We omit the proof of this result. The two special instances
proved in Lemma~\ref{LemGarvesh} and~\ref{LemSahand} have been proved
in the papers~\cite{RasWaiYu09} and~\cite{NegWai09} respectively. We
now show how Proposition~\ref{PropGenGauss} can be used to recover
various lemmas required in our proofs.

\paragraph{Proof of Lemma~\ref{LemGarvesh}:}  We begin by establishing
this auxiliary result required in the proof of
Corollary~\ref{CorSparse}.  When $\Reg{\cdot} = \|\cdot\|_1$, we have
$\Regdual{\cdot} = \|\cdot\|_\infty$.  Moreover, the random vector
$x_i \sim N(0, \CovMat)$ can be written as $x_i =
\CovMatSqrt w$, where $w \sim N(0, I_{\pdim \times \pdim})$ is
standard normal.  Consequently, using properties of Gaussian
maxima~\cite{LedTal91} and defining $\rhovec(\CovMat) = \max_{j = 1, 2,
\ldots, \pdim} \CovMat_{jj}$, we have the bound
\begin{equation*}
(\Exs[\|x_i\|_\infty])^2 \; \leq \; \rhovec(\CovMat) \;
(\Exs[\|w\|_\infty])^2 \; \leq \; 3 \rhovec(\CovMat) \; \sqrt{\log
\pdim}.
\end{equation*}
Substituting into Proposition~\ref{PropGenGauss} yields the
claims~\eqref{EqnGarveshLower} and~\eqref{EqnGarveshUpper}.

\paragraph{Proof of Lemma~\ref{LemSahand}:}  

In order to prove this claim, we view each random observation matrix
$X_i \in \real^{\mdim \times \mdim}$ as a $\pdim = \mdim^2$ vector
(namely the quantity $\vect(X_i)$), and apply
Proposition~\ref{PropGenGauss} in this vectorized setting.  Given the
standard Gaussian vector $w \in \real^{\mdim^2}$, we let $W \in
\real^{\mdim \times \mdim}$ be the random matrix such that $\vect(W) =
w$.  With this notation, the term $\Regdual{\vect(X_i)}$ is equivalent
to the operator norm $\opnorm{X_i}$.  As shown in Negahban and
Wainwright~\cite{NegWai09}, $\Exs[\opnorm{X_i}] \leq
24\rhomat(\CovMat) \; \sqrt{\mdim}$, where $\rhomat$ was previously 
defined~\eqref{EqnDefnRhoMat}.

%

\section{Auxiliary results for Corollary~\robrefcorcomp}

In this section, we provide the proofs of Lemmas~\ref{LemMCRSC}
and~\ref{LemMCRSM} that play a central role in the proof of
Corollary~\ref{CorMatComp}.  In order to do so, we require the
following result, which is a re-statement of a theorem due to Negahban
and Wainwright~\cite{NegWai10b}:
\bprops
\label{PropManU}
For the matrix completion operator $\Xop_\numobs$, there are universal
positive constants $(\plaincon_1, \plaincon_2)$ such that
\begin{align}
\biggr | \frac{\|\XopN(\Theta) \|^2_2}{\numobs} -
\matsnorm{\Theta}{F}^2 \biggr | & \leq \plaincon_1 \, \mdim
\|\Theta\|_\infty \; \matsnorm{\Theta}{1} \sqrt{\frac{\mdim \log
\mdim}{\numobs}} \; + \; \plaincon_2 \biggr( \mdim \|\Theta\|_\infty
\sqrt{\frac{\mdim \log \mdim}{\numobs}}\biggr)^2 \qquad \mbox{for all
$\Theta \in \real^{\mdim \times \mdim}$}
\end{align}
with probability at least $1 - \exp(-\mdim \log \mdim)$.
\eprops

\subsection{Proof of Lemma~\ref{LemMCRSC}}
\label{AppLemMCRSC}

Applying Proposition~\ref{PropManU} to $\DelHatIt{t}$ and using the
fact that $\mdim \|\DelHatIt{t}\|_\infty \leq 2\spike$ yields
\begin{align}
\label{EqnMCInitialLower}
\frac{\|\XopN(\DelHatIt{t})\|_2^2}{\numobs} & \geq
\matsnorm{\DelHatIt{t}}{F}^2 - \PLAINCON_1 \spike\matsnorm{\DelHatIt{t}}{1}
\; \sqrt{\frac{\mdim \log \mdim}{\numobs}} - \PLAINCON_2 \,
\spike^2\frac{\mdim \log \mdim}{\numobs},
\end{align}
where we recall our convention of allowing the constants to change
from line to line.  From Lemma~\ref{LemMUFCvsMCBoring},
\begin{align*}
\nuclear{\DelHatIt{t}} & \leq 2 \, \Compat(\Bset^\perp) \,
\frob{\DelHatIt{t}} + 2 \nuclear{\BigProjModelPerp{\thetastar}} + 2
\nuclear{\DeltaStar} + 
\Compat(\Bset^\perp) \frob{\DeltaStar}.
\end{align*}
Since $\radcon \leq \nuclear{\ThetaStar}$, Lemma~\ref{LemConeOpt}
implies that $\nuclear{\DeltaStar} \leq 2 \Compat(\Bset^\perp)
\frob{\DeltaStar} +
\nuclear{\BigProjModelPerp{\thetastar}}$, and hence that
\begin{align}
\nuclear{\DelHatIt{t}} & \leq 2 \, \Compat(\Bset^\perp) \,
\matsnorm{\DelHatIt{t}}{F} + 4 \nuclear{\BigProjModelPerp{\thetastar}}
+ 5 \Compat(\Bset^\perp) \frob{\DeltaStar}.
\label{EqnCompCone}
\end{align}
Combined with the lower bound, we obtain that
$\frac{\|\XopN(\DelHatIt{t})\|_2^2}{\numobs}$ is lower bounded by
\begin{align*}
\frob{\DelHatIt{t}}^2 \Biggr \{ 1 - \frac{2 \PLAINCON_1 \,
  \spike\Compat(\Bset^\perp) \sqrt{\frac{\mdim \log
      \mdim}{\numobs}}}{\frob{\DelHatIt{t}}} \Biggr \} - 2 \PLAINCON_1
\, \spike\sqrt{\frac{\mdim \log \mdim}{\numobs}} \Big \{ 4
\nuclear{\BigProjModelPerp{\thetastar}} + 5 \Compat(\Bset^\perp)
\frob{\DeltaStar} \Big \} - \PLAINCON_2 \, \spike^2\frac{\mdim \log
  \mdim}{\numobs}.
\end{align*}
Consequently, for all iterations such that $\frob{\DelHatIt{t}} \geq 4
\PLAINCON_1 \Compat(\Bset^\perp) \sqrt{\frac{ \mdim \log \mdim}{\numobs}}$,
we have
\begin{align*}
\frac{\|\XopN(\DelHatIt{t})\|_2^2}{\numobs} \geq \frac{1}{2}
\frob{\DelHatIt{t}}^2 - 2 \PLAINCON_1 \, \spike\sqrt{\frac{\mdim \log
    \mdim}{\numobs}} \Big \{ 4 \nuclear{\BigProjModelPerp{\thetastar}}
+ 5 \Compat(\Bset^\perp) \frob{\DeltaStar} \Big \} - \PLAINCON_2 \,
\spike^2\frac{\mdim \log \mdim}{\numobs}.
\end{align*}
By subtracting off an additional term, the bound is valid for all
$\DelHatIt{t}$---viz.
\begin{align*}
\frac{\|\XopN(\DelHatIt{t})\|_2^2}{\numobs} &\geq \frac{1}{2}
\frob{\DelHatIt{t}}^2 - 2 \PLAINCON_1 \, \spike\sqrt{\frac{\mdim \log
    \mdim}{\numobs}} \Big \{ 4 \nuclear{\BigProjModelPerp{\thetastar}}
+ 5 \Compat(\Bset^\perp) \frob{\DeltaStar} \Big \} \\
&\qquad \qquad \qquad \qquad \qquad - \PLAINCON_2 \,
\spike^2\frac{\mdim \log \mdim}{\numobs} - 16 \PLAINCON_1^2
\spike^2\Compat^2(\Bset^\perp) \frac{\mdim \log \mdim}{\numobs}.
\end{align*}

%


\subsection{Proof of Lemma~\ref{LemMCRSM}}
\label{AppLemMCRSM}

Applying Proposition~\ref{PropManU} to $\SHORT$ and using the fact
that $\mdim \|\SHORT\|_\infty \leq 2\spike$ yields
\begin{align}
\label{EqnMCInitialUpper}
\frac{\|\XopN(\SHORT)\|_2^2}{\numobs} & \leq \matsnorm{\SHORT}{F}^2 +
\PLAINCON_1 \spike\matsnorm{\SHORT}{1} \; \sqrt{\frac{\mdim \log
    \mdim}{\numobs}} + \PLAINCON_2 \, \spike^2\frac{\mdim \log
  \mdim}{\numobs},
\end{align}
where we recall our convention of allowing the constants to change
from line to line.  By triangle inequality, we have $\nuclear{\SHORT}
\leq \nuclear{\Iter{t} - \ThetaHat} + \nuclear{\Iter{t+1} - \ThetaHat}
\; = \; \nuclear{\DelHatIt{t}} +
\nuclear{\DelHatIt{t+1}}$. Equation~\ref{EqnCompCone} gives us bounds
on $\nuclear{\DelHatIt{t}}$ and $\nuclear{\DelHatIt{t+1}}$. Substituting
them into the upper bound~\eqref{EqnMCInitialUpper} yields the claim.


\bibliographystyle{plain}
\bibliography{mjwain_super}

\begin{thebibliography}{10}

\bibitem{AgarwalNegWai11}
A.~Agarwal, S.~Negahban, and M.~J. Wainwright.
\newblock Noisy matrix decomposition via convex relaxation: Optimal rates in
  high dimensions.
\newblock {\em To appear in Annals of Statistics}, 2011.
\newblock Appeared as http://arxiv.org/abs/1102.4807.

\bibitem{AmiWai08}
A.~A. Amini and M.~J. Wainwright.
\newblock High-dimensional analysis of semdefinite relaxations for sparse
  principal component analysis.
\newblock {\em Annals of Statistics}, 37:2877--2921, 2009.

\bibitem{BeckTeb09}
A.~Beck and M.~Teboulle.
\newblock A fast iterative shrinkage-thresholding algorithm for linear inverse
  problems.
\newblock {\em SIAM Journal on Imaging Sciences}, 2(1):183--202, 2009.

\bibitem{becker09nesta}
S.~Becker, J.~Bobin, and E.~J. Candes.
\newblock Nesta: a fast and accurate first-order method for sparse recovery.
\newblock {\em SIAM Journal on Imaging Sciences}, 4(1):1--39, 2011.

\bibitem{Bertsekas_nonlin}
D.P. Bertsekas.
\newblock {\em Nonlinear programming}.
\newblock Athena Scientific, Belmont, MA, 1995.

\bibitem{BiRiTsy08}
P.~J. Bickel, Y.~Ritov, and A.~Tsybakov.
\newblock Simultaneous analysis of {L}asso and {D}antzig selector.
\newblock {\em Annals of Statistics}, 37(4):1705--1732, 2009.

\bibitem{Boyd02}
S.~Boyd and L.~Vandenberghe.
\newblock {\em Convex optimization}.
\newblock Cambridge University Press, Cambridge, UK, 2004.

\bibitem{BreLor08}
K.~Bredies and D.~A. Lorenz.
\newblock Linear convergence of iterative soft-thresholding.
\newblock {\em Journal of Fourier Analysis and Applications}, 14:813--837,
  2008.

\bibitem{Bun07}
F.~Bunea, A.~Tsybakov, and M.~Wegkamp.
\newblock Sparsity oracle inequalities for the {L}asso.
\newblock {\em Electronic Journal of Statistics}, pages 169--194, 2007.

\bibitem{CandesLiMaWr2009}
E.~J. {Candes}, X.~{Li}, Y.~{Ma}, and J.~{Wright}.
\newblock {Robust Principal Component Analysis?}
\newblock {\em J. ACM}, 58:11:1--11:37, 2011.

\bibitem{CanRec08}
E.~J. Cand{\`e}s and B.~Recht.
\newblock Exact matrix completion via convex optimization.
\newblock {\em Found. Comput. Math.}, 9(6):717--772, 2009.

\bibitem{Chand09}
V.~Chandrasekaran, S.~Sanghavi, P.~Parrilo, and A.~Willsky.
\newblock Rank-sparsity incoherence for matrix decomposition.
\newblock {\em SIAM J. on Optimization}, 21(2):572--596, 2011.

\bibitem{Chen98}
S.~Chen, D.~L. Donoho, and M.~A. Saunders.
\newblock Atomic decomposition by basis pursuit.
\newblock {\em SIAM J. Sci. Computing}, 20(1):33--61, 1998.

\bibitem{DuchiSSSiCh08}
J.~Duchi, S.~Shalev-Shwartz, Y.~Singer, and T.~Chandra.
\newblock Efficient projections onto the $\ell_1$-ball for learning in high
  dimensions.
\newblock In {\em ICML}, 2008.

\bibitem{Dykstra85}
R.~L. Dykstra.
\newblock An iterative procedure for obtaining i-projections onto the
  intersection of convex sets.
\newblock {\em Annals of Probability}, 13(3):975--984, 1985.

\bibitem{Fa02}
M.~Fazel.
\newblock {\em Matrix Rank Minimization with Applications}.
\newblock PhD thesis, Stanford, 2002.
\newblock Available online:
  http://faculty.washington.edu/mfazel/thesis-final.pdf.

\bibitem{GarKha09}
R.~Garg and R.~Khandekar.
\newblock Gradient descent with sparsification: an iterative algorithm for
  sparse recovery with restricted isometry property.
\newblock In {\em ICML}, 2009.

\bibitem{HalWotZha08}
E.~T. Hale, Y.~Wotao, and Y.~Zhang.
\newblock Fixed-point continuation for $\ell_1$-minimization: Methodology and
  convergence.
\newblock {\em SIAM J. on Optimization}, 19(3):1107--1130, 2008.

\bibitem{HsuKaZh2010}
D.~Hsu, S.M. Kakade, and Tong Zhang.
\newblock Robust matrix decomposition with sparse corruptions.
\newblock {\em IEEE Trans. Info. Theory}, 57(11):7221 --7234, 2011.

\bibitem{HuaZha09}
J.~Huang and T.~Zhang.
\newblock The benefit of group sparsity.
\newblock {\em The Annals of Statistics}, 38(4):1978--2004, 2010.

\bibitem{Yi09}
S.~Ji and J.~Ye.
\newblock An accelerated gradient method for trace norm minimization.
\newblock In {\em ICML}, 2009.

\bibitem{KoltchinskiiLoTs2011}
V.~Koltchinskii, K.~Lounici, and A.~B. Tsybakov.
\newblock Nuclear-norm penalization and optimal rates for noisy low-rank matrix
  completion.
\newblock {\em Annals of Statistics}, 39:2302--2329, 2011.

\bibitem{LedTal91}
M.~Ledoux and M.~Talagrand.
\newblock {\em Probability in Banach Spaces: Isoperimetry and Processes}.
\newblock Springer-Verlag, New York, NY, 1991.

\bibitem{LeeBres09}
K.~Lee and Y.~Bresler.
\newblock Guaranteed minimum rank approximation from linear observations by
  nuclear norm minimization with an ellipsoidal constraint.
\newblock Technical report, UIUC, 2009.
\newblock Available at arXiv:0903.4742.

\bibitem{Lou09}
K.~Lounici, M.~Pontil, A.~B. Tsybakov, and S.~van~de Geer.
\newblock Taking advantage of sparsity in multi-task learning.
\newblock In {\em COLT}, 2009.

\bibitem{LuoTse93}
Z.~Q. Luo and P.~Tseng.
\newblock Error bounds and convergence analysis of feasible descent methods: a
  general approach.
\newblock {\em Annals of Operations Research}, 46-47:157--178, 1993.

\bibitem{Meinshausen06}
N.~Meinshausen and P.~B\"uhlmann.
\newblock High-dimensional graphs and variable selection with the {L}asso.
\newblock {\em Annals of Statistics}, 34:1436--1462, 2006.

\bibitem{NegRavWaiYu09}
S.~Negahban, P.~Ravikumar, M.~J. Wainwright, and B.~Yu.
\newblock A unified framework for high-dimensional analysis of {M}-estimators
  with decomposable regularizers.
\newblock In {\em NIPS}, 2009.
\newblock To appear in Statistical Science.

\bibitem{NegWai09}
S.~Negahban and M.~J. Wainwright.
\newblock Estimation of (near) low-rank matrices with noise and
  high-dimensional scaling.
\newblock {\em Annals of Statistics}, 39(2):1069--1097, 2011.

\bibitem{NegWai10b}
S.~Negahban and M.~J. Wainwright.
\newblock Restricted strong convexity and (weighted) matrix completion: Optimal
  bounds with noise.
\newblock {\em Journal of Machine Learning Research}, 13:1665--1697, May 2012.

\bibitem{Nesterov04}
Y.~Nesterov.
\newblock {\em Introductory Lectures on Convex Optimization}.
\newblock Kluwer Academic Publishers, New York, 2004.

\bibitem{Nesterov07}
Y.~Nesterov.
\newblock Gradient methods for minimizing composite objective function.
\newblock Technical Report~76, Center for Operations Research and Econometrics
  (CORE), Catholic University of Louvain (UCL), 2007.

\bibitem{NgailPe2008}
H.~V. Ngai and J.~P. Penot.
\newblock Paraconvex functions and paraconvex sets.
\newblock {\em Studia Mathematica}, 184:1--29, 2008.

\bibitem{RasWaiYu10}
G.~Raskutti, M.~J. Wainwright, and B.~Yu.
\newblock Restricted eigenvalue conditions for correlated {G}aussian designs.
\newblock {\em Journal of Machine Learning Research}, 11:2241--2259, August
  2010.

\bibitem{RasWaiYu09}
G.~Raskutti, M.~J. Wainwright, and B.~Yu.
\newblock Minimax rates of estimation for high-dimensional linear regression
  over $\ell_q$-balls.
\newblock {\em IEEE Trans. Info. Theory}, 57(10):6976---6994, 2011.

\bibitem{Recht09}
B.~Recht.
\newblock A simpler approach to matrix completion.
\newblock {\em Journal of Machine Learning Research}, 12:3413--3430, 2011.

\bibitem{RecFazPar10}
B.~Recht, M.~Fazel, and P.~Parrilo.
\newblock Guaranteed minimum-rank solutions of linear matrix equations via
  nuclear norm minimization.
\newblock {\em SIAM Review}, 52(3):471--501, 2010.

\bibitem{RohTsy10}
A.~Rohde and A.~Tsybakov.
\newblock Estimation of high-dimensional low-rank matrices.
\newblock {\em Annals of Statistics}, 39(2):887--930, 2011.

\bibitem{RudZho11}
M.~Rudelson and S.~Zhou.
\newblock Reconstruction from anisotropic random measurements.
\newblock Technical report, University of Michigan, July 2011.

\bibitem{SreAloJaa05}
N.~Srebro, N.~Alon, and T.~S. Jaakkola.
\newblock Generalization error bounds for collaborative prediction with
  low-rank matrices.
\newblock In {\em NIPS}, 2005.

\bibitem{Tibshirani96}
R.~Tibshirani.
\newblock Regression shrinkage and selection via the lasso.
\newblock {\em Journal of the Royal Statistical Society, Series B},
  58(1):267--288, 1996.

\bibitem{TroppGil07}
J.~A. Tropp and A.~C. Gilbert.
\newblock Signal recovery from random measurements via orthogonal matching
  pursuit.
\newblock {\em IEEE Trans. Info. Theory}, 53(12):4655--4666, 2007.

\bibitem{GeerBuhl09}
S.~van~de Geer and P.~Buhlmann.
\newblock On the conditions used to prove oracle results for the lasso.
\newblock {\em Electronic Journal of Statistics}, 3:1360--1392, 2009.

\bibitem{XuCaSa2010}
H.~Xu, C.~Caramanis, and S.~Sanghavi.
\newblock Robust {PCA} via outlier pursuit.
\newblock {\em IEEE Trans. Info. Theory}, 58(5):3047 --3064, May 2012.

\bibitem{Huang06}
C.~H. Zhang and J.~Huang.
\newblock The sparsity and bias of the lasso selection in high-dimensional
  linear regression.
\newblock {\em Annals of Statistics}, 36(4):1567--1594, 2008.

\bibitem{ZhaRocYu09}
P.~Zhao, G.~Rocha, and B.~Yu.
\newblock Grouped and hierarchical model selection through composite absolute
  penalties.
\newblock {\em Annals of Statistics}, 37(6{A}):3468--3497, 2009.

\end{thebibliography}

\end{document}